\documentclass[journal]{IEEEtran}
\usepackage[T1]{fontenc}
\usepackage{placeins}
\usepackage{amsmath,amsfonts}
\usepackage{algorithmic}
\usepackage{algorithm}
\usepackage{array}
\usepackage[caption=false,font=normalsize,labelfont=sf,textfont=sf]{subfig}
\usepackage{textcomp}
\usepackage{stfloats}
\usepackage{url}
\usepackage{verbatim}
\usepackage{graphicx}
\usepackage[square,sort,comma,numbers]{natbib}
\usepackage{booktabs}
\usepackage{multirow}
\usepackage{multicol}
\usepackage{mathtools}
\usepackage{ragged2e}
\usepackage{minted}
\usepackage[export]{adjustbox}
\usepackage{listings}

\usepackage{newfloat}
\DeclareCaptionStyle{ruled}{labelfont=normalfont,labelsep=colon,strut=off} % DO NOT CHANGE THIS
\floatstyle{ruled}
\newfloat{listing}{tb}{lst}{}
\floatname{listing}{Listing}

\usepackage[capitalize]{cleveref}
\crefname{section}{Sec.}{Secs.}
\Crefname{section}{Section}{Sections}
\Crefname{table}{Table}{Tables}
\crefname{table}{Tab.}{Tabs.}
\crefname{paragraph}{para.}{paras.}
\Crefname{paragraph}{Para.}{Paras.}
\usepackage{xcolor}
\definecolor{cadmiumgreen}{rgb}{0.0, 0.42, 0.24}
\definecolor{custom}{cmyk}{0.1,0.48,0.49,0.2}
\definecolor{OliveGreen}{cmyk}{0.64,0,0.95,0.40}
\definecolor{new}{rgb}{0.81,0.05,0.9}
\definecolor{BrickRed}{rgb}{0.81,0.1,0.1}
\definecolor{RoyalBlue}{rgb}{0.2,0.2,0.75}

   \usepackage[normalem]{ulem}
   \usepackage{bm}
\makeatletter
\DeclareRobustCommand\onedot{\futurelet\@let@token\@onedot}
\def\@onedot{\ifx\@let@token.\else.\null\fi\xspace}

\makeatother

\def\clap#1{\hbox to 0pt{\hss #1\hss}}%
\def\initials#1{\protect\clap{\protect\smash{\protect\raisebox{1.4ex}{\protect\tiny{\protect\textsf{\protect\textit{#1}}}}}}}%
\makeatletter
\newcommand{\EDIT}[4][]{\protect\@ifundefined{hidecomments}{%
  \protect\strut{\color{#3}{\hspace{0pt}\initials{#2}\protect\sout{#1}{~#4}}}%
  }{#4}}
\newcommand{\NOTEboxed}[3]{\protect\@ifundefined{hidecomments}{%
  {\begin{center}\fbox{\parbox{0.97\linewidth}{\protect\EDIT{#1}{#2}{#3}}}\end{center}}
  }{}}
\newcommand{\COMM}[3]{\protect\@ifundefined{hidecomments}{%
  {\protect\EDIT{#1}{#2}{#3}}%
  }{}}
\newcommand{\DefAuthor}[2] % initials, color
{%
  \expandafter\newcommand\csname #1edit\endcsname[2][]{\protect\EDIT[##1]{#1}{#2}{##2}}
  \expandafter\newcommand\csname #1\endcsname[1]{\protect\COMM{#1}{#2}{##1}}
  \expandafter\newcommand\csname #1boxed\endcsname[1]{\protect\NOTEboxed{#1}{#2}{##1}}
}
\definecolor{dfltgreen}       {rgb}{0.0,0.5,0.0}
\definecolor{dfltred}         {rgb}{0.7,0.0,0.0}
\newcommand{\REVadd}[1]{\protect\@ifundefined{hidecomments}{%
  \strut{\color{dfltgreen}{#1}}}{#1}}
\newcommand{\REVedit}[2][]{\protect\@ifundefined{hidecomments}{%
  \strut{\color{dfltred}{\protect\sout{#1}}\color{dfltgreen}{~#2}}}%
  {#2}}
\makeatother

\definecolor{dkgreen}       {rgb}{0.0,0.5,0.0}
\definecolor{dkblue}        {rgb}{0.0,0.0,0.7}
\definecolor{dkcyan}        {rgb}{0.0,0.5,0.5}
\definecolor{dkmagenta}     {rgb}{0.5,0.0,0.5}
\DefAuthor{MK}{dkmagenta} % MARGRET'S COMMENTS
\DefAuthor{AB}{dkgreen} % ANDRES' COMMENTS
\DefAuthor{JS}{dkblue} % JENNY'S COMMENTS
\DefAuthor{SC}{dkcyan} % STUDENTS' COMMENTS
\DefAuthor{SA}{orange} % SHASHANK'S COMMENTS
\usepackage{pifont}% http://ctan.org/pkg/pifont

\begin{document}

\title{Beware of Aliases -- Signal Preservation is Crucial for Robust Image Restoration}

\author{Shashank Agnihotri$^1$\thanks{$^1$University of Mannheim}, Julia Grabinski$^{1,2}$\thanks{$^2$Institute for Machine Learning and Analytics, Offenburg University, Offenburg, Germany}, Janis Keuper$^{1,2}$, Margret Keuper$^{1,3}$\thanks{$^3$Max Planck Institute for Informatics, Saarland Informatics Campus, Saarbrücken, Germany}
}

%\markboth{Under Review IEEE Transactions on Image Processing}%
%{Agnihotri \MakeLowercase{\textit{et al.}}: Beware of Aliases}

%\IEEEpubidadjcol

\maketitle

\begin{abstract}
Anti-aliasing methods have recently shown promising results in enhancing deep neural networks (DNNs) to learn better feature representations.
Most of these works are limited to image classification tasks, though, for real-world tasks such as image restoration the high-frequency information discarded by the anti-aliasing methods can be of paramount importance to restore image details like edges.
However, providing the DNN with a path to propagate high-frequency information comes at the risk of spectral artifacts forming in the restored images, which might not always be visible to the human eye. These artifacts can be analysed using adversarial attacks which make them more apparent.
In this work, we show that providing alias-free paths in state-of-the-art restoration networks improves model robustness at low costs on the restoration performance. 
We do so by proposing \textbf{B}eware \textbf{o}f \textbf{A}liases (BoA)-Networks, a modification to the state-of-the-art image restoration models that executes downsampling and upsampling operations partly in the frequency domain to ensure alias-free paths along the entire model while potentially preserving all relevant high-frequency information. 
We modify both CNN-based and Transformer-based image restoration models with Beware Of Aliases (BoA) modifications and study their performance across image restoration tasks like image deblurring and image deraining.
\end{abstract}

\begin{IEEEkeywords}
Anti-aliasing, robustness, adversarial robustness, image restoration, image deraining, image deblurring, signal processing, image processing, Fourier transform, low-pass filter.
\end{IEEEkeywords}

\section{Introduction}
\label{sec:intro}
\IEEEPARstart{C}{omputer} vision, an area with tremendous advances in recent years due to improved machine learning models, is fundamentally a form of digital signal processing. % which has greatly benefited from the recent advances in machine learning, especially Deep Neural Networks (DNNs).
However, in the course of modifying machine learning methods for vision tasks, some essential signal processing fundamentals have been overlooked, potentially risking unexpected model behavior.
This leads to deep neural networks (DNNs) learning sub-optimal feature representations from the provided data resulting in aliasing and other spectral artifacts in the learned feature maps. 
Most models can compensate for those artifacts on in-domain data while significant problems arise under distribution shifts like those encountered under adversarial attacks.
Recent works such as \cite{grabinski2022aliasing, zou2020delving, grabinski2022AAAIw} have explored this problem for image classification models.
However, spectral artifacts occurring in the feature maps due to inadequate sampling are equally crucial for pixel-wise prediction tasks such as image restoration which are often involved in real-world applications that might affect human safety.
Therefore, reducing these spectral artifacts is of paramount importance for a model's reliability in real-world settings. % under noise and for the model's robustness
%~\cite{hoffmann2021towards}.
Spectral artifacts might emerge as grid artifacts or checkerboard artifacts~\cite{checkerboard_odena2016deconvolution}, ringing artifacts~\cite{ringing_artifacts}, or lattice artifacts~\cite{lattice_structure}.
\begin{figure}[t]
    \centering % <-- added
    \scalebox{0.92}{
    \begin{tabular}{@{}c@{\hspace{0.03cm}}c@{}}
    {\small FSNet~\cite{cui2023selective}} & {\small Restormer~\cite{zamir2022restormer}} \\
    \includegraphics[width=0.5\linewidth]{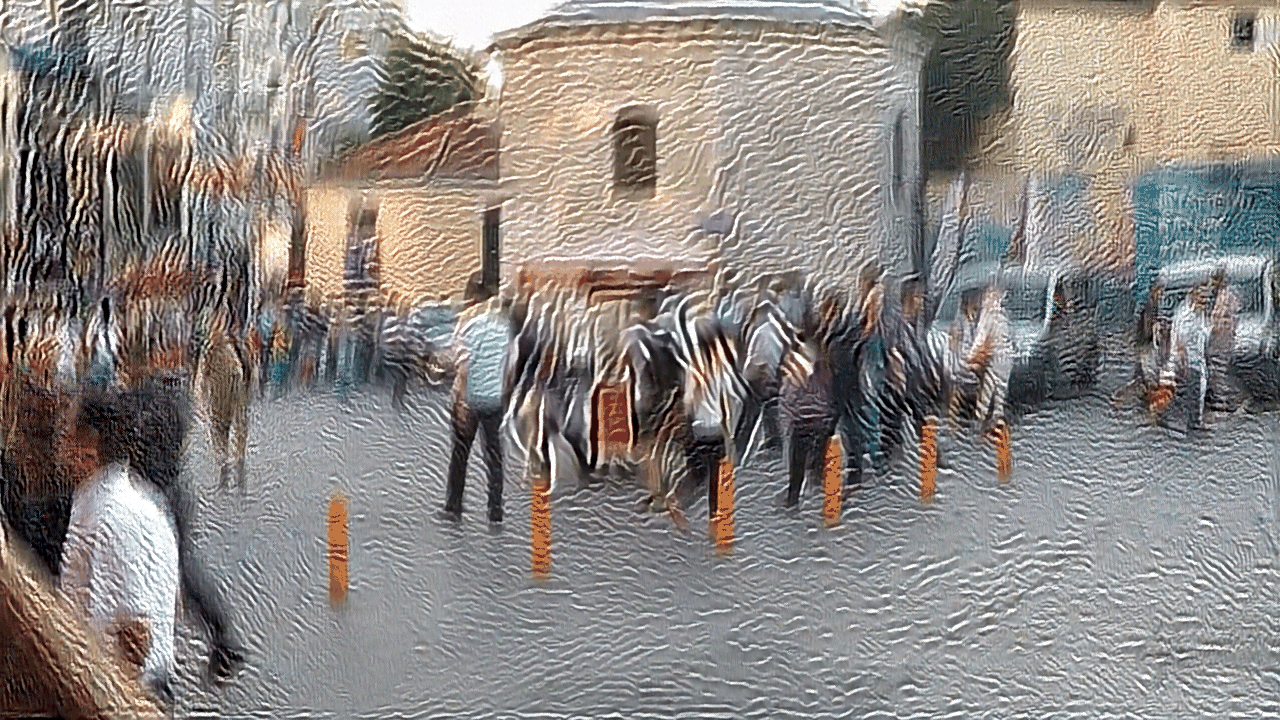} &
    \includegraphics[width=0.5\linewidth]{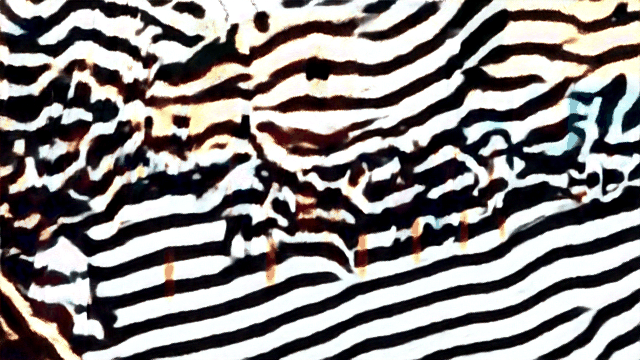} \\
    PSNR: 16.54 & PSNR: 9.04 \\
    {\small FLC Pooling~\cite{grabinski2022frequencylowcut}} & {\small \textbf{BoA}-Restormer \textbf{(proposed)}} \\
    \includegraphics[width=0.5\linewidth]{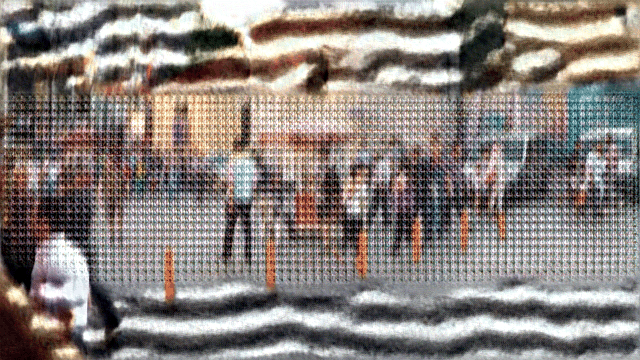} &
    \includegraphics[width=0.5\linewidth]{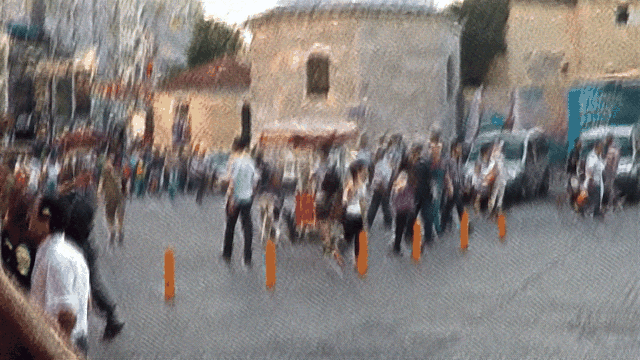} \\
    PSNR: 16.00 & \textbf{PSNR: 18.91} \\
\end{tabular}
}
\caption{Comparing images deblurred by image restoration models after 10 iterations of PGD attack using the GoPro dataset.
While other approaches exhibit strong spectral artifacts in restored images, our proposed \textbf{BoA}-modifications significantly reduce these artifacts. PSNR values are averaged over the entire test set.}
\label{fig:teaser}
\end{figure}
Often these are visible to the human eye. 
However in some independently and identically distributed (i.i.d.) scenarios, the network might have learned shortcuts~\cite{shortcut} around it, to project a sample signal from the input distribution to the target distribution.
In such scenarios, these spectral artifacts, while still existing in the learned feature representations, might not appear in the restored images and provide a wrong sense of safety~\cite{agnihotri2023improving}.
Thus, slightly perturbing the input distribution such that the perturbed and original input samples still look similar to the human eye, helps in exposing these underlying inadequately learned feature representations.
%These would again be visible to the human eye when performing pixel-wise prediction tasks.

This can be achieved by performing white-box adversarial attacks on the input images, as they optimize the perturbations by utilizing the availability of a DNN's learned weights.
Therefore, we use white-box adversarial attacks to accentuate the spectral artifacts in the restored images and expose the vulnerabilities of the DNN's learned feature representations~\cite{agnihotri2023unreasonable}.
\cref{fig:teaser} provides an example of this effect, where FSNet~\cite{cui2024image}, Restormer~\cite{zamir2022restormer} and FLC Pooling~\cite{grabinski2022frequencylowcut} have visible artifacts under PGD~\cite{pgd} attack.

Some proposed computer vision models attempt to reduce spectral artifacts using low-pass filters in the Fourier domain when downsampling feature maps~\cite{grabinski2022frequencylowcut,cui2023selective,grabinski2023fix}, motivated by the fact that convolution kernels widely used for downsampling in DNNs sample below the Nyquist rate and thus introduce artifacts in the learned feature representations~\cite{grabinski2022frequencylowcut}.
However, depriving the DNN of the high-frequency information might have detrimental effects on tasks such as edge detection, image restoration, or shape analysis.
One such task is image deblurring, where the blurring has already deprived the DNN of some essential high-frequency information like edges, and shapes. 
The restoration DNNs are expected to learn to restore them.
However, the contrary is true for tasks such as image denoising and image deraining, where the DNN is expected to learn to remove high-frequency noise from an image.

Thus, in this work, we study both the downsampling and upsampling operations performed in a DNN for the image restoration tasks.
We propose a novel and theoretically motivated sampling methods, \textbf{FrequencyPreservedPooling (FP)} for downsampling and \textbf{FreqAvgUp} for upsampling that sample information efficiently to ensure that the artifacts in the learned representations are significantly reduced, as shown in \cref{fig:teaser} (bottom right).

The key contributions of this work are as follows:
\begin{itemize}
    \item A novel downsampling method \textbf{FrequencyPreservedPooling}, that uses the Fourier domain to downsample feature maps \emph{without} compromising the high-frequency features and simultaneously making the network focus on low-frequency features. This increases the stability of the learned features.
    \item A novel upsampling method \textbf{FreqAvgUp} which uses the frequency domain to help restore some high-frequency information lost during downsampling. We refer to both modules used together as \emph{Beware of Aliases (BoA)}-Pooling.
    \item We study these architectural design choices for both CNN-based and Vision Transformer-based SotA image restoration networks over different image restoration tasks.
\end{itemize}

\section{Related Work}
\label{sec:related}

\paragraph{Adversarial Robustness}
\label{subsec:related:robustness}

Adversarial robustness refers to  a model's ability to withstand intentionally crafted image perturbations designed to fool the network. 
Such attacks are commonly used to evaluate the quality of a network's learned representations~\cite{schmalfuss2022attacking,schrodi2022towards,agnihotri2023unreasonable}.
While the first adversarial attacks focused on classification networks \cite{c_and_w,fgsm,pgd} subsequent work introduced adversarial attacks for a broader range of pixel-wise tasks~\cite{agnihotri2023cospgd,segpgd,schmalfuss2022perturbationconstrained}. 
Our work shows that a model's robustness can be enhanced by providing an aliasing-free path for the downsampling and upsampling in image restoration models.

\paragraph{Aliasing Artifacts}
\label{subsec:related:artifacts}

Recent work on model robustness demonstrates that aliasing artifacts %~\cite{grabinski2022aliasing,li2021wavecnet,zhang2019making,zou2020delving} 
tend to increase model vulnerability in classification models. 
\cite{zhang2019making,zou2020delving} enhance a model's shift-invariance and reduce aliases by blurring before downsampling, \cite{li2021wavecnet} use wavelet representations, and %against aliasing and for more robustness against common corruptions, 
\cite{hossain2023anti} introduce a novel activation function in combination with blurring before downsampling as a remedy. % and \cite{grabinski2022aliasing} show that robust models learn to exhibit less aliasing after downsampling. 
These works primarily focus on image encoding and artifacts introduced during downsampling. However, \cite{durall2020watch,jung2021spectral,karras2021alias} point out that aliasing is equally harmful during upsampling in the context of GANs. Recent work by \cite{agnihotri2023improving} leverages this insight to improve representations learned for encoder-decoder networks using Large Transposed Convolution Kernels (LCTC) for upsampling. 
%Also in the field of image generation reducing aliasing improves the quality of the generated images~\cite{durall2020watch,jung2021spectral,karras2021alias}. 
%Hence, anti-aliasing components within the networks \cite{karras2021alias} or specially adapted loss terms \cite{jung2021spectral,durall2020watch} can improve the quality of the generated images. 

\paragraph{Frequency Learning in Image Restoration}
Operating parts of a network in the frequency domain can enhance global information~\cite{ffc2020lu,grabinski2023large,rao2021global} or improve downsampling in classification tasks~\cite{grabinski2022frequencylowcut,li2021wavecnet}. Yet, pixel-wise prediction tasks like image restoration also apply upsampling. 
Thus prior work in this direction used discrete wavelet transforms to replace the up- and downsampling in a UNet~\cite{hao2019discrete}. 
Furthermore, FSNet~\cite{cui2024image} uses a multi-branch approach to select the most informative frequencies during network training and enhance several image restoration tasks. 
%\cite{xint2023freqsel} apply a ReLU function on the frequencies of the feature map to gain insightful information on the blur pattern in image deblurring.
In MRLPFNets~\cite{Dong_2023_ICCV} a low-pass filter is learned %to reduce high-frequency noise 
and the multi-scale features from different stages of the network are fused via a wavelet-based feature fusion.

All these approaches demonstrate that frequency learning can enhance image restoration tasks. However, while they focus on clean task improvements, our approach is inherently more robust due to the anti-aliasing path we establish.

\paragraph{Sampling Methods}
\label{subsec:related:sampling}
Since aliasing usually arises from incorrect sampling, most existing work focused on the improvement of downsampling by dedicated blurring techniques. %~\cite{hossain2023anti,karras2021alias,li2021wavecnet,zhang2019making,zou2020delving}. 
While these approaches only reduce the amount of aliasing, \cite{grabinski2022frequencylowcut,grabinski2023fix} propose a \textit{provably} aliasing-free downsampling operation. However, this approach only includes the downsampling while for pixel-wise prediction tasks like image restoration upsampling is equally important. Further, all these approaches remove aliases by removing high-frequencies, the exact information that one would want to restore for deblurring. 
Thus, we propose a pair of down- and upsampling operations that provide the network with a completely aliasing-free path \textit{while being able to retain high-frequency information}.

Upsampling is traditionally done via Transpose Convolution \cite{segnet,goodfellow2020generative, long2015fully, noh2015learning,unet} or Interpolation \cite{psmnet,semsegzhao2017pspnet,zhao2018psanet} which both suffer from spectral artifacts introduced due to incorrect sampling~\cite{checkerboard_odena2016deconvolution,agnihotri2023improving}. 
%In Transpose Convolutions, the kernels can overlap leading to uneven contributions of the different pixels which lead to grid-like artifacts~\cite{checkerboard_odena2016deconvolution}. Linear interpolation can lead to aliasing artifacts which are caused by the limited context during interpolation~\cite{agnihotri2023improving}. 

Recent transformer-based models employ \textbf{Pixel Unshuffle and Shuffle}~\cite{pixel_shuffle_unshuffle} to spatially aggregate or upsample feature maps, respectively. Therefore, our proposed sampling operator builds upon these operations. 
Pixel Shuffle combines $r^2$ (i.e.~ usually four) channels by organizing the values from the same channel dimension in a grid such as to increasing the spatial resolution (height and width) of the feature maps by a factor $r$ (i.e.~usually two). Pixel Unshuffle is the inverse operation and decomposes each feature map channel into $r^2$ channels of decrease spatial resolution by factor $r$. %  in height and width to increase the channel dimension by a factor of $r^2$.  Hence, the spatial resolution of the feature maps can easily be increased (upsampling) by aggregating channels or decreased (downsampling) by expanding the channel dimension while
Unlike transposed convolutions, Pixel Shuffle operations have no learnable parameters. They maintain all information present in the feature maps and preserve the overall feature map dimension. Yet, due to the re-organization of channel dimensions, %Pixel Unshuffle performs downsampling with sampling rate $r$ resulting in 
severe aliasing artifacts~\cite{grabinski2022frequencylowcut} in each channel.

%All these methods are prone to spectral artifacts like aliasing. 
In contrast, we propose modified Pixel Shuffle and Unshuffle operations that separate high- and low-frequency information, allowing the model to access alias-free data while preserving relevant high-frequency details.
%In contrast, we propose Pixel Shuffle and Unshuffle operations that learn high- and low-frequency information in separate paths. Thus, the model has access to alias-free information at any point during sampling while retaining relevant high frequency information. % for  and thus avoid aliases in one path of our down- and upsampling method, while still incorporating the strength of Pixel Unshuffle and Pixel Shuffle in the other path.

\section{Method}
\label{sec:method}

\begin{figure*}[t]
\begin{center}
\scriptsize
    \begin{tabular}{@{}c@{\hspace{2cm}}c@{}}
   FrequencyPreservedPooling Downsampling block (Encoder) \qquad FreqAvgUp Upsampling block  (Decoder)& Abstract BoA-Network Architecture \\        
{\includegraphics[height=6.1cm]{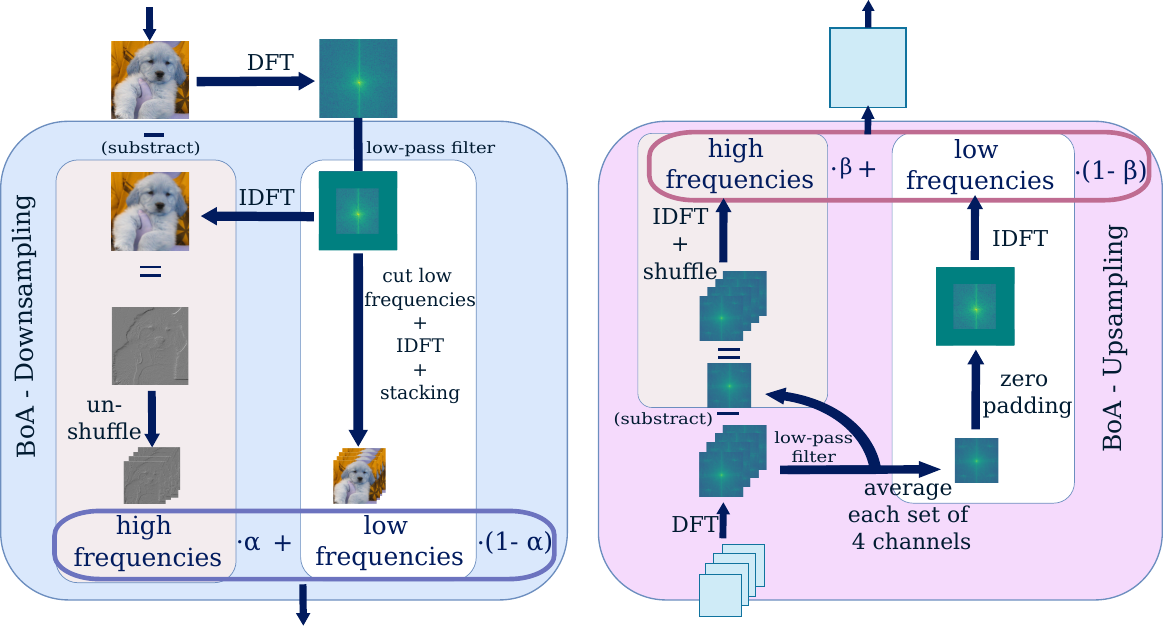}}&
      \includegraphics[width=0.23\linewidth]{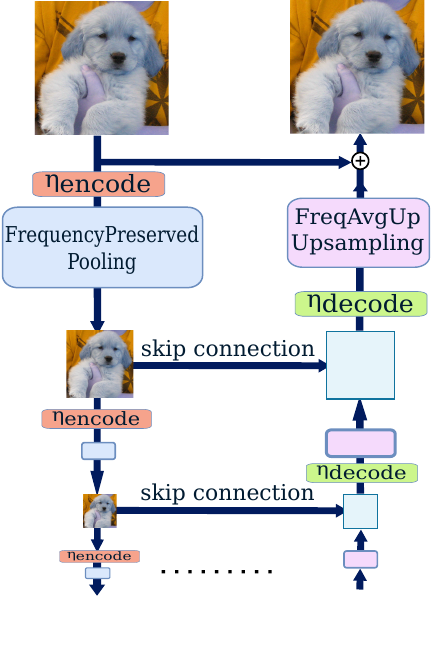}     
%\caption{BoA-network architecture using our up- and downsampling from \cref{fig:complete_boa} and skip connections between encoder and decoder.}

    \end{tabular}
\caption{A visual representation of the proposed sampling operations. Left: Flow diagram for our proposed downsampling operation ``FrequencyPreservedPooling'' used in the encoder of the image restoration model's architecture. Center: flow diagram for our proposed upsampling operation ``FreqAvgUp'' used in the decoder of the BoA architecture. Right: Overview of the architectural change. }
\label{fig:complete_boa}
\end{center}    

\end{figure*}

We propose a method to preserve the high-frequency information while making the network also focus on low-frequency information (please refer to \cref{fig:complete_boa}). 
To this effect, we split the feature maps into their low-frequency and high-frequency components. We then learn the proper mixture to enhance the stability of the feature maps that the network learns, while also providing the network with an \emph{alias-free} information path.

Let $\eta_\mathit{encode}$ be the repeating blocks of the encoder and $\eta_\mathit{decode}$ be the repeating blocks of the decoder, such that each $\eta_\mathit{encode}$ is followed by a downsampling operation $\zeta_{down}$ and each decoder block is followed by an upsampling operation $\zeta_{up}$. 

Most works~\cite{zamir2022restormer, chen2022simple} use Pixel Unshuffle~\cite{pixel_shuffle_unshuffle} for downsampling and Pixel Shuffle~\cite{pixel_shuffle_unshuffle} for upsampling.
However, as discussed in \ref{sec:related}, these operations can introduce artifacts in the feature maps.

Thus in our work, we modify the $\zeta_{down}$ and $\zeta_{up}$ operations in the encoder and decoder parts of the network, respectively, while keeping the encoder and decoder building blocks the same.

Following we formulate these changes,

Let feature $\mathrm{X}_s$ of shape $[\mathrm{B}, \mathrm{C}, \mathrm{M}, \mathrm{N}]$ consist of ${[\mathrm{x}, \mathrm{y}]}_{\mathrm{M}\times\mathrm{N}}$ coordinates in the spatial domain, where $\mathrm{B}$ is the batch size, $\mathrm{C}$ are the number of feature channels, and the spatial resolution of the feature map is $\mathrm{M}\times \mathrm{N}$.

\subsubsection{Downsampling}
\label{subsec:method:downsampling}

A na\"{i}ve implementation of FLC Pooling~\cite{grabinski2022frequencylowcut} for downsampling would not work for pixel-wise tasks such as image restoration due to the inherent nature of a low-pass filter in the Fourier domain.
However, high-frequency information is crucial for high-quality image restoration. Additionally, the application of a rectangle function in the frequency domain generates lattice-like artifacts as shown by \cite{lattice_structure} due to Gibb's phenomenon~\cite{gibbs}.
A low-pass filter is a circular filter in space, so we need padding to effectively avoid the circular convolution which creates artifacts.
Thus, similar to \cite{grabinski2023fix}, we use padding to reduce this kind of artifacts and keep the size of spatial dimensions odd for additional numerical stability.
The padding is only required to be large enough to avoid artifacts from circular convolutions.
Thus, if the spatial resolution of the feature maps before appropriate padding was $[{\mathrm{M}}\times {\mathrm{N}}]$, then after the padding it changes to $[\frac{5}{2}{\mathrm{M}}+1 \times \frac{5}{2}{\mathrm{N}}+1]$,  for simplicity we refer to it as $[\mathcal{M}\times \mathcal{N}]$.

So, first we pad $\mathrm{X}_s$ on all sides such that,

\begin{equation}
\label{eqn:downsampling:padding}
\mathrm{X}_{s_{\mathrm{B}\times \mathrm{C}\times {\mathrm{M}}\times {\mathrm{N}}}} \xrightarrow[]{Padding} \mathrm{X}_{s_{\mathrm{B}\times \mathrm{C}\times \frac{5}{2}\mathrm{M}+1 \times \frac{5}{2}\mathrm{N}+1}} =:\mathrm{X}_{s_{\mathrm{B}\times \mathrm{C}\times {\mathcal{M}}\times {\mathcal{N}}}}
\end{equation}

\iffalse
\begin{equation}
\label{eqn:downsampling:padding}
\mathrm{X}_{s_{\mathrm{B}\times \mathrm{C}\times \bar{\mathrm{M}}\times \bar{\mathrm{N}}}} \xrightarrow[]{Padding} \mathrm{X}_{s_{\mathrm{B}\times \mathrm{C}\times \mathrm{M} \times \mathrm{N}}} %=:\mathrm{X}_{s_{\mathrm{B}\times \mathrm{C}\times {\mathrm{M}}\times {\mathrm{N}}}}
\end{equation}
\fi

Please note, this padding is performed on the feature maps during downsampling, right before the DFT operations, and is removed right after the IDFT operation.
Such that, all operations in the frequency domain are performed on these padded feature maps, while all operations in the spatial domain are performed on unpadded feature maps.

First, we obtain %$\mathrm{X_{\omega}}$ 
$\hat{\mathrm{X}}_{\omega}$ by performing a %2D Discrete Fourier Transform (DFT) and 
2D Discrete Fourier Transform (DFT) followed by a low-pass filter on the padded feature maps $\mathrm{X}_s$ , i.e.,
%
%\begin{equation}
%    \label{eqn:downsampling:DFT}
%     \mathrm{X_{\omega}\left[k, l\right]} = \frac{1}{M\times N} \sum_{m=0}^{M-1} \sum_{n=0}^{N-1} X_s[x_m, y_n] \cdot e^{-j2\pi\left(\frac{k}{M}m+\frac{l}{N}n\right)}
%\end{equation}
%
\begin{equation}
    \label{eqn:downsampling:DFT_low_pass}
    \hat{\mathrm{X}}_{\omega}\left[k,l\right] = \frac{1}{{\mathcal{M}}\times \mathcal{N}} \sum_{m=0}^{{\mathcal{M}}-1} \sum_{n=0}^{\mathcal{N}-1} \mathrm{X}_s[x_m, y_n] \cdot e^{-j2\pi\left(\frac{k}{\mathcal{M}}m+\frac{l}{\mathcal{N}}n\right)}
\end{equation}
for $k,l$ in $0, \dots, \frac{\mathcal{M}-1}{2}$ and $0, \dots, \frac{\mathcal{N}-1}{2}$, respectively.
Now, $\hat{\mathrm{X}}_{\omega_{\mathrm{B}\times \mathrm{C}\times \frac{\mathcal{{M}}}{2} \times \frac{\mathcal{{N}}}{2}}}$ are the low frequency features in $\mathrm{X}_{s_{\mathrm{B}\times \mathrm{C}\times {\mathcal{M}}\times {\mathcal{N}}}}$.

Please note, here \cref{eqn:downsampling:DFT_low_pass} corresponds to \cref{eqn:downsampling:DFT_low_pass} in the main paper.

Then, \textbf{for the low-frequencies path}, we perform a full Inverse Discrete Fourier Transform on $\hat{\mathrm{X}}_\omega$ to convert it to the spatial domain and obtain $\hat{\mathrm{X}}_s$
\begin{equation}
    \label{eqn:downsampling:IDFT_low_pass}
 \hat{\mathrm{X}}_{s}[m, n] = \sum_{k=0}^{\frac{\mathcal{M}-1}{2}}\sum_{l=0}^{\frac{\mathcal{N}-1}{2}} \hat{\mathrm{X}}_{\omega}\left[k,l\right] \cdot e^{j2\pi\left(\frac{k}{{\mathcal{M}}}m+\frac{l}{\mathcal{N}}n\right)},
\end{equation}

To emulate the increase in the number of channels done by Pixel Unshuffle operation on the high-frequencies, we concatenate $\hat{\mathrm{X}}_{s}$ along the $\mathrm{C}$ dimension four times:
\begin{align}
    \label{eqn:downsampling:concat}
    %\begin{split}
&\hat{\mathrm{X}}_{s_{\mathrm{B}\times \mathrm{4C}\times \frac{\mathcal{{M}}}{2} \times \frac{\mathcal{{N}}}{2}}} = \\ &{\hat{\mathrm{X}}_{s_{\mathrm{B}\times \mathrm{C}\times \frac{\mathcal{{M}}}{2} \times \frac{\mathcal{{N}}}{2}}}}\mathbin\Vert{\hat{\mathrm{X}}_{s_{\mathrm{B}\times \mathrm{C}\times \frac{\mathcal{{M}}}{2} \times \frac{\mathcal{{N}}}{2}}}}\mathbin\Vert{\hat{\mathrm{X}}_{s_{\mathrm{B}\times \mathrm{C}\times \frac{\mathcal{{M}}}{2} \times \frac{\mathcal{{N}}}{2}}}}\mathbin\Vert{\hat{\mathrm{X}}_{s_{\mathrm{B}\times \mathrm{C}\times \frac{\mathcal{{M}}}{2} \times \frac{\mathcal{{N}}}{2}}}}.\nonumber
    %\end{split}     
\end{align}

Lastly, we crop out the feature map, to get rid of the padding introduced in the spatial domain in \cref{eqn:downsampling:padding}, before the DFT. %\cref{eqn:downsampling:padding}. 
This completes the alias-free path of the low-frequencies.

Next, \textbf{for the path propagating the high-frequencies},
%As shown by \cref{eqn:downsampling:DFT_low_pass}, we use the median frequency as the threshold for the low-pass filter.
we zero pad $\hat{\mathrm{X}}_{\omega_{\mathrm{B}\times \mathrm{C}\times \frac{\mathcal{{M}}}{2} \times \frac{\mathcal{{N}}}{2}}}$  to obtain 
$\check{\mathrm{X}}_{\omega_{\mathrm{B}\times \mathrm{C}\times \mathcal{{M}}\times \mathcal{{N}}}}$ %which is of shape $[\mathrm{B}, \mathrm{C}, \mathrm{\bar{M}}, \mathrm{\bar{N}}]$
\begin{equation}
    \label{eqn:downsampling:zero_pad_low}
\hat{\mathrm{X}}_{\omega_{\mathrm{B}\times \mathrm{C}\times \frac{\mathcal{{M}}}{2} \times \frac{\mathcal{{N}}}{2}}}  \xrightarrow[]{Padding} \check{\mathrm{X}}_{\omega_{\mathrm{B}\times \mathrm{C}\times \mathcal{{M}}\times \mathcal{{N}}}} % = 0_{\mathrm{B}\times \mathrm{C}\times \mathrm{M}\times \mathrm{N}} + 
\end{equation}

Please note, \cref{eqn:downsampling:DFT_low_pass} and \cref{eqn:downsampling:zero_pad_low} together represent a low-pass filter such that the threshold is the median of the frequencies present.

Next, in the high-frequency path, we perform a full Inverse DFT on $\check{\mathrm{X}}_\omega$ to convert it to the spatial domain and obtain $\check{\mathrm{X}}_s$.
\begin{equation}
    \label{eqn:downsampling:IDFT_zero_padd}
 \check{\mathrm{X}}_{s}[m, n] = \sum_{k=0}^{\mathcal{M}-1}\sum_{l=0}^{\mathcal{N}-1} \check{\mathrm{X}}_{\omega}\left[k,l\right] \cdot e^{j2\pi\left(\frac{k}{\mathcal{M}}m+\frac{l}{\mathcal{N}}n\right)}
\end{equation}
for $k,l$ in $0, \dots, \mathcal{M}-1$ and $0, \dots, \mathcal{N}-1$, respectively.
Here, \cref{eqn:downsampling:IDFT_zero_padd} corresponds to \cref{eqn:downsampling:IDFT_zero_padd} in the main paper. 
%In \cref{eqn:downsampling:IDFT_zero_padd}, $\check{\mathrm{X}}_\omega$ represents a zero padded $\hat{\mathrm{X}}_\omega$ in \cref{eqn:downsampling:IDFT_zero_padd}, padded using \cref{eqn:downsampling:zero_pad_low}.

%
Next, we remove the padding from introduced in \cref{eqn:downsampling:padding}, before the DFT
% \cref{eqn:downsampling:padding} 
from ${\mathrm{X}}_s$ and $\check{\mathrm{X}}_s$, and  
we obtain the features $\mathrm{X'}_{s_{\mathrm{B}\times \mathrm{C}\times \mathrm{{M}}\times \mathrm{{N}}}}$ that correspond to the high frequencies in $\mathrm{X}_{s_{\mathrm{B}\times \mathrm{C}\times \mathrm{M}\times \mathrm{N}}}$ using \cref{eqn:downsampling:high_freq_DoS}
\begin{equation}
 \label{eqn:downsampling:high_freq_DoS}
       \mathrm{X'}_{s_{\mathrm{B}\times \mathrm{C}\times \mathrm{M}\times \mathrm{N}}} = \mathrm{X}_{s_{\mathrm{B}\times \mathrm{C}\times \mathrm{M}\times \mathrm{N}}} -  \check{\mathrm{X}}_{s_{\mathrm{B}\times \mathrm{C}\times \mathrm{M}\times \mathrm{N}}}
\end{equation}
%This corresponds to \cref{eqn:downsampling:high_freq_DoS} in the main paper.
Here, \cref{eqn:downsampling:high_freq_DoS}, represents a high-pass filter in the frequency domain.

Then, we downsample $\mathrm{X'}_{s_{\mathrm{B}\times \mathrm{C}\times \mathrm{M}\times \mathrm{N}}}$ to $\mathrm{X'}_{s_{\mathrm{B}\times \mathrm{4C}\times \mathrm{\frac{M}{2}}\times \mathrm{\frac{N}{2}}}}$ using Pixel Unshuffle. 
For PixelUnshuffle, we know the information being removed, making the operation perfectly invertible.
This property is crucial later when we attempt to mirror the downsampling operation, to perform an upsampling operation.

Now, we formulate the downsampling operations of the features maps, i.e.~$\zeta_{down}$ as shown in  \cref{eqn:downsampling:learnable} in the main paper, and here in \cref{eqn:downsampling:learnable}
\begin{equation}
    \label{eqn:downsampling:learnable}
  \zeta_{down}(\mathrm{X}_{s_{\mathrm{B}\times \mathrm{C}\times \mathrm{M}\times \mathrm{N}}}) = (1 - \alpha) \cdot \hat{\mathrm{X}}_{s_{\mathrm{B}\times \mathrm{4C}\times \mathrm{\frac{M}{2}}\times \mathrm{\frac{N}{2}}}} + \alpha \cdot \mathrm{X'}_{s_{\mathrm{B}\times \mathrm{4C}\times \mathrm{\frac{M}{2}}\times \mathrm{\frac{N}{2}}}}
\end{equation}
where $\alpha$ is a learnable parameter.
We learn a value for alpha such that we mix the features with low and high frequencies optimally.
We initialize $\alpha$=0.3 to have a bias towards low-frequency information.

Thus, using $\zeta_{down}$ we obtain $\mathrm{X}_{s_{\mathrm{B}\times \mathrm{4C}\times \mathrm{\frac{M}{2}}\times \mathrm{\frac{N}{2}}}}$ which is  downsampled $\mathrm{X}_{s_{\mathrm{B}\times \mathrm{C}\times \mathrm{M}\times \mathrm{N}}}$ as shown by \cref{eqn:downsampling:operation}.
\begin{equation}
\label{eqn:downsampling:operation}
    \zeta_{down}(\mathrm{X}_{s_{\mathrm{B}\times \mathrm{C}\times \mathrm{M}\times \mathrm{N}}}) = \mathrm{X}_{s_{\mathrm{B}\times \mathrm{4C}\times \mathrm{\frac{M}{2}}\times \mathrm{\frac{N}{2}}}}
\end{equation}
We name this method \textbf{FrequencyPreservedPooling (FP)}, because it allows the downsampling operation to be fully signal preserving, depending on $\alpha$, while providing an alias-free path.
We represent these operations as a flow diagram in \cref{fig:complete_boa} (left). 
%
%
%
% \hat{\mathrm{X}}_{s_{\mathrm{B}\times \mathrm{C}\times \frac{\mathrm{M}}{2} \times \frac{\mathrm{N}}{2}}}[m, n]
%
For ablation, we also consider the scenario in which we randomly drop the high frequencies with a probability of 30\%, i.e. we use only the low frequencies $\hat{\mathrm{X}}_{\omega_{\mathrm{B}\times \mathrm{4C}\times \frac{\mathrm{M}}{2} \times \frac{\mathrm{N}}{2}}}$ during learning 30\% of the time (batch-wise), in this scenario \cref{eqn:downsampling:learnable} effectively changes to 
\begin{equation}
    \label{eqn:downsampling:only_low}
  \Tilde{\zeta}_{down}(\mathrm{X}_{s_{\mathrm{B}\times \mathrm{C}\times \mathrm{M}\times \mathrm{N}}}) =  \hat{\mathrm{X}}_{s_{\mathrm{B}\times \mathrm{4C}\times \mathrm{\frac{M}{2}}\times \mathrm{\frac{N}{2}}}}.
\end{equation}

%%%%%%%%%%%%%%%%%%%%%%%%%%%%%%%%%%%%%%%%%%%%%  UPSAMPLING       %%%%%%%%%%%%%%%%%%%%%%%%%%%%%%%%%%%%%%%%%%%%%

\subsubsection{Upsampling}
\label{subsec:method:upsampling}

For upsampling the feature maps, we propose an approach symmetric to the downsampling operation. 
Here the objective is to perform upsampling operation $\zeta_{up}$ on $\mathrm{X}_{s_{\mathrm{B}\times \mathrm{C}\times \mathrm{M}\times \mathrm{N}}}$ and obtain $\mathrm{X}_{s_{\mathrm{B}\times \mathrm{\frac{C}{4}}\times \mathrm{2{M}}\times \mathrm{2{N}}}}$.
%This is shown by \cref{eqn:upsampling:operation},
\begin{equation}
    \label{eqn:upsampling:operation}
    \zeta_{up}(\mathrm{X}_{s_{\mathrm{B}\times \mathrm{C}\times \mathrm{M}\times \mathrm{N}}}) = \mathrm{X}_{s_{\mathrm{B}\times \mathrm{\frac{C}{4}}\times \mathrm{2{M}}\times \mathrm{2{N}}}}
\end{equation}
Following we describe the operations that comprise $\zeta_{up}$.

First, we perform a DFT on the feature maps $\mathrm{X}_s$ that exist in the spatial domain to obtain the feature maps in the frequency domain denoted by $\mathrm{X}_{\omega}$ as shown by \cref{eqn:upsampling:DFT}.
\begin{equation}
    \label{eqn:upsampling:DFT}
     \mathrm{X_{\omega}\left[k, l\right]} = \frac{1}{\mathrm{M}\times \mathrm{N}} \sum_{m=0}^{\mathrm{M}-1} \sum_{n=0}^{\mathrm{N}-1} X_s[x_m, y_n] \cdot e^{-j2\pi\left(\frac{k}{\mathrm{M}}m+\frac{l}{\mathrm{N}}n\right)}
\end{equation}
Since, during the downsampling step we concatenated four feature maps at a time, as shown in \cref{eqn:downsampling:concat}, now we calculate $\Bar{\mathrm{X}}_{\omega}$, in which replace a set of four feature maps from $\mathrm{X}_{\omega}$ with average over those four feature maps along the channel dimension. 
This is shown by \cref{eqn:upsampling:avg_four},
\begin{align}
    \label{eqn:upsampling:avg_four}
    &\Bar{\mathrm{X}}_{\omega_{\mathrm{B}\times \mathrm{C}\times \mathrm{M}\times \mathrm{N}}} = \\
    &\frac{1}{4}\left(\sum_{i=0}^{3}\mathrm{X}_{\omega_{\mathrm{B}\times i\times \mathrm{M}\times \mathrm{N}}} \mathbin\Vert \sum_{i=0}^{3}\mathrm{X}_{\omega_{\mathrm{B}\times (4+i)\times \mathrm{M}\times \mathrm{N}}} \mathbin\Vert ... \mathbin\Vert \sum_{i=0}^{3}\mathrm{X}_{\omega_{\mathrm{B}\times (\mathrm{C}-4+i)\times \mathrm{M}\times \mathrm{N}}} \right)\nonumber
\end{align}

Please note ${\mathrm{X}}_{\omega}$ contains both low-frequency and high-frequency information, but since during downsampling, we focus on learning low-frequency information, ${\mathrm{X}}_{\omega}$ contains more low-frequency information and by extension same is true for $\Bar{\mathrm{X}}_{\omega}$.
Thus, effectively resulting in \cref{eqn:upsampling:DFT} and \cref{eqn:upsampling:avg_four} together representing a low-pass filter.

Next \textbf{for the high-frequency information path}, as also shown by \cref{eqn:upsampling:high_freq}, we separate the frequencies higher than the mean to obtain $\mathrm{X^*}_{\omega}$ for each four-tuple of channels
\begin{equation}
    \label{eqn:upsampling:high_freq}
    \mathrm{X^*}_{\omega} = \mathrm{X}_{\omega} - \Bar{\mathrm{X}}_{\omega},
\end{equation}
and perform IDFT on $\mathrm{X^*}_{\omega_{\mathrm{B}\times \mathrm{C}\times \mathrm{M}\times \mathrm{N}}}$ to convert the feature maps to $\mathrm{X^*}_{s_{\mathrm{B}\times \mathrm{C}\times \mathrm{M}\times \mathrm{N}}}$ in the spatial domain as shown by \cref{eqn:upsampling:idft_high_freq},
\begin{equation}
    \label{eqn:upsampling:idft_high_freq}
    \mathrm{X^*}_{s}[m, n] = \sum_{k=0}^{\mathrm{M}-1}\sum_{l=0}^{\mathrm{N}-1} \mathrm{X^*}_{\omega}\left[k,l\right] \cdot e^{j2\pi\left(\frac{k}{\mathrm{M}}m+\frac{l}{\mathrm{N}}n\right)}.   
\end{equation}
This operation is followed by a Pixel Shuffle operation on $\mathrm{X^*}_{s_{\mathrm{B}\times \mathrm{C}\times \mathrm{M}\times \mathrm{N}}}$ to upsample it to $\mathrm{X^*}_{s_{\mathrm{B}\times \mathrm{\frac{C}{4}}\times \mathrm{2M}\times \mathrm{2N}}}$.
This completes the high-frequency path.

Next \textbf{for the low-frequency information path}, from $\Bar{\mathrm{X}}_{{\omega}_{\mathrm{B}\times \mathrm{C}\times \mathrm{M}\times \mathrm{N}}}$, we select the first of every four identical channels and obtain $\Bar{\mathrm{X}}_{{\omega}_{\mathrm{B}\times \mathrm{\frac{C}{4}}\times \mathrm{M}\times \mathrm{N}}}$.

In Pixel Shuffle and Unshuffle, the spatial shuffle is not given. 
Thus, combining the 4 maps in the frequency domain is not trivial as they are shuffled in space.

Then, we are pad with zeros along the x-axis and y-axis of the feature maps in the frequency domain to obtain $\Bar{\mathrm{X}}_{{\omega}_{\mathrm{B}\times \mathrm{\frac{C}{4}}\times \mathrm{2M}\times \mathrm{2N}}}$ to complete the upsampling.
This completes the path for propagating the low-frequency information.

This is followed by an IDFT operation on $\Bar{\mathrm{X}}_{\omega}$ to obtain $\Bar{\mathrm{X}}_{s}$
\begin{align}
    \label{eqn:upsampling:idft_avg_freq}
     \Bar{\mathrm{X}}_{s}[m, n] =    \sum_{k=0}^{2\mathrm{M}-1}\sum_{l=0}^{2\mathrm{N}-1} \Bar{\mathrm{X}}_{\omega}\left[k,l\right] \cdot e^{j2\pi\left(\frac{k}{\mathrm{M}}m+\frac{l}{\mathrm{N}}n\right)}   %\nonumber
\end{align}

Lastly, we learn a parameter $\beta$ to mix the low-frequency information and high-frequency information feature maps optimally. 
Thus, we express $\zeta_{up}$ by \cref{eqn:upsampling:beta_mixing},
\begin{equation}
    \label{eqn:upsampling:beta_mixing}
    \zeta_{up}(\mathrm{X}_s) = (1-\beta)\cdot\Bar{\mathrm{X}}_{s} + \beta\cdot\mathrm{X^*}_{s}
\end{equation}

Here, similar to $\alpha$ in \cref{subsec:method:downsampling}, we initialize $\beta$ with 0.3 to induce a bias towards low-frequency information.
We name this upsampling method as \textbf{FreqAvgUp}. 
We represent these operations as a flow diagram in \cref{fig:complete_boa} (right) titled ``Upsampling''. \\

Additionally for ablation, we explore another upsampling technique that we name \textbf{SplitUp}. 
Here we first, upsample the feature maps $\mathrm{X}_{s_{\mathrm{B}\times \mathrm{C}\times \mathrm{M}\times \mathrm{N}}}$ using Pixel Shuffle to $\mathrm{X}_{s_{\mathrm{B}\times \mathrm{\frac{C}{4}}\times \mathrm{2M}\times \mathrm{2N}}}$.

Then, we convert the feature maps $\mathrm{X}_{s}$ to the frequency domain using DFT and obtain $\mathrm{X}_{\omega}$ similar to as shown in \cref{eqn:upsampling:DFT}.

Next, we split the upsampled feature maps into low-frequencies $\Tilde{\hat{X}}_{\omega}$ by using a low-pass filter with the threshold being the median frequency in the feature maps, and high-frequencies $\Tilde{X}^{*}_{\omega}$ by subtracting the low-frequencies from the entire feature map in the frequency domain.

Then, we convert these split feature maps to the spatial domain to obtain $\Tilde{\hat{X}}_{s}$ (corresponds to low-frequencies) and $\Tilde{{X}^*}_{s}$ (corresponds to high-frequencies).

Lastly, we learn the parameter $\beta$ to learn and mix the low-frequency and high-frequency information in the spatial domain as shown by \cref{eqn:upsampling:split_up}, to get the upsampling operation $\Tilde{\zeta}_{up}$,
\begin{equation}
    \label{eqn:upsampling:split_up}
    \Tilde{\zeta}_{up}(\mathrm{X}_{s_{\mathrm{B}\times \mathrm{C}\times \mathrm{M}\times \mathrm{N}}}) = (1 - \beta) \cdot \Tilde{\hat{X}}_{\omega} + \beta \cdot \Tilde{X}^{*}_{s}
\end{equation}

Inspired by \cite{grabinski2023fix}, to have symmetry in the Fast Fourier Transform (FFT) along the x-axis and y-axis, for every alternate downsampling and upsampling operation, we use the transposed feature maps in the spatial domain, for the FFT operation, and transpose them back to their original dimensions after having finished all operations.

For every alternate downsampling step and alternate upsampling step, we transpose the spatial feature, such that

\begin{equation}
    \label{eqn:downsampling:transpose}
{\mathrm{X}_{s_{\mathrm{B}\times \mathrm{C}\times \mathrm{M} \times \mathrm{N}}}}^T = \mathrm{X}_{s_{\mathrm{B}\times \mathrm{C}\times \mathrm{N} \times \mathrm{M}}}
\end{equation}

\iffalse
\begin{equation}
    \label{eqn:downsampling:transpose}
{\mathrm{X}_{s_{\mathrm{B}\times \mathrm{C}\times \frac{5}{2}\mathrm{M}+1 \times \frac{5}{2}\mathrm{N}+1}}}^T = \mathrm{X}_{s_{\mathrm{B}\times \mathrm{C}\times \frac{5}{2}\mathrm{N}+1 \times \frac{5}{2}\mathrm{M}+1}}
\end{equation}
\fi

leading to interchanging $\mathrm{M}$ and $\mathrm{N}$ and its appropriate variants in the subsequent operations in the frequency domain.
And we transpose back, when converting back from the frequency domain to the spatial domain.
%\sct{Maybe we don't need to mention this transpose here.}

%%%%%%%%%%%%%%%%%%%%%%%%%%%%%%%%%%%%%%%%%%%%%%%%%%%%%%%%%%%%%%%%%%%%           EXPERIMENTS           %%%%%%%%%%%%%%%%%%%%%%%%%%%%%%%%%%%%%%%%
\section{Experiments}
\label{sec:experiments}

\subsection{Experimental Setup}
\label{subsec:experiments:setup}
In the following, we provide the experimental setup for this work.
Please refer to \cref{sec:appendix:exp_setup_details} for additional details.

\paragraph{Downstream Task. }
We focus on image deblurring and deraining. Image deblurring is chosen because preserving high-frequency details is crucial for restoring sharp edges and object boundaries while minimizing spectral artifacts from incorrect sampling. In contrast, image deraining requires removing some high-frequency noise while preserving high-frequency details, presenting challenges different from deblurring. Please refer to \cref{subsec:appendix:results:denoising} for image denoising.
% We focus on the image deblurring task, image denoising, and image deraining. We choose image deblurring as it is a good example of a task in which preserving high-frequency information is essential, especially to restore sharp edges and object boundaries in an image, while attempting to reduce the spectral artifacts infused in the image, due to incorrect sampling.
% Next, we study image denoising and image deraining, as here removing high-frequency noise is essential, and might have different requirements compared to image deblurring.

\iffalse
\paragraph{Dataset. }Similar to \cite{zamir2022restormer, chen2022simple}, for our experiments we use the GoPro image deblurring dataset~\cite{gopro}, and several image deraining datasets. Please refer to \cref{sec:appendix:exp_setup_details} for more details. There, we also report results on the SIDD image denoising dataset.
\fi
%This dataset consists of 3214 real-world images, split into 2103 training images and 1111 test images, with realistic blur and their corresponding ground truth (deblurred images) captured using a high-speed camera.

\paragraph{Evaluation Metrics. }A higher Peak Signal-to-Noise Ratio (PSNR) and Structural similarity (SSIM)~\cite{ssim} indicate a better image quality i.e.~an image closer to the ground truth image. 
These metrics serve as a community-adopted benchmark for these datasets and provide some information regarding the quality of the restorations.

\paragraph{Networks. }For our proposed architectural design changes, we use the SotA image restoration architectures Restormer~\cite{zamir2022restormer}, a Vision Transformer-based network, and NAFNet~\cite{chen2022simple}, a CNN-based network, both have a UNet-like architecture.
Additionally, we compare against FSNet~\cite{cui2024image}, a recently proposed image deblurring model that uses Fourier transforms in their downsampling operations as well.

\paragraph{Adversarial Attacks. }We use PGD~\cite{pgd} and CosPGD~\cite{agnihotri2023cospgd} white-box attacks as they provide a good overview of the quality of the features learned by the network. 
%As proposed by \cite{pgd, agnihotri2023cospgd}, 
For both attacks we use $\epsilon\approx\frac{8}{255}$ and $\alpha$=0.01.
We evaluate against these adversarial attacks over the number of iterations$\in$\{5, 10, 20\}. Please note, 0 attack iterations in figures correspond to no adversarial attack.

\subsection{Frequency Analysis of the Image Restoration tasks}
\label{sec:appendix:freq_analysis}
\begin{figure*}[ht]
\centering
%\scriptsize
    \begin{tabular}{@{}c@{}}
   %BoA-modified Architecture\\
      \includegraphics[width=0.95\linewidth]{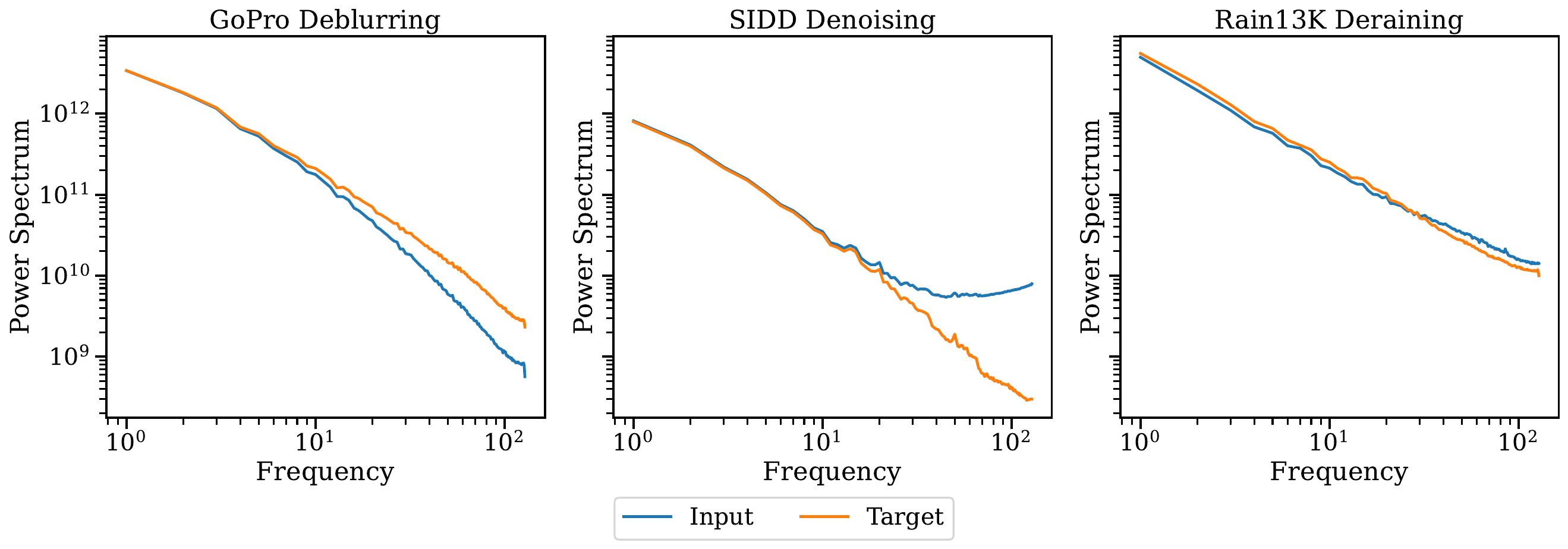}     
    \end{tabular}
\caption{Power v/s Frequency Distribution of the three considered tasks in this work. We form this distribution using 100 random samples from the respective training datasets. ``Input'' denotes the distribution of the input images (to be restored images) used by the restoration network, and ``Target'' denotes the distribution of the ground truth restored images.}
\label{fig:dataset_distribution}   

\end{figure*}

\begin{figure*}[ht]
\centering
%\scriptsize
    \begin{tabular}{@{}c@{}}
   %BoA-modified Architecture\\
      \includegraphics[width=0.95\linewidth]{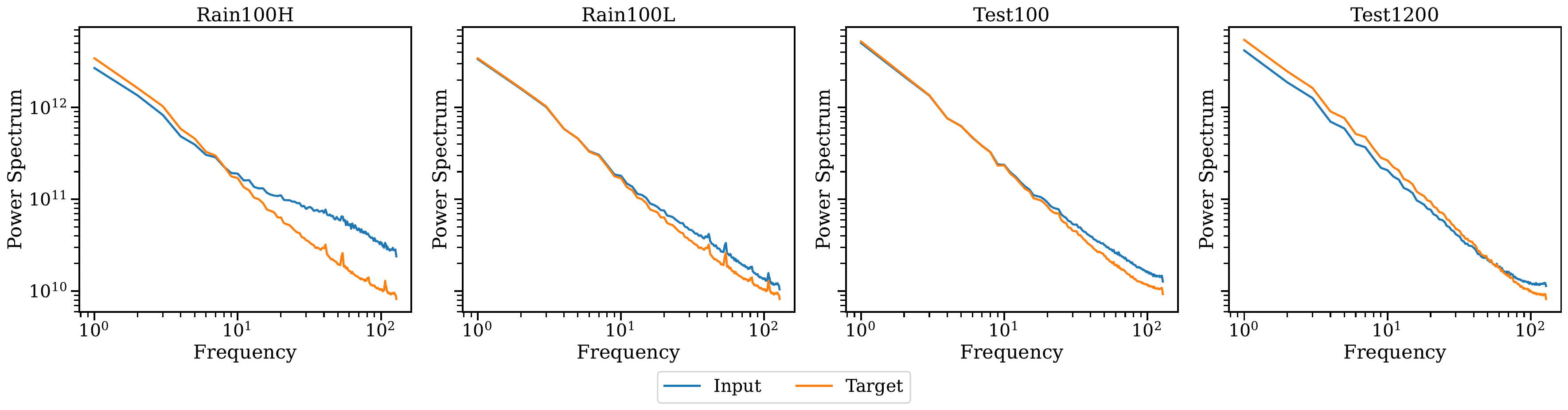}     
    \end{tabular}
\caption{Power v/s Frequency Distribution of the four considered \underline{\textbf{Image Deraining}} datasets in this work. We form this distribution using 100 random samples from the respective test datasets. ``Input'' denotes the distribution of the input images (to be derained images) used by the deraining network, and ``Target'' denotes the distribution of the ground truth restored (derained) images.}
\label{fig:deraining_test_dataset_distribution}   

\end{figure*}
%The three considered tasks of Image Deblurring, Image Denoising and Image Deraining are very different from each other, especially when looked at in the Frequency domain.
The three tasks considered — Image Deblurring, Image Denoising, and Image Deraining — differ significantly from each other, especially when analyzed in the frequency domain.
This is shown in \Cref{fig:dataset_distribution}, where we plot the distribution of the ``Input'' and ``Target'' or Ground-Truth images for each of the training datasets used for the respective tasks.
For image deblurring, we observe that, up until approximately the median frequency, both the input image and the ground-truth image have very similar power spectra. However, after the median frequency, the power and concentration of high-frequencies are lower in the input image than in the ground-truth image.
This is expected, as blurring causes image boundaries and edges to be blurred out, reducing the amount of high-frequencies in the blurred input image.
Thus, for the task of image deblurring, preserving high-frequencies is of paramount importance.

In contrast, the observations made for image deblurring do not hold for image denoising.
The frequency spectra of image denoising are opposite to image deblurring.
Here, until the median frequency, both the noisy input image and the denoised ground-truth image have very similar frequency spectra. However, after the median frequency, the power spectrum of the noisy input images is significantly higher than the ground-truth images.
This indicates that the noisy input images contain high-frequency noise which is to be removed.
Therefore, preserving high frequencies in the noisy input image can negatively impact the restoration network's performance in image denoising.
%Thus, for the task of image denoising, preserving the high-frequencies in the noisy input image can harm the restoration network's performance.

On the other hand, image deraining serves as the middle ground between image deblurring and image denoising.
Across the frequency spectra, the amplitude of the frequencies in the rainy input image and the derained ground-truth image are very similar.
Until the median frequency, the rainy input image has low-frequencies at a marginally lower amplitude and after the median frequency, this trend reverses. 
Thus, for the task of image deraining, preserving high-frequencies is important, however, in some scenarios when the rainy input image has a lot more high-frequency than the ground-truth, losing some high-frequency noise could be helpful, for example, Rain100H as seen in \Cref{fig:deraining_test_dataset_distribution}.
\Cref{fig:deraining_test_dataset_distribution} also helps us better understand the behaviour of different architectural design choices in \Cref{fig:deraining_cospgd_full} and \Cref{fig:deraining_pgd_full}.

\subsection{Image Deraining}
\label{subsec:results:deraining}
\begin{figure}[ht]
    \centering % <-- added
   \includegraphics[width=0.95\linewidth]{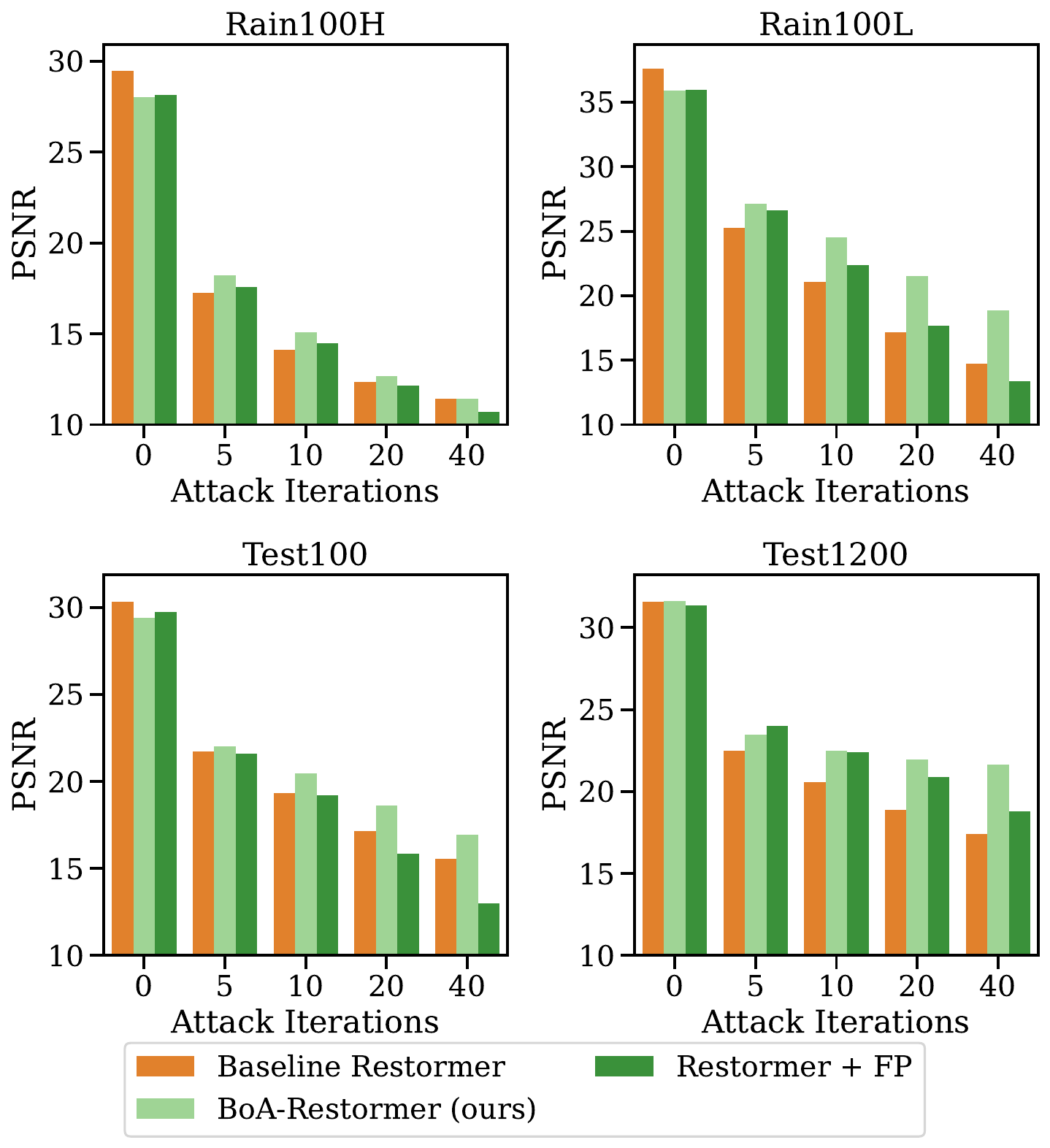}
\caption{\underline{\textbf{Image deraining}} results under CosPGD attack. Titles indicate the dataset set. Please refer to \cref{fig:deraining_cospgd_full} and \cref{fig:deraining_pgd_full} for more results.}
\label{fig:deraining_cospgd}
\end{figure}
\begin{figure}[t]
    \centering % <-- added
\scalebox{0.92}{
   \begin{tabular}{@{}c@{\hspace{0.1cm}}c@{\hspace{0.1cm}}c@{}}
 & NO ATTACK & CosPGD 20 iterations\\

  \rotatebox{90}{\textbf{\small \phantom{b}Baseline Restormer}} & 
  \includegraphics[height=2.7cm, width=0.24\textwidth]{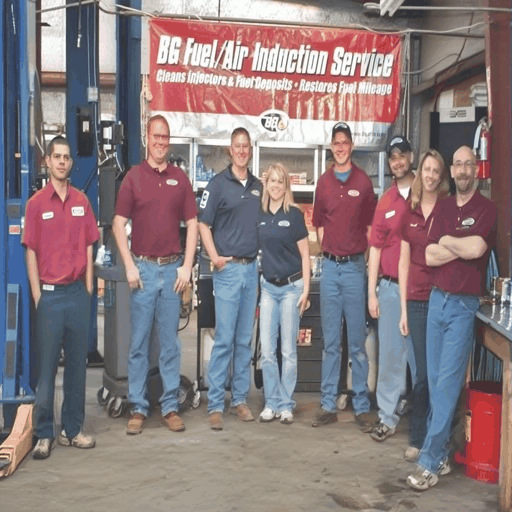}&
  \includegraphics[height=2.7cm, width=0.24\textwidth]{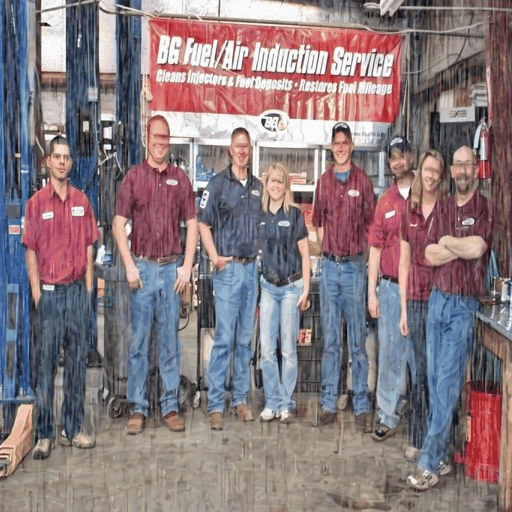}

\\

  \rotatebox{90}{\textbf{\small \phantom{bb}Restormer + FP}} &  
  \includegraphics[height=2.7cm, width=0.24\textwidth]{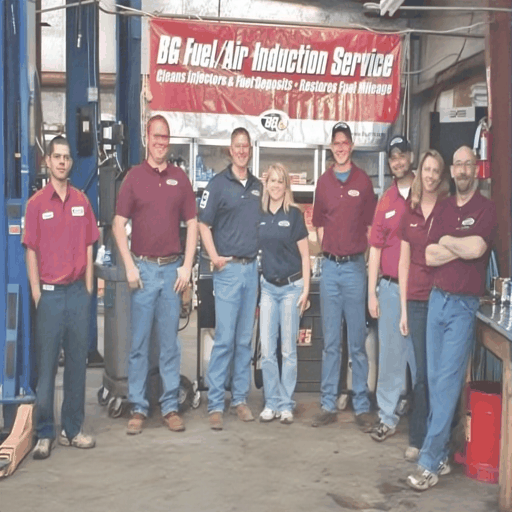}&
  \includegraphics[height=2.7cm, width=0.24\textwidth]{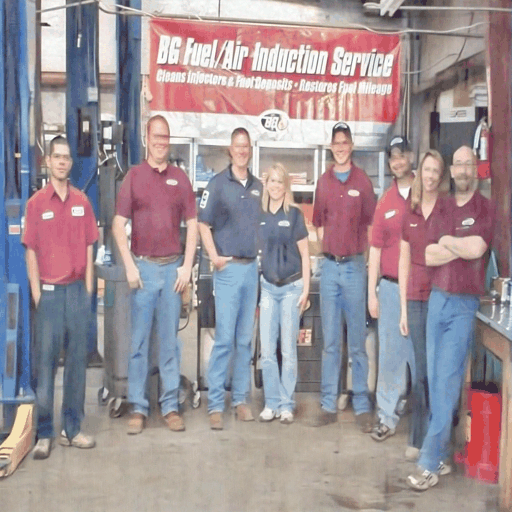}

  \\

  \rotatebox{90}{\textbf{\small \phantom{}BoA-Restormer (ours)}} &
  \includegraphics[height=2.7cm, width=0.24\textwidth]{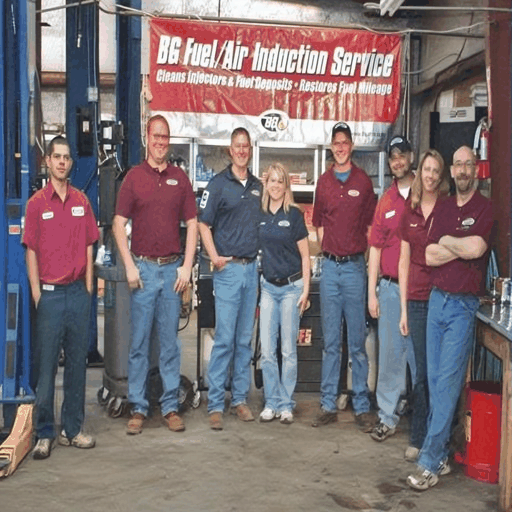}&
  \includegraphics[height=2.7cm, width=0.24\textwidth]{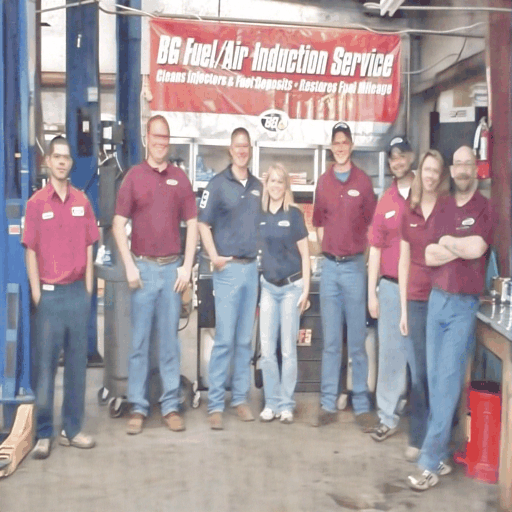}
  \\
\end{tabular}
}
\caption{\underline{\textbf{Image deraining}} restored images from the Test1200 dataset. Comparing sampling methods on rainy input images and CosPGD attacked rain images.}
\label{fig:deraining_images_restormer_cospgd}
\end{figure}
% As a task, image deraining is significantly different from the previously studied two image restoration tasks.
% Here, the high-frequency noise is large enough to disrupt texture and shape information in the input images.
% Thus, while noise is significant enough to adversely affect the performance of a DNN that preserves high-frequency information, preserving this information helps the DNN better restore shapes and certain textures.
As shown in \Cref{sec:appendix:freq_analysis}, image deraining differs significantly from the other image restoration tasks. The high-frequency noise in this case is strong enough to disrupt texture and shape information. While this noise can hinder a DNN that preserves high-frequency details, maintaining this information is crucial for better restoration of shapes and textures.
This can be observed in \cref{fig:deraining_cospgd}, as the adversarial noise is absent or less, BoA performs at par or slightly worse than the baseline or FP.
Yet, as soon as the adversarial noise is optimized by using more steps, BoA outperforms the other two, indicating better learning of feature representations.

This is observed in \cref{fig:deraining_images_restormer_cospgd}, when under no attack, all three considered models are performing almost at par. 
However, under adversarial attack, Restormer is not able to sufficiently de-rain the image, leading to visible rain droplets in the restored image.
\textbf{Restormer + FP} is able to de-rain the image, however, the image quality is not optimal.
Only \textbf{BoA-Restormer} is able to de-rain the image while maintaining optimal image quality, demonstrating the benefits of BoA modifications.

\begin{figure}[t]
    \centering % <-- added
\scalebox{0.92}{
   \begin{tabular}{@{}c@{\hspace{0.1cm}}c@{\hspace{0.1cm}}c@{}}
 & NO ATTACK & CosPGD 20 iterations\\

  \rotatebox{90}{\textbf{\small  \phantom{bbb}Baseline}}  \rotatebox{90}{\small \textbf{\phantom{bb}Restormer}} & 
  \includegraphics[width=0.23\textwidth]{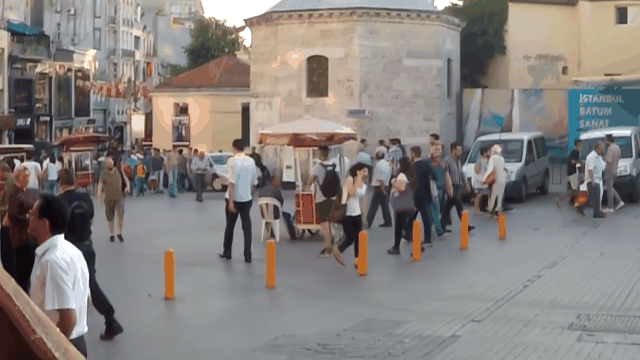}&
  \includegraphics[width=0.23\textwidth]{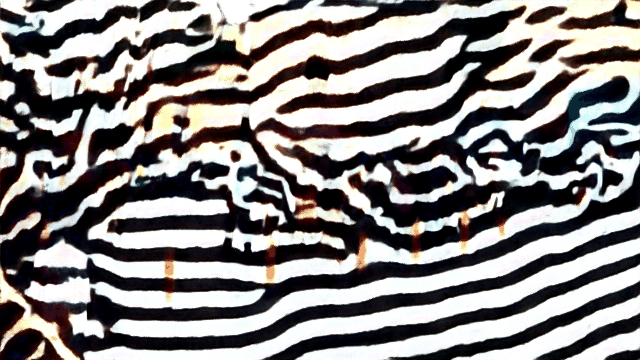}
\\

  \rotatebox{90}{\textbf{\small Restormer + FLC}} & 
  \includegraphics[width=0.23\textwidth]{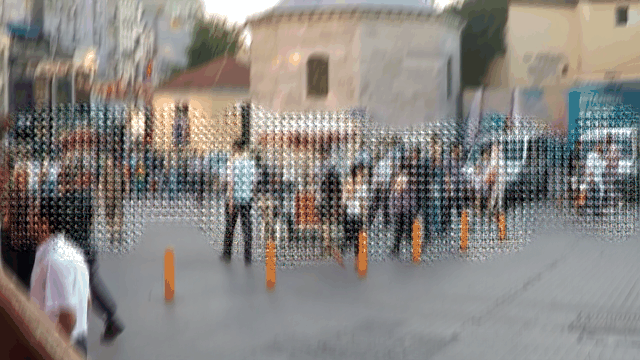}&
  \includegraphics[width=0.23\textwidth]{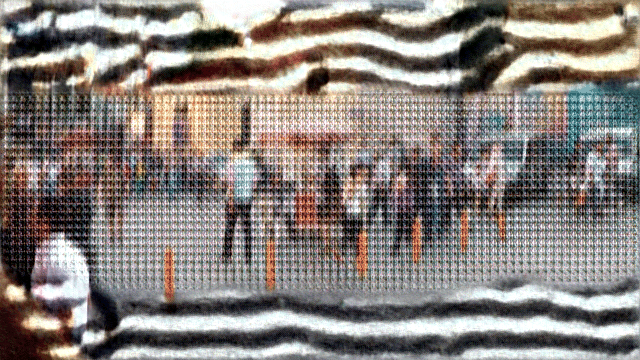}
\\
  \rotatebox{90}{\textbf{\small Restormer + FP}} &  
  \includegraphics[width=0.23\textwidth]{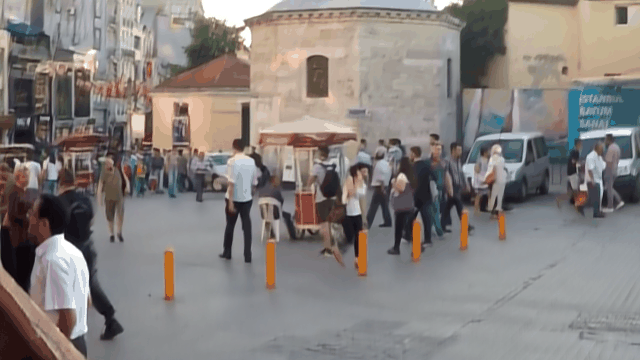}&
  \includegraphics[width=0.23\textwidth]{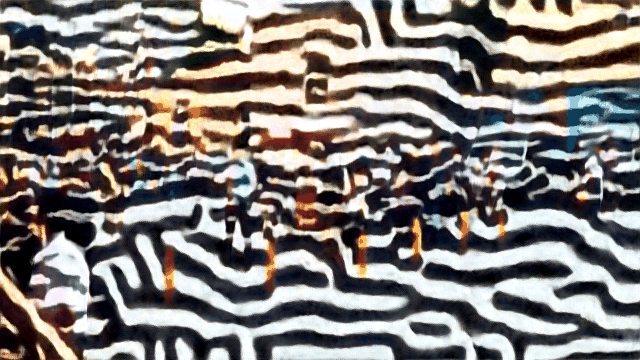}
  \\

  \rotatebox{90}{\textbf{\small Restormer + FP}}  \rotatebox{90}{\small + DropHigh} &
  \includegraphics[width=0.23\textwidth]{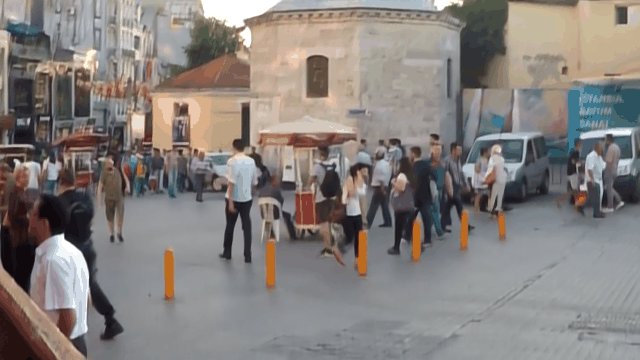}&
  \includegraphics[width=0.23\textwidth]{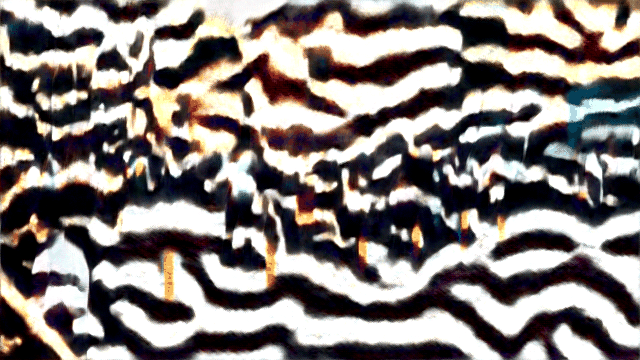}
  \\

  \rotatebox{90}{\textbf{\small Restormer + FP}}  \rotatebox{90}{\tiny + FirstLayerDrop} &
  \includegraphics[width=0.23\textwidth]{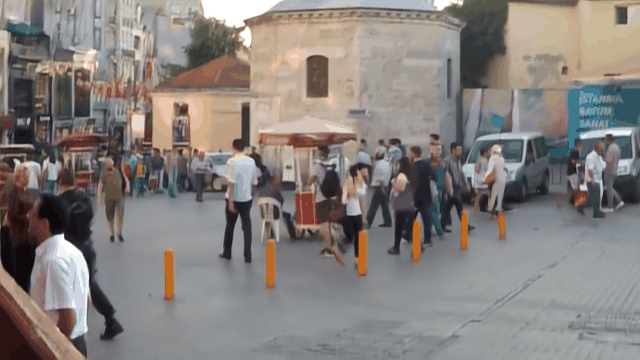}&
  \includegraphics[width=0.23\textwidth]{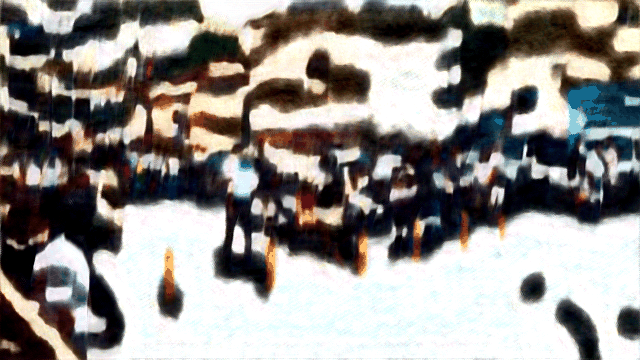}
  \\

  \rotatebox{90}{\textbf{\small BoA-Restormer}}  \rotatebox{90}{\small \textbf{\phantom{bbbb}(Ours)}} &
  \includegraphics[width=0.23\textwidth]{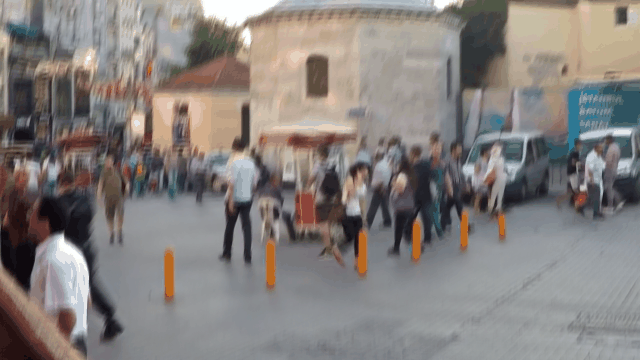}&
  \includegraphics[width=0.23\textwidth]{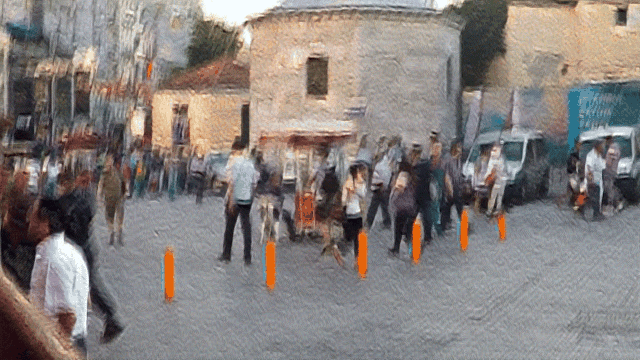}
  \\
\end{tabular}
}
\caption{\underline{\textbf{Image deblurring}} restored images from the GoPro dataset. Qualitatively comparing sampling methods on clean blurry input images and CosPGD attacked input images. Symbolic notations are the same as in \cref{tab:restormer_encoder_ablation}.}
\label{fig:restormer_cospgd_attack_different_downnsamplings5}
\end{figure}

\subsection{Image Deblurring}
\label{subsec:experiments:results}

\begin{figure*}[t]
    \centering % <-- added
   \includegraphics[width=\linewidth]{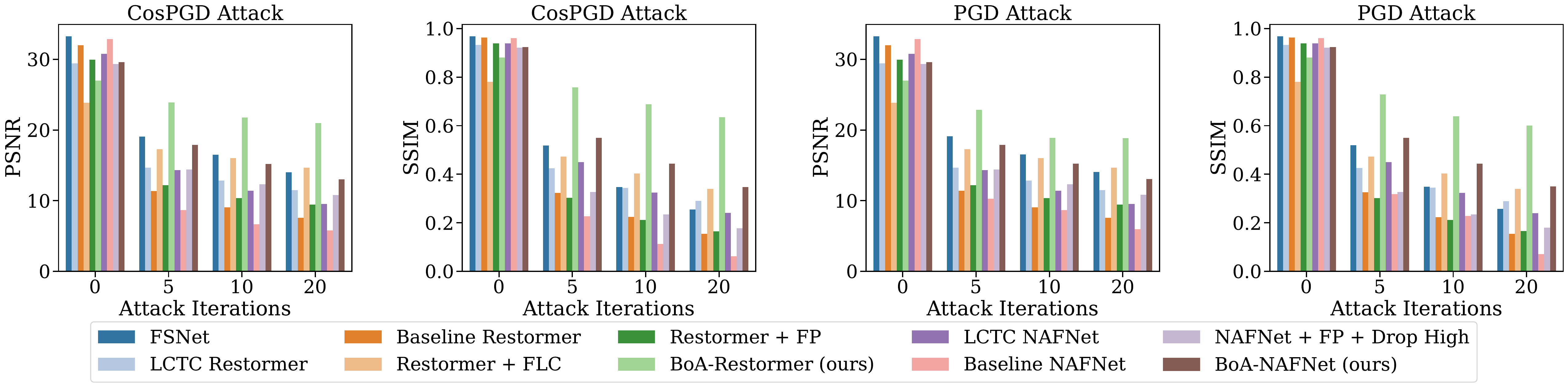}
\caption{\underline{\textbf{Image deblurring}} results under adversarial attack using the GoPro dataset.}
\label{fig:deblurring_main}
\end{figure*}

In the following, we discuss the experimental results from using our proposed \textbf{BoA-Restormer}, employing our downsampling method \textbf{FrequencyPreservedPooling} and our upsampling method \textbf{FreqAvgUp} for image restoration. 
We report additional quantitative results in \cref{subsec:appendix:results} and provide additional qualitative results in \cref{sec:appendix:pgd_images}.
We provide additional ablations on the design choices in \cref{subsec:appendix:proof_low_frequencies_focus} and \cref{subsec:analysis:symmetry}.

We observe in \cref{fig:teaser}, that while both Restormer and FSNet restore blurry images reasonably well, under adversarial attack Restormer's restorations have significant visual artifacts.
These visual artifacts, while reduced, also exist in the images restored by FSNet and are accentuated under strong adversarial attack ($>$10 attack iterations).
We first ablate over na\"{i}vely implementing FLC Pooling~\cite{grabinski2022frequencylowcut} in the Restormer encoder for downsampling. While this reduces spectral artifacts under attack compared to the baseline Restormer, the restored images remain suboptimal and significantly worse than those from FSNet. % and we observe that while under attack, the spectral artifacts that exist in baseline Restormer have reduced slightly, but the restored images are far from ideal and are significantly worse compared to FSNEt.
% Additionally, due to the sinc-interpolation artifacts in FLC Pooling, we observe additional spectral artifacts in a window in the center of the images restored when using FLC Pooling for downsampling.
% These artifacts are called lattice artifacts by \cite{lattice_structure} and occur due to the inherent nature of low-pass filters in the Fourier domain as shown by \cite{gibbs}.
Moreover, the images restored are still significantly blurry.
We hypothesize that this is due to the removal of high-frequency information in the images, which hampers the image restoration model in restoring sharp edges and image boundaries.
Thus, we replace the FLC Pooling in the downsampling with \textbf{FrequencyPreservedPooling (FP)}, and observe that the spectral artifacts that existed in the restored blurry images have been removed (\cref{fig:restormer_cospgd_attack_different_downnsamplings5}). However, the blurry images restored under adversarial attacks still have significant spectral artifacts.
Since the Restormer architecture, Restormer with FLC Pooling for downsampling and Restormer with FrequencyPreservedPooling for downsampling all use PixelShuffle for upsampling, we hypothesize that the decoder is unable to interpret the information encoded during downsampling correctly.
Thus, we introduce symmetry in the downsampling-upsampling operation by replacing the PixelShuffle in \textbf{Restormer+FP} with \textbf{FreqAvgUp} for upsampling to obtain \textbf{BoA-Restormer}. 
This symmetry helps the deblurring model perform significantly better, and we observe that the deblurred images are relatively artifact-free even under strong adversarial attacks.
We discuss importance of this symmetry in detail in \cref{subsec:analysis:symmetry}

From a quantitative perspective, in \cref{fig:deblurring_main}, we observe that the metrics reflect the observations from \cref{fig:restormer_cospgd_attack_different_downnsamplings5}.
While FSNet performs significantly better for blurry images without attack, under adversarial attacks, it lacks stability in its restorations.
A similar observation is made for the baseline Restormer and baseline NAFNet, which performs worse under adversarial attacks.
This indicates that all these models are merely learning shortcuts from the input to the target distribution and not reliable features, leading to spectral artifacts being formed in restored images.
Including FLC Pooling for downsampling, does help increase the adversarial robustness of the models. 
However, the quality of the blurry images restored when not attacked is severely lower with previously discussed spectral artifacts visible.
Replacing FLC Pooling with FrequencyPreservedPooling, helps in increasing the performance of the models on blurry images when not attacked. 
However, it is less robust than using FLC Pooling.
Lastly, we introduce symmetry in Restormer+FP and NAFNet+FP, with BoA modifications, by replacing the upsampling operation from PixelShuffle to FreqAvgUp and observe an increase in the adversarial robustness with the resulting BoA-Restormer, and BoA-NAFNet respectively. 
We observe a slight trade-off in the adversarial and non-adversarial performance, however, this trade-off is expected as shown by \cite{zhang2019theoretically,tsipras2018robustness}.

\iffalse
\begin{itemize}
    \item Restormer performance bad under attack
    \item FLC Pooling increases some robustness but huge drop in clean, visible artifacts, write reason of these artifacts
    \item Only using Frequency Pooling during downsampling results good. More robust than Restormer, but less robust than FLC
    \item Finally, using FreqAvgUp in the upsampling, giving symmetry (refer \cref{subsec:analysis:symmetry}), makes results better. Hopefully, also compare with FSNet. 
\end{itemize}
\fi

\subsection{Analysing Focus on Low-Frequencies}
\label{subsec:analysis:focus}
%Here we talk about the case in which we drop high frequencies.
\begin{table}[t]
    \centering
    \caption{\underline{\textbf{Image deblurring}} ablations on the GoPro dataset. Different downsampling techniques, as discussed in \cref{subsec:method:downsampling}, are compared based on performance on clean samples and against 20 iterations of CosPGD attack. Refer \cref{tab:restormer_encoder_ablation_full} for additional results.}
    %\scriptsize
    \scalebox{0.9}{
    \begin{tabular}{@{}l@{\,}cc|cc@{}}
    \toprule
    \multirow{2}{*}{Architecture} & \multicolumn{2}{c|}{Test} & \multicolumn{2}{c}{CosPGD} \\
   %& & &  \multicolumn{2}{c}{20 attack itrs } \\
   & PSNR & SSIM & PSNR & SSIM  \\
    \toprule

        \multicolumn{5}{c}{Restormer variants} \\        
        \midrule
         \textbf{Restormer} & \textbf{31.99} & \textbf{0.9635} &  7.59 & 0.1548  \\

          ~~~~~ + FLC & 23.85 & 0.7811 & \textbf{14.66} & \textbf{0.3401} \\

          ~~~~~ + FP & 29.95 & 0.9395 & 9.44 & 0.1651 \\
         
          ~~~~~ + FP + DropHigh & 29.93 & 0.9395  & 9.25 & 0.2355 \\

          ~~~~~ + FP + FirstLayerDrop & 29.96 & 0.9402 & 9.75 & 0.2899 \\

          \midrule 
        \multicolumn{5}{c}{NAFNet variants} \\        
        \midrule

         \textbf{NAFNet} & \textbf{32.87} & \textbf{0.9606}& 5.81 & 0.0617  \\

          ~~~~~ + FP & 31.17 & 0.9439 & 4.89 & 0.0001  \\
         
          ~~~~~ + FP + DropHigh & 29.37 & 0.9204 & \textbf{10.78} & \textbf{0.1770}  \\

          ~~~~~ + FP + FirstLayerDrop & 30.85 & 0.9408  & 7.83 & 0.1632 \\

    \bottomrule
    \end{tabular}    
    }
    \label{tab:restormer_encoder_ablation}
\end{table}
In the following, we ablate design decisions to encourage the focus on low-frequency information, while providing both high and low frequencies to the model.
As discussed in \cref{subsec:experiments:results}, we observe that na\"{i}vely implementing FLC Pooling for a pixel-wise task such as image deblurring does not result in usable restored images.
The better performance of \textbf{FrequencyPreservedPooling} over FLC Pooling for image deblurring indicates that retaining high-frequency information from the feature maps is crucial to the image deblurring task.
However, the extent to which this high-frequency information needs to be retained needs further exploration.
Given that CNN-based architectures have a bias towards high-frequency information~\cite{lukasik2023improving,abello2021dissecting}, NAFNet's performance might vary to that of Restormer.
Thus, we study \textbf{FrequencyPreservedPooling} in more detail.
To this effect, we modify \textbf{FrequencyPreservedPooling}, such that for 30\% of the function calls, we drop the high-frequency information, to effectively reduce \textbf{FrequencyPreservedPooling} to a low-pass filter similar to \cite{grabinski2022frequencylowcut} for these calls.
We implement this ablation in two possible settings, first: allowing all downsampling steps to drop high frequencies in 30\% of the calls. We denote this as \textbf{DropHigh}. Second, we allow only the first downsampling step to drop high frequencies 30\% of the time. We denote this as \textbf{FirstLayerDrop}.
The intuition behind \textbf{FirstLayerDrop} is that, since the image itself might have noise that might correspond to high frequencies, reducing them to only the first downsampling step might be enough.
Further reducing high-frequencies at random in all downsampling steps might deprive the model of some important information as well.
We observe in \cref{tab:restormer_encoder_ablation}, that this intuition is validated and indeed \textbf{FP + FirstLayerDrop} outperforms the other ablations over \textbf{FP} in the case of Restormer.
However, when viewed qualitatively in \cref{fig:restormer_cospgd_attack_different_downnsamplings5}, we observe a massive loss of information in the restored images, with considerable areas simply being white.
%Thus, for the ablations with the upsampling using Restormer, we continue using \textbf{FrequencyPreservedPooling} to form BoA-Restormer.
In the case of NAFNet, due to the implicit high-frequency bias of CNNs, we observe that \textbf{DropHigh} performs the best for image deblurring under adversarial attack.
%Thus, we continue using this setting for NAFNet for further ablations with upsampling to form BoA-NAFNet.

%\input{deraining_restormer}
%Please refer \cref{table:deraining}.

\section{Conclusion}
\label{sec:conclusion}
With the increased use of machine learning in computer vision, fundamental principles of signal processing are often overlooked.
These foundations %help guide computer vision tasks and can 
should be utilized by ML methods to ensure that the model is learning meaningful representations of the data.
For pixel-wise tasks like image restoration, disregarding these foundations leads to 
%Fortunately, for pixel-wise tasks like image restoration, this can be validated by the human eye by spotting 
spectral artifacts in the images restored. Additionally, 
%For some scenarios, one might have to use 
adversarial attacks can be used as a tool to accentuate these spectral artifacts which otherwise might not be visible to the human eye, thus helping in their study. % to be visible.
%Some works\cite{agnihotri2023unreasonable} propose using adversarial training as a tool to improve the stability of the features learned by a model, however, adversarial training is computationally and time-wise extremely expensive.

In this work, we focus on the downsampling and upsampling operations being used by some SotA architectures for image deblurring and propose more stable solutions.
We propose \textbf{FrequencyPreservedPooling}, a novel downsampling method, and \textbf{FreqAvgUp} a novel upsampling method, together called \textbf{BoA}-modifications, which are theoretically motivated sampling operations that lead to more stable feature representation being learned by the model without the need for adversarial training. 
These operations preserve important parts of the signal while discarding noise and significantly reducing spectral artifacts in the restored images. 

\paragraph{Limitations and Future Work. }
%While, this work shows that focusing on low-frequency information, while still retaining some high-frequency information, helps the image deblurring model to restore images with significantly reduced spectral artifacts, even under adversarial attack. Yet, the performance on blurry images when not adversarially attacked is not ideal.
This work demonstrates that focusing on low-frequency information while retaining some high-frequency details helps reduce spectral artifacts in image deblurring, even under adversarial attacks. 
However, performance on non-attacked blurry images is not ideal.
%Here, a point needs to be made for the currently used evaluation metrics that do not provide a reliable evaluation of image restoration methods, especially as they ignore spectral artifacts that are visible to the human eye.
It is important to note that current evaluation metrics may not reliably assess image restoration methods, as they often overlook spectral artifacts visible to the human eye.
This can be observed in \cref{fig:teaser}, here FLC Pooling visibly generates significantly more artifacts than FSNet, however, both achieve very similar PSNR values.
%Still, these metrics do serve as a community-adopted benchmark for these datasets and provide some information regarding the quality of the restorations.
%Thus, improving our proposed methods, to perform better on these metrics like PSNR would be additionally helpful to the community.

\section*{Ethics and Broader Impact Statement}
To the best of our knowledge, all literature used in this work has been referenced correctly.
Our work did not involve any human subjects and does not pose a threat to humans or the environment.

Assessing the quality of representations learned by a machine learning model is of paramount importance.
This makes sure that the model is not learning shortcuts from the input distribution to the target distribution~\cite{shortcut} but learning something meaningful.
Adversarial attacks are a reliable tool for gauging the quality of a model's learned representations. 
However adversarial attacks are time and computation exhaustive, especially if used during training.
Thus, our proposed architectural design choices are aimed towards increasing a model's robustness and reducing its vulnerabilities with much resource requirement by avoiding adversarial training and is theoretically motivated. 
Thus, our work helps advance the field of machine learning.

\section*{Acknowledgements}
We acknowledge support by the DFG research unit 5336 - Learning to Sense.
The authors acknowledge support by the state of Baden-Württemberg through bwHPC
and the German Research Foundation (DFG) through grant INST 35/1597-1 FUGG.

\newpage

%\begin{thebibliography}{1}
\bibliographystyle{IEEEtran}
\bibliography{aaai25}
%\end{thebibliography}

\begin{IEEEbiography}[{\includegraphics[width=1in,height=1.25in,clip,keepaspectratio]{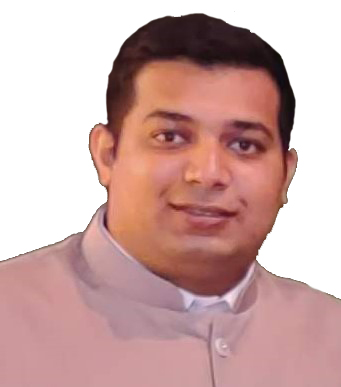}}]{Shashank Agnihotri}
is currently a PhD student in Prof. Dr.-Ing Margret Keuper’s Machine Learning Lab in the DWS Group at the University of Mannheim. He received his MSc. Computer Science from the University of Freiburg in September 2021. He works primarily on analyzing and improving the adversarial and OOD robustness of deep learning architectures. He has published some papers at prestigious venues like ICML, ECCV, NeurIPS, ICCV, and others.\end{IEEEbiography}

\begin{IEEEbiography}[{\includegraphics[width=1in,height=1.25in,clip,keepaspectratio]{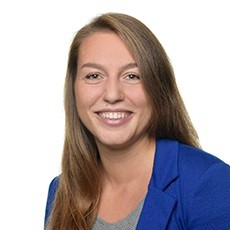}}]{Julia Grabinski} is a PhD candidate at the Fraunhofer Institute for Industrial Mathematics and in the Chair of Machine Learning at the University of Mannheim. 
She received her M.Sc. degree from the University of Mannheim, Germany, in 2021. 
Her research interests include robustness in computer vision under the lens of signal processing. She has published some papers at prestigious venues like NeurIPS, ICCV, ECCV, CVPR, ICML and others. \end{IEEEbiography}

\begin{IEEEbiography}[{\includegraphics[width=1in,height=1.25in,clip,keepaspectratio]{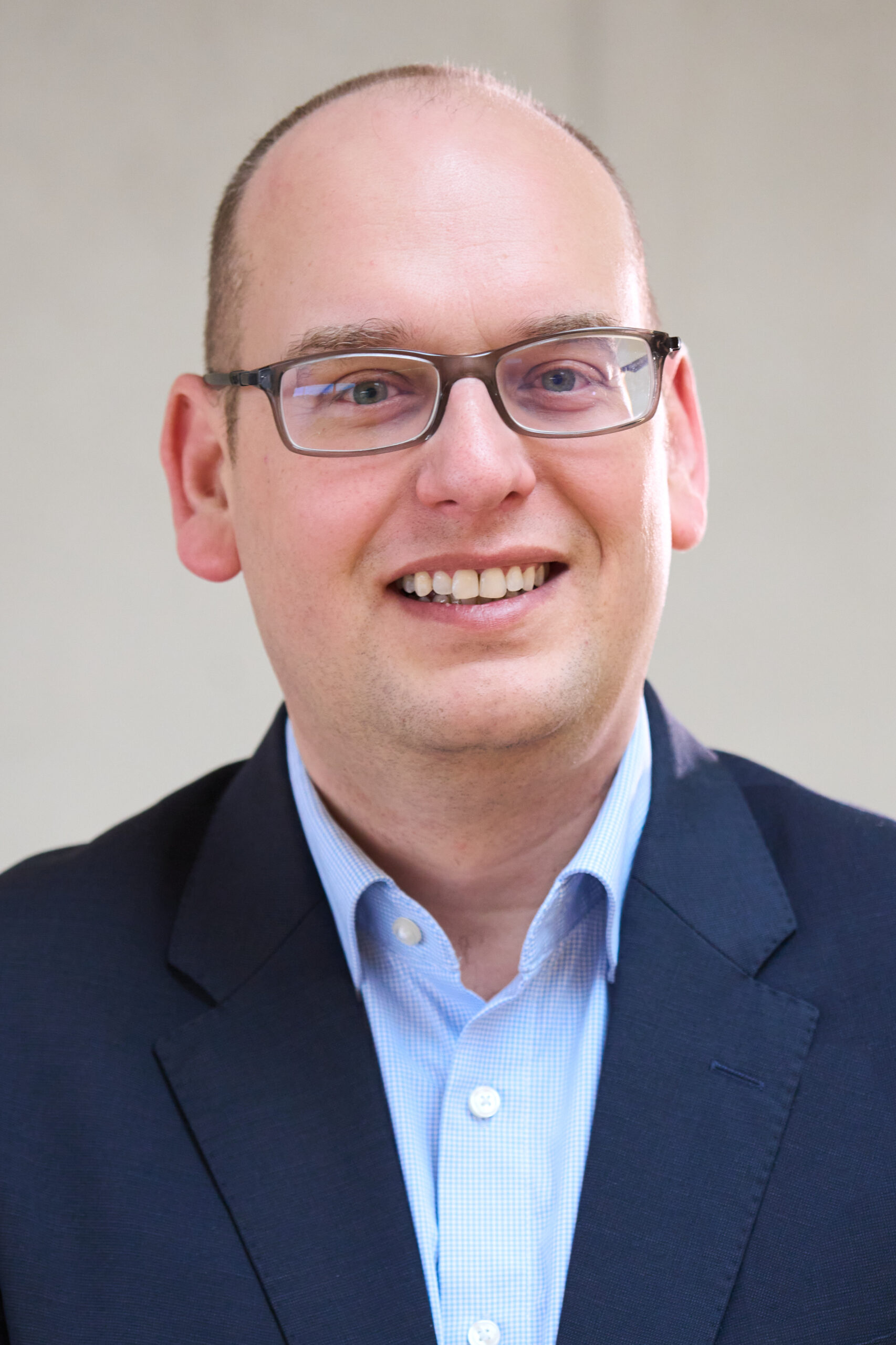}}]{Janis Keuper}
received the PhD degree from the University of Freiburg.
He is a full professor for data science and head of the Institute for Machine Learning and Analytics (IMLA) at Offenburg University and an associated Professor at the University of Mannheim. His research focuses on the correctness and application of learning models in the context of physical systems.\end{IEEEbiography}

\begin{IEEEbiography}[{\includegraphics[width=1in,height=1.25in,clip,keepaspectratio]{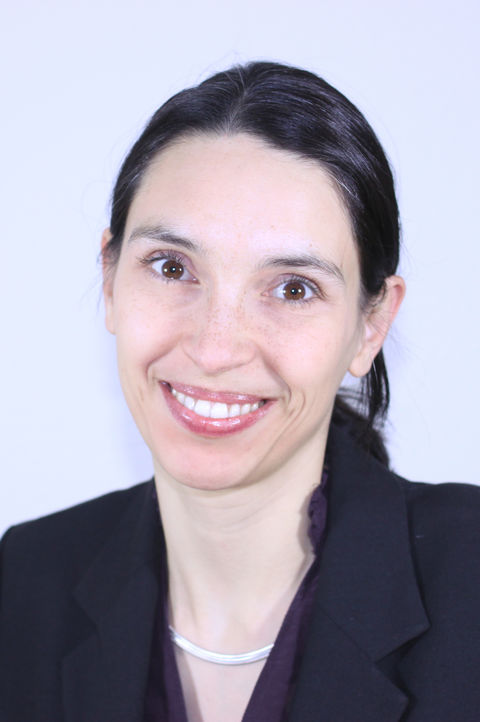}}]{Margret Keuper}
is a full professor for Machine Learning with a focus on computer vision at the University of Mannheim. She is also an affiliated research leader at the Max Planck Institute for Informatics, Saarbruecken. She received her PhD degree from the University of Freiburg under the supervision of Prof Thomas Brox with her thesis entitled ”Segmentation of Cells and Sub-cellular Structures from Microscopic Recordings”. She worked as a postdoctoral researcher at the University of Freiburg working on topics related to motion estimation, segmentation, and grouping. She is also a member of ELLIS.\end{IEEEbiography}

\appendix
\onecolumn
%\section*{}
{
    \centering
    \Large
    \textbf{Beware of Aliases – Signal Preservation is Crucial for Robust Image Restoration} \\
    \vspace{0.5em}Supplementary Material \\
    \vspace{1.0em}
}

\section*{Table of Content}
For better readability of the appendix, following we provide a table of contents:
\begin{itemize}
     \item \textbf{\Cref{sec:appendix:pixel_shuffle_unshuffle}}: We provide a brief overview of the PixelShuffle and PixelUnshuffle Operations and relate it to our work.
    %\item \textbf{\Cref{sec:appendix:method}}: We provide detailed steps for our proposed Downsampling (\cref{subsec:appendix:method:downsampling}) and our proposed Upsampling (\cref{subsec:appendix:method:upsampling}) operations.
    \item \textbf{\Cref{subsec:appendix:results}}: We provide additional quantitative results for:
    \begin{itemize}
        \item \Cref{subsec:appendix:results:deblurring}: Image Deblurring
        \item \Cref{subsec:appendix:results:denoising}: Image Denoising
        \item \Cref{subsec:appendix:results:denoising}: Image Deraining
    \end{itemize}
    \item \textbf{\Cref{sec:appendix:pgd_images}}: We provide additional qualitative results.
    \item \textbf{\Cref{subsec:appendix:proof_low_frequencies_focus}}: We provide a deep-dive study in the learned parameter values showcasing that the network learned to focus of low-frequencies.
    \item \textbf{\Cref{subsec:analysis:symmetry}}: We provide an ablation study for our proposed upsampling method and its importance.
    \item \textbf{\Cref{sec:appendix:exp_setup_details}}: We provide in-depth experimental setup details.
    %\item \textbf{\Cref{sec:appendix:freq_analysis}}: We provide an in-depth frequency analysis of the image restoration tasks and the datasets used to highlight the differences between each task and dataset in the frequency domain.
    \item \textbf{\Cref{sec:appendix:code}}: We provide the code for our proposed prominent sampling operations:
    \begin{itemize}
        \item \Cref{subsec:appendix:code:fp_code}: Code for \textbf{FrequencyPreservedPooling (FP)}.
        \item \Cref{subsec:appendix:code:fp_drop_high_code}: Code for \textbf{FP + DropHigh}.
        \item \Cref{subsec:appendix:code:FreqAvgUp}: Code for \textbf{FreqAvgUp}.
    \end{itemize}
\end{itemize}

\subsection{Background on Pixel Shuffle and PixelUnshuffle}
\label{sec:appendix:pixel_shuffle_unshuffle}
\begin{figure*}[ht]
\centering
%\scriptsize
    \begin{tabular}{@{}c@{}}
   %BoA-modified Architecture\\
      \includegraphics[width=0.7\linewidth]{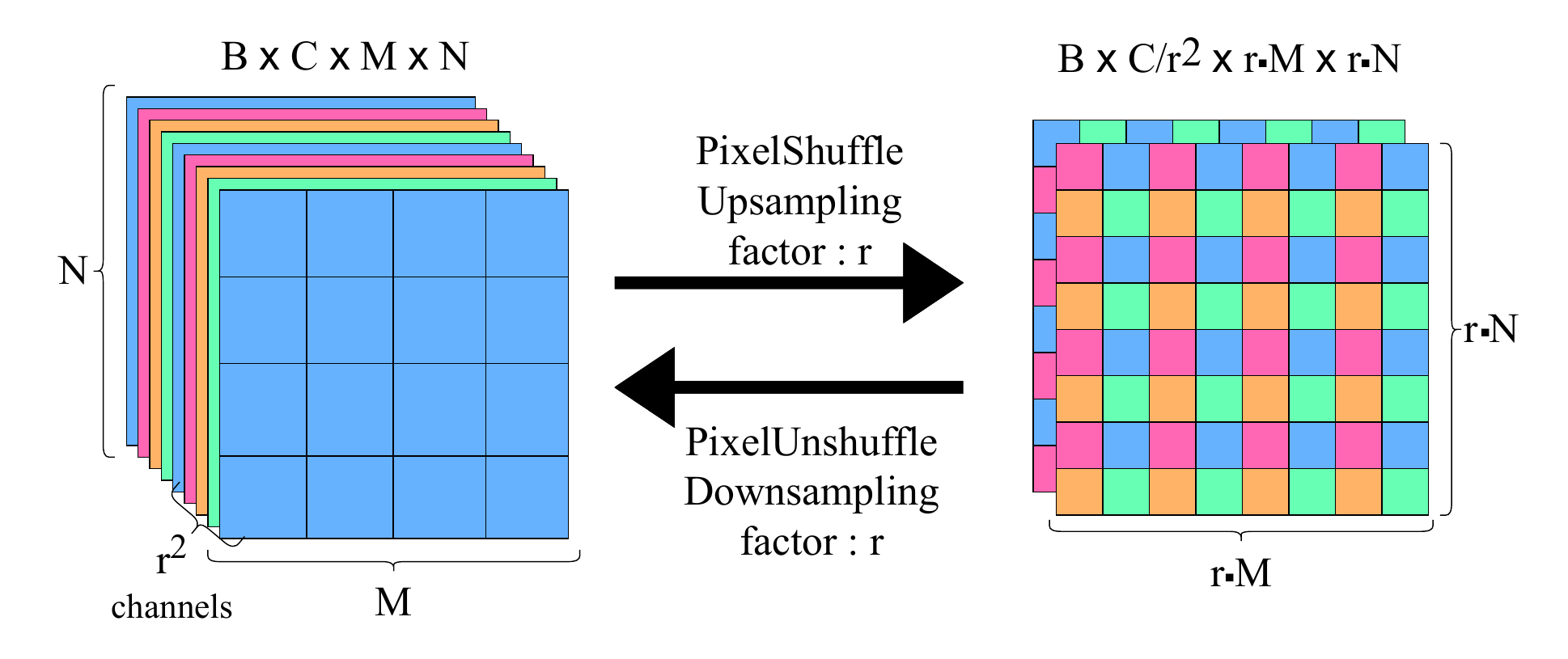}     
    \end{tabular}
\caption{Overview of the PixelShuffle and PixelUnshuffle operations by \cite{pixel_shuffle_unshuffle}.}
\label{fig:pixel_shuffle_unshuffle_overview}   

\end{figure*}
In their work, \cite{pixel_shuffle_unshuffle} proposed an upsampling and downsampling operation, PixelShuffle and PixelUnshuffle respectively. 
\Cref{fig:pixel_shuffle_unshuffle_overview} shows an abstract overview of these sampling operations and how they change the dimensions of the feature maps.
In this work, the models used, Restormer and NAFNet, both use an upsampling and downsampling factor of 2, i.e.,~$\mathrm{r}=2$.
Thus, when upsampling using PixelShuffle, a feature map with dimensions $\mathrm{B} \times \mathrm{C} \times \mathrm{M} \times \mathrm{N}$ becomes of the size $\mathrm{B} \times \frac{\mathrm{C}}{4} \times 2\mathrm{M} \times 2\mathrm{N}$, and the vice-versa is true when downsampling using PixelUnshuffle.

\subsection{Additional Detailed Results}
\label{subsec:appendix:results}

\subsubsection{Image Deblurring}
\label{subsec:appendix:results:deblurring}
\begin{table*}[htbp]
    \centering
    \caption{ \underline{\textbf{Image deblurring}} results. Different downsampling and upsampling techniques are compared based on performance on clean samples and performance against CosPGD and PGD attacks with various attack strengths. Baseline Restormer uses Pixel Unshuffle for downsampling and PixelShuffle for upsampling. As discussed in \cref{subsec:method:downsampling}, we ablate on dropping high frequencies during downsampling with a 30\% probability, denoted as \textbf{+ DropHigh}. Additionally, we ablate on dropping high frequencies with a 30\% probability only in the first downsampling step, denote as \textbf{+ FirstLayerDrop}.}
    %\scriptsize
    \scalebox{0.7}{
    \begin{tabular}{@{}l@{\,}cc|cc|cc|cc|cc|cc|cc@{}}
    \toprule
    \multirow{3}{*}{Architecture} & \multicolumn{2}{c|}{Test} & \multicolumn{6}{c|}{CosPGD} & \multicolumn{6}{c}{PGD} \\
   & & & \multicolumn{2}{c|}{5 attack itrs } & \multicolumn{2}{c|}{10 attack itrs } & \multicolumn{2}{c|}{20 attack itrs } & \multicolumn{2}{c|}{5 attack itrs } & \multicolumn{2}{c|}{10 attack itrs } & \multicolumn{2}{c}{20 attack itrs } \\
   & PSNR & SSIM & PSNR & SSIM & PSNR & SSIM & PSNR & SSIM & PSNR & SSIM & PSNR & SSIM & PSNR & SSIM \\
    \toprule

        \multicolumn{15}{c}{Restormer variants} \\        
        \midrule
         \textbf{Restormer} & \textbf{31.99} & \textbf{0.9635} & 11.36 & 0.3236 & 9.05 & 0.2242 & 7.59 & 0.1548 & 11.41 & 0.3256 & 9.04 & 0.2234 & 7.58 & 0.1543 \\

          ~~~~~ + FLC & 23.85 & 0.7811 & 17.3 & 0.4725 & 16.01 & 0.4031 & 14.66 & 0.3401 & 17.3 & 0.4723 & 16.00 & 0.4030 & 14.66 & 0.3402 \\

          ~~~~~ + FP & 29.95 & 0.9395 & 12.17 & 0.3024 & 10.35 & 0.2117 & 9.44 & 0.1651 & 12.17 & 0.3018 & 10.35 & 0.2112 & 9.45 & 0.1655 \\
         
          ~~~~~ + FP + DropHigh & 29.93 & 0.9395 & 12.67 & 0.3707 & 10.34 & 0.2875 & 9.25 & 0.2355 & 12.17 & 0.371 & 10.34 & 0.2878 & 9.24 & 0.235 \\

          ~~~~~ + FP + FirstLayerDrop & 29.96 & 0.9402 & 13.11 & 0.4458 & 11.00 & 0.3513 & 9.75 & 0.2899 & 13.10 & 0.4458 & 10.98 & 0.3502 & 9.76 & 0.2907 \\

           ~~~~~ + FreqAvgUp & 30.11 & 0.9427 & 13.35 & 0.3831 & 11.20 & 0.2837 & 10.09 & 0.2317 & 13.34 & 0.3828 & 11.21 & 0.2839 & 10.09 & 0.2316 \\

          % ~~~~~ + FP + DropHigh + FreqAvgUp &  \\

          % ~~~~~ + FP + FirstLayerDrop + FreqAvgUp &  \\

          ~~~~~ + FP + FreqAvgUp \textbf{(BoA)} & 26.99 & 0.8806 & \textbf{23.9} & \textbf{0.7578} & \textbf{21.76} & \textbf{0.6875} & \textbf{21.00} & \textbf{0.6351} & \textbf{22.84} & \textbf{0.7290} & \textbf{18.91} & \textbf{0.6390} & \textbf{18.86} & \textbf{0.601}  \\

          \midrule 
        \multicolumn{15}{c}{NAFNet variants} \\        
        \midrule

         \textbf{NAFNet} & \textbf{32.87} & \textbf{0.9606} & 8.67 & 0.2264 & 6.68 & 0.1127  & 5.81 & 0.0617 & 10.27 & 0.3179  & 8.66 & 0.2282  &  5.95 & 0.0714 \\

          ~~~~~ + FP & 31.17 & 0.9439 & 5.60 & 0.0536 & 4.94 & 0.0074 & 4.89 & 0.0001 & 5.60 & 0.0536 & 4.94 & 0.0072 & 4.89 & 0.0002  \\
         
          ~~~~~ + FP + DropHigh & 29.37 & 0.9204 & 14.41 & 0.3266 & 12.33 & 0.2338 & 10.78 & 0.1770 & 14.41 & 0.3266 & 12.34 & 0.2344 & 10.82 & 0.1798  \\

          ~~~~~ + FP + FirstLayerDrop & 30.85 & 0.9408 & 11.0 & 0.3245 & 9.11 & 0.2322 & 7.83 & 0.1632 & 11.00 & 0.3247 & 9.10 & 0.2317 & 7.88 & 0.1662 \\

          ~~~~~ + FreqAvgUp & 31.01 & 0.9420 & 6.97 & 0.0736 & 6.17 & 0.0412 & 5.78 & 0.0305 & 6.97 & 0.0738 & 6.18 & 0.0416 & 5.79 & 0.0299 \\

          ~~~~~ + FP + FreqAvgUp & 30.46 & 0.9356 & 5.13 & 0.0042 & 5.12 & 0.0043 & 5.12 & 0.0043 & 5.13 & 0.0042 & 5.12 & 0.0043 & 5.12 & 0.0043 \\
          
          ~~~~~ + FP + DropHigh + FreqAvgUp \textbf{(BoA)} & 29.58 & 0.9242 & \textbf{17.88} & \textbf{0.5499} & \textbf{15.21} & \textbf{0.4432} & \textbf{13.00} & \textbf{0.3470} & \textbf{17.88} & \textbf{0.5497} & \textbf{15.23} & \textbf{0.4441} & \textbf{13.04} & \textbf{0.3491}  \\

          ~~~~~ + FP + FirstLayerDrop + FreqAvgUp & 30.28 & 0.9337 & 12.13 & 0.2692 & 9.50 & 0.1589 & 7.87 & 0.1036 & 12.13 & 0.2693 & 9.45 & 0.1567 & 7.86 & 0.1037 \\

          %BoA-NAFNet &  \\

    \bottomrule
    \end{tabular}    
    }
    \label{tab:restormer_encoder_ablation_full}
\end{table*}
Following we present additional quantitative results for image deblurring. 
We provide these results in \Cref{tab:restormer_encoder_ablation_full}.
We indicate the ablation setting chosen as \textbf{BoA} for the respective network and task.

\subsubsection{Image Denoising}
\label{subsec:appendix:results:denoising}
This task, unlike image deblurring, requires removing high-frequencies from the images.
This is later discussed in \Cref{sec:appendix:freq_analysis}, where in \Cref{fig:dataset_distribution}, we show how different the three considered tasks are in the frequency domain. Specifically, we show that image denoising input images have a significantly higher power and concentration of high frequencies that need to be removed to restore the images.
Thus, in image denoising, the DNN preserving high-frequency information can adversely affect the DNN's performance.
We observe this in \cref{tab:denoising}, quantitatively.
The \textbf{FreqAvgUp} helps preserve high-frequency information during upsampling, and this is harming the performance of BoA modifications.
However, it's important to note that, high frequencies can also include important features like shapes and edges, so preserving them is essential.
%However, apart from the noise, the high-frequencies could also comprise important features, for example, shape and edges and thus preserving them to some extent is required, and the 
\textbf{FrequencyPreservedPooling} downsampling ensures this, helping the DNN perform better under adversarial attacks by distinguishing important information from noise.
\begin{table*}[htbp]
    \centering
    \caption{\underline{\textbf{Image denosing}} results. Different downsampling and upsampling techniques are compared based on performance on clean samples and performance against CosPGD and PGD attacks with various attack strengths. Baseline Restormer uses Pixel Unshuffle for downsampling and PixelShuffle for upsampling. As discussed in \cref{subsec:method:downsampling}, we ablate over dropping high frequencies during downsampling with a 30\% probability, denoted as \textbf{+ DropHigh}. Additionally, we ablate over dropping high frequencies with a 30\% probability only in the first downsampling step, denoted as\textbf{+ FirstLayerDrop}.}
    %\scriptsize
    \scalebox{0.7}{
    \begin{tabular}{@{}lcc|cc|cc|cc|cc|cc|cc@{}}
    \toprule
    \multirow{3}{*}{Architecture} & \multicolumn{2}{c|}{Test} & \multicolumn{6}{c|}{CosPGD} & \multicolumn{6}{c}{PGD} \\
   & & & \multicolumn{2}{c|}{5 attack itrs } & \multicolumn{2}{c|}{10 attack itrs } & \multicolumn{2}{c|}{20 attack itrs } & \multicolumn{2}{c|}{5 attack itrs } & \multicolumn{2}{c|}{10 attack itrs } & \multicolumn{2}{c}{20 attack itrs } \\
   & PSNR & SSIM & PSNR & SSIM & PSNR & SSIM & PSNR & SSIM & PSNR & SSIM & PSNR & SSIM & PSNR & SSIM \\
    \toprule

        \multicolumn{15}{c}{Restormer variants} \\
        \midrule
        
         \textbf{Restormer} & 40.02 & 0.9706 & 26.76 & 0.9196 & 25.64 & 0.9067 & 25.05 & 0.8961 & 26.73 & 0.9194 & 25.66 & 0.9068 & 25.01 & 0.8961 \\      
                  
          ~~~~~ + FP & 39.88 & 0.9701 & 28.28 & 0.9217 & 27.26 & 0.9075 & 26.75 & 0.8993 & 28.31 & 0.9221 & 27.23 & 0.9082 & 26.72 & 0.8986 \\

          ~~~~~ + FP + DropHigh & 39.88 & 0.9791 & 28.03 & 0.9257 & 26.72 & 0.9058 & 26.08 & 0.8916 & 28.02 & 0.9260 & 26.71 & 0.9056 & 26.09 & 0.8913 \\

          ~~~~~ + FP + FirstLayerDrop  & 39.88 & 0.9701 & 27.56 & 0.9252 & 26.67 & 0.9143 & 26.08 & 0.9034 & 27.56 & 0.9251 & 26.63 & 0.9128 & 26.06 & 0.9028 \\

           ~~~~~ + FreqAvgUp & 39.86 & 0.9701 & 27.66 & 0.9251 & 27.13 & 0.9186 & 26.65 & 0.9137 & 27.66 & 0.9252 & 27.13 & 0.9180 & 26.66 & 0.9143  \\

          ~~~~~ + FP + DropHigh + FreqAvgUp & 39.85 & 0.9700 & 26.63 & 0.8749 & 25.64 & 0.8601 & 25.13 & 0.8462 & 26.61 & 0.8736 & 25.64 & 0.8598 & 25.08 & 0.8428  \\

          ~~~~~ + FP + FirstLayerDrop + FreqAvgUp & 39.86 & 0.9701 & 27.20 & 0.9138 & 26.42 & 0.9041 & 25.92 & 0.8965 & 27.20 & 0.9139 & 26.42 & 0.9046 & 25.92 & 0.8961 \\

          ~~~~~ + FP + FreqAvgUp \textbf{(BoA)} & 39.85 & 0.9701 & 27.10 & 0.9158 & 25.95 & 0.8976 & 25.30 & 0.8830 & 27.09 & 0.9157 & 25.94 & 0.8970 & 25.31 & 0.8839 \\

          \midrule 
        \multicolumn{15}{c}{NAFNet variants} \\        
        \midrule
        
        \textbf{NAFNet} &  39.97 & 0.9599 & 29.36 & 0.8494 & 28.47 & 0.8260 & 27.97 & 0.8047 & 29.36 & 0.8494 & 28.43 & 0.8243 & 27.96 &  0.8032 \\

        ~~~~~ + FP  & 39.73 & 0.9584 & 29.11 & 0.8277 & 27.91 & 0.7841 & 26.93 & 0.7366 & 29.11 & 0.8277 & 27.92 & 0.7841 & 26.94 & 0.7369 \\
        
        ~~~~~ + FP + DropHigh  & 36.57 & 0.9574 & 28.22 & 0.7914 & 26.41 & 0.7244 & 24.55 & 0.6426 & 28.22 & 0.7914 & 26.41 & 0.7243 & 24.57 & 0.6437 \\

          ~~~~~ + FP + FirstLayerDrop  & 39.71 & 0.9583 & 29.14 & 0.8337 & 27.98 & 0.7953 & 27.31 & 0.7636 & 29.14 & 0.8337 & 27.98 & 0.7952 & 27.33 & 0.7637 \\

          ~~~~~ + FreqAvgUp & 39.75 & 0.9585 & 27.90 & 0.7839 & 26.69 & 0.7395 & 25.69 & 0.6931 & 27.91 & 0.7844 & 26.65 & 0.7383 & 25.67 & 0.6921 \\

          ~~~~~ + FP + DropHigh + FreqAvgUp & 39.54 & 0.9577 & 28.58 & 0.8099 & 26.90 & 0.7383 & 24.59 & 0.6252 & 28.58 & 0.8099 & 26.90 & 0.7383 & 24.58 & 0.6246  \\

          ~~~~~ + FP + FirstLayerDrop + FreqAvgUp & 39.72 & 0.9582 & 28.39 & 0.7838 & 25.90 & 0.6938 & 23.78 & 0.6152 & 28.39 & 0.7838 & 25.88 & 0.6931 & 23.80 & 0.6163 \\

           ~~~~~ + FP + FreqAvgUp \textbf{(BoA)} & 39.64 & 0.9577 & 28.58 & 0.8099 & 26.90 & 0.7383 & 24.59 & 0.6252 & 28.58 & 0.8099 & 26.90 & 0.7383 & 24.58 & 0.6246 \\

    \bottomrule
    \end{tabular}    
    }
    \label{tab:denoising}
\end{table*}

\subsubsection{Image Deraining}
\label{subsec:appendix:results:deraining}
\begin{figure*}[ht]
    \centering % <-- added
   \includegraphics[width=\linewidth]{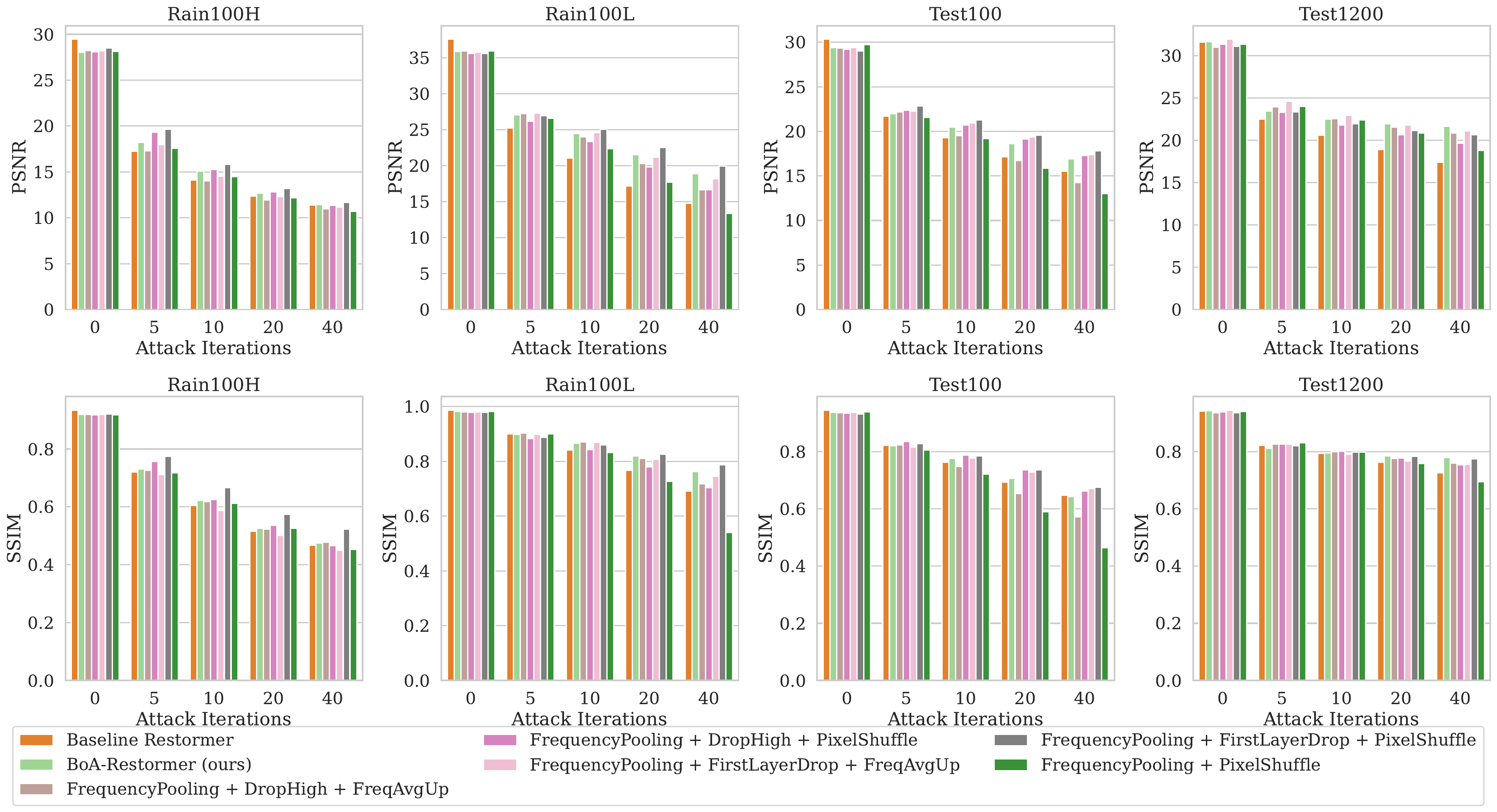}
\caption{\underline{\textbf{Image deraining}} results under CosPGD attack.}
\label{fig:deraining_cospgd_full}
\end{figure*}

\begin{figure*}[ht]
    \centering % <-- added
   \includegraphics[width=\linewidth]{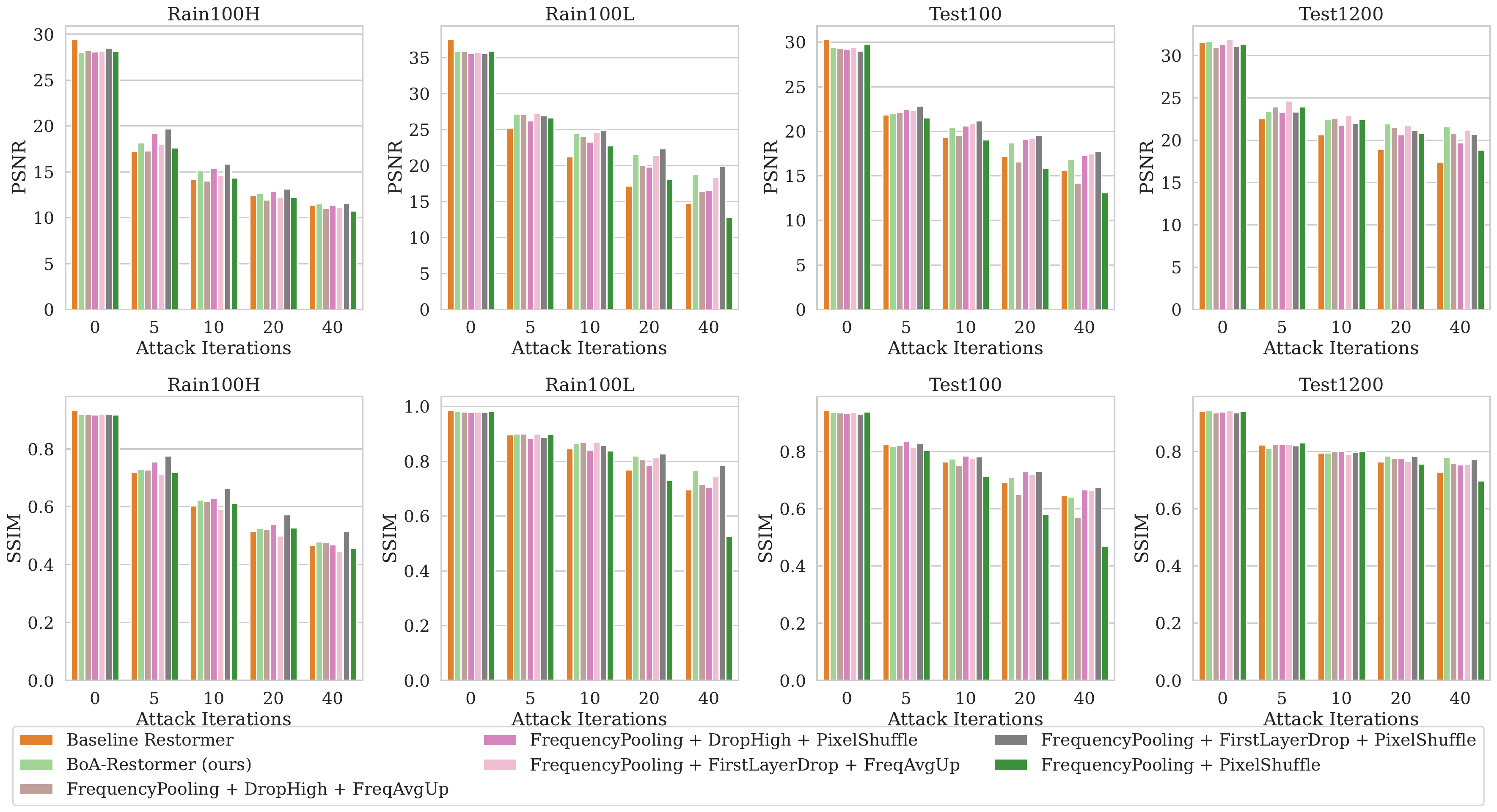}
\caption{\underline{\textbf{Image deraining}} results under PGD attack.}
\label{fig:deraining_pgd_full}
\end{figure*}
We provide additional quantitative results for image deraining in \Cref{fig:deraining_cospgd_full} and \Cref{fig:deraining_pgd_full} under CosPGD and PGD attack, respectively.

\subsection{Additional Qualitative Results}
\label{sec:appendix:pgd_images}

Following we provide qualitative results for the models considered in \cref{fig:restormer_cospgd_attack_different_downnsamplings5} and \cref{fig:deraining_images_restormer_cospgd} under PGD attack and CosPGD and more attack iterations for image deblurring and image deraining, respectively.

\begin{figure*}[htb]
    \centering % <-- added
\scalebox{0.92}{
   \begin{tabular}{@{}c@{\hspace{0.2cm}}c@{\hspace{0.1cm}}c@{\hspace{0.1cm}}c@{\hspace{0.1cm}}c@{\hspace{0.1cm}}c@{\hspace{0.1cm}}c@{}}
    \multicolumn{3}{c}{MODEL} & NO ATTACK & 5 iterations & 10 iterations & 20 iterations\\
  \rotatebox{90}{\textbf{FSNet\cite{cui2023selective}}} & & &
  \includegraphics[width=0.23\textwidth]{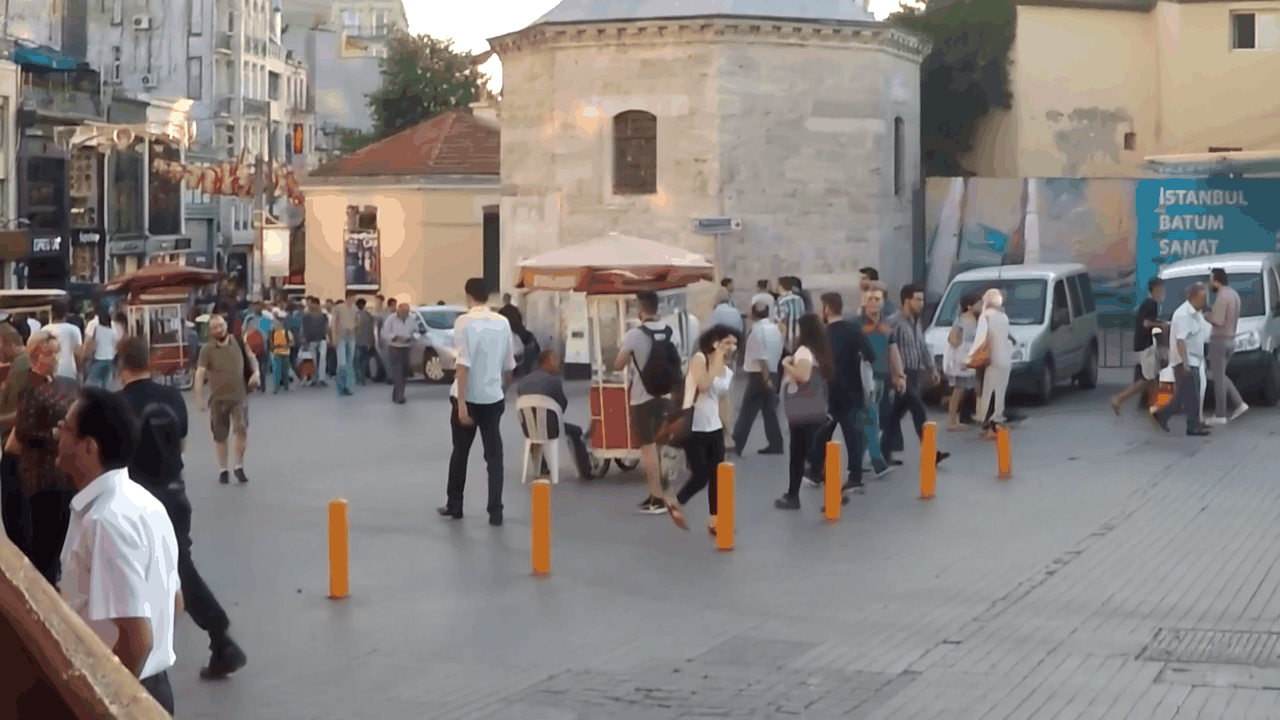}&
  \includegraphics[width=0.23\textwidth]{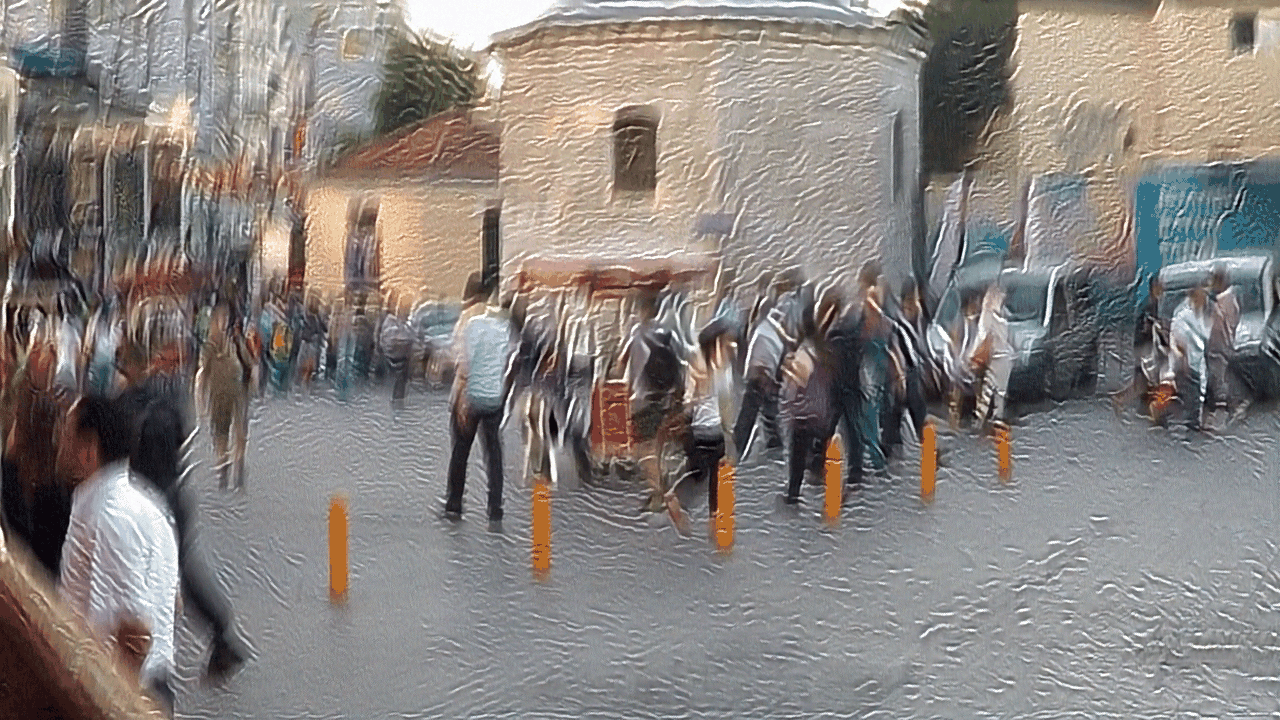}&
  \includegraphics[width=0.23\textwidth]{figures/FSNet_deblurring2/FSNet_pgd_10_GOPR0384_11_00_000002.png}&
  \includegraphics[width=0.23\textwidth]{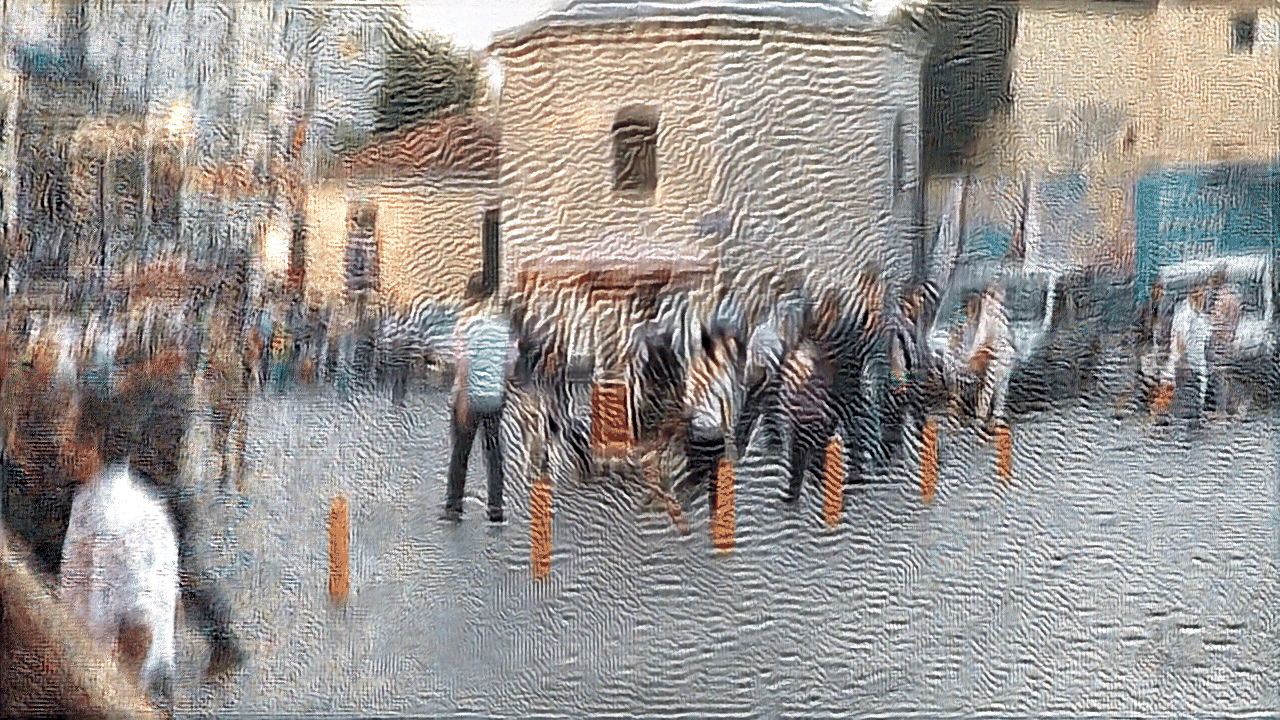}\\

  \rotatebox{90}{\textbf{Restormer}} & & &
  \includegraphics[width=0.23\textwidth]{figures/old_figures/Restormer_no_attack_GOPR0384_11_00-000002.png}&
  \includegraphics[width=0.23\textwidth]{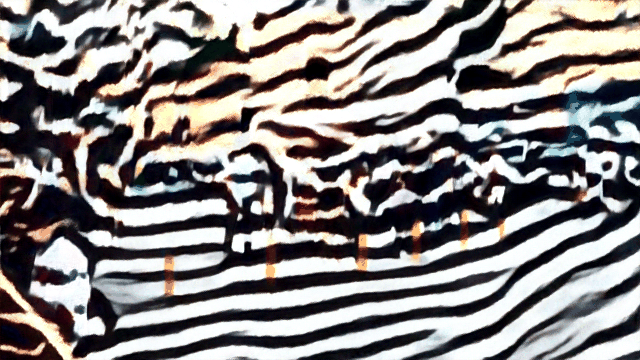}&
  \includegraphics[width=0.23\textwidth]{figures/old_figures/Restormer_pgd_10_GOPR0384_11_00-000002.png}&
  \includegraphics[width=0.23\textwidth]{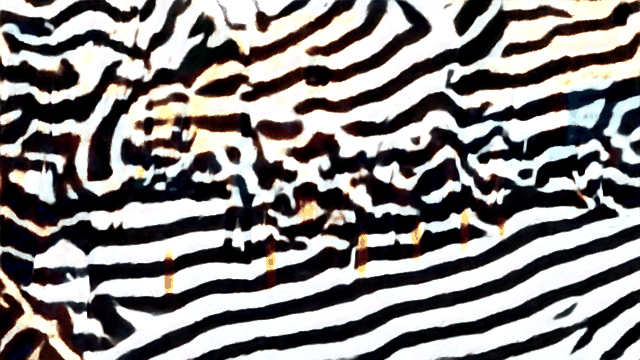}\\

  \rotatebox{90}{\textbf{Restormer}} & \rotatebox{90}{ + FLC} & &
  \includegraphics[width=0.23\textwidth]{figures/old_figures/Restormer_FLC_no_attack_GOPR0384_11_00-000002.png}&
  \includegraphics[width=0.23\textwidth]{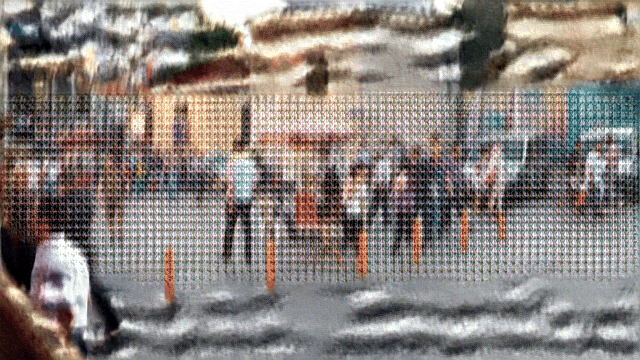}&
  \includegraphics[width=0.23\textwidth]{figures/old_figures/Restormer_FLC_pgd_10_GOPR0384_11_00-000002.png}&
  \includegraphics[width=0.23\textwidth]{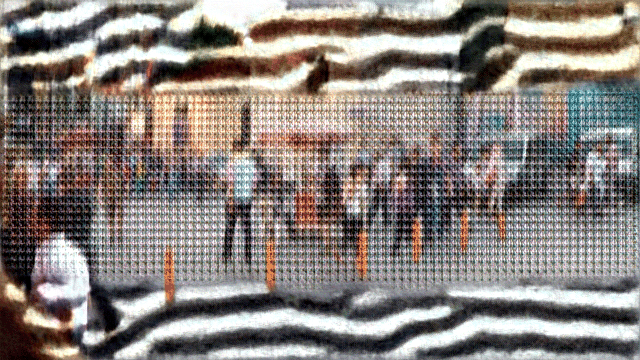}\\

  \rotatebox{90}{\textbf{Restormer}} & \rotatebox{90}{+ FP} &  &
  \includegraphics[width=0.23\textwidth]{figures/Restormer_FP/no_drop_no_attack_GOPR0384_11_00-000002.png}&
  \includegraphics[width=0.23\textwidth]{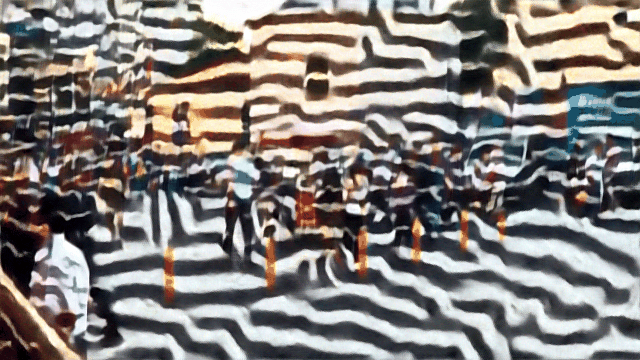}&
  \includegraphics[width=0.23\textwidth]{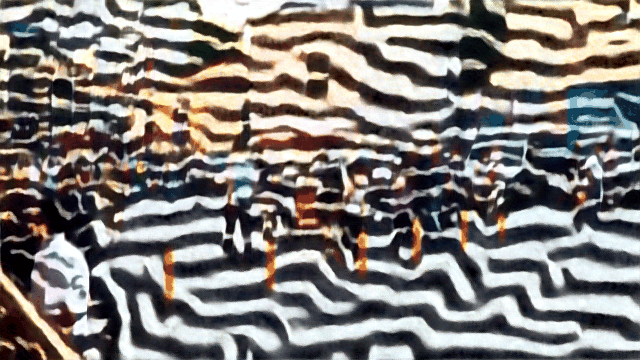}&
  \includegraphics[width=0.23\textwidth]{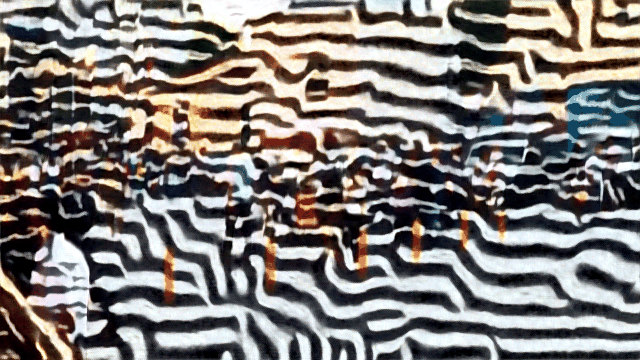}
  \\

\rotatebox{90}{\textbf{Restormer}} &  \rotatebox{90}{+ FreqAvgUp} & &
  \includegraphics[width=0.23\textwidth]{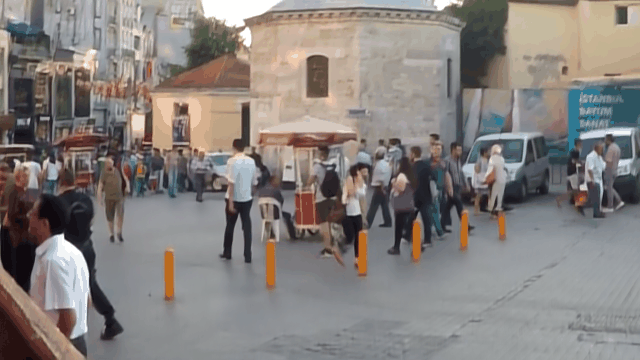}&
  \includegraphics[width=0.23\textwidth]{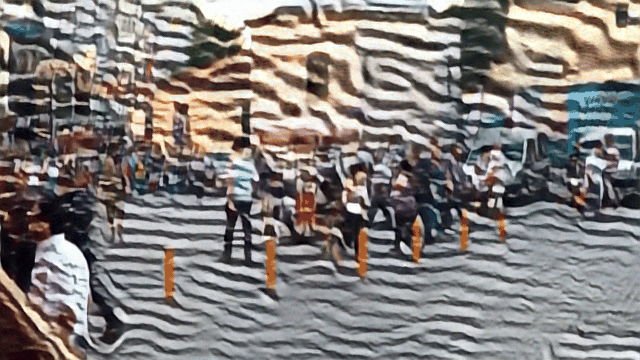}&
  \includegraphics[width=0.23\textwidth]{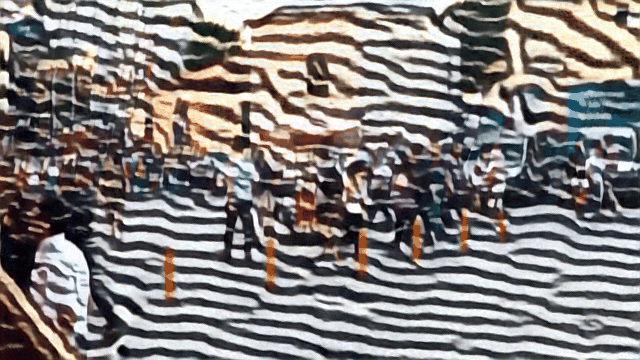}&
  \includegraphics[width=0.23\textwidth]{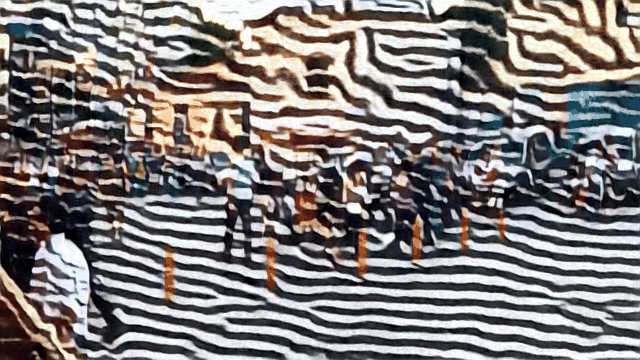}
  \\
 
  \rotatebox{90}{\textbf{Restormer}} & \rotatebox{90}{+ FP} & \rotatebox{90}{+ FreqAvgUp} &
  \includegraphics[width=0.23\textwidth]{figures/Restormer_FP_FreqAvgUp/no_drop_freqAvgUp_no_attack_GOPR0384_11_00-000002.png}&
  \includegraphics[width=0.23\textwidth]{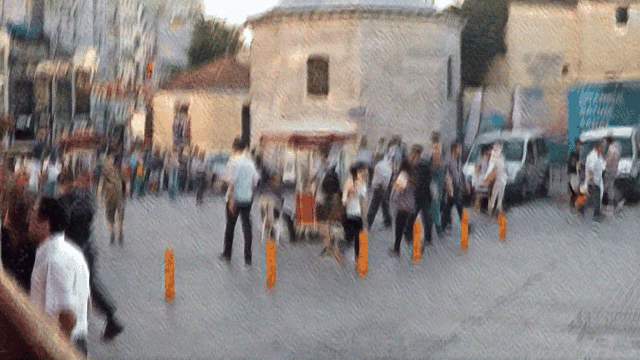}&
  \includegraphics[width=0.23\textwidth]{figures/Restormer_FP_FreqAvgUp/no_drop_freqAvgUp_pgd_10_GOPR0384_11_00-000002.png}&
  \includegraphics[width=0.23\textwidth]{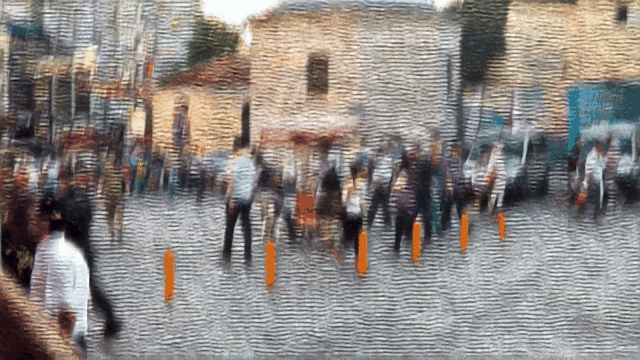}
  \\

\end{tabular}
}
\caption{\underline{\textbf{Image Deblurring}} results using \textbf{Restormer}. Comparing the proposed architectural design choices qualitatively against other baselines for image restoration, on normal blurry input images and input images adversarially attacked using PGD. Symbolic notations are the same as those in \cref{fig:deblurring_main}.}
\label{fig:restormer_pgd_attack}
\end{figure*}
\begin{figure*}[htb]
    \centering % <-- added
\scalebox{0.92}{
   \begin{tabular}{@{}c@{\hspace{0.2cm}}c@{\hspace{0.1cm}}c@{\hspace{0.1cm}}c@{\hspace{0.1cm}}c@{\hspace{0.1cm}}c@{\hspace{0.1cm}}c@{}}
    \multicolumn{3}{c}{MODEL} & NO ATTACK & 5 iterations & 10 iterations & 20 iterations\\
  \rotatebox{90}{\textbf{Restormer}} & & &
  \includegraphics[width=0.23\textwidth]{figures/old_figures/Restormer_no_attack_GOPR0384_11_00-000002.png}&
  \includegraphics[width=0.23\textwidth]{figures/old_figures/Restormer_pgd_5_GOPR0384_11_00-000002.png}&
  \includegraphics[width=0.23\textwidth]{figures/old_figures/Restormer_pgd_10_GOPR0384_11_00-000002.png}&
  \includegraphics[width=0.23\textwidth]{figures/old_figures/Restormer_pgd_20_GOPR0384_11_00-000002.png}\\

  \rotatebox{90}{\textbf{Restormer}} & \rotatebox{90}{+ FP} &  &
  \includegraphics[width=0.23\textwidth]{figures/Restormer_FP/no_drop_no_attack_GOPR0384_11_00-000002.png}&
  \includegraphics[width=0.23\textwidth]{figures/Restormer_FP/no_drop_pgd_5_GOPR0384_11_00-000002.png}&
  \includegraphics[width=0.23\textwidth]{figures/Restormer_FP/no_drop_pgd_10_GOPR0384_11_00-000002.png}&
  \includegraphics[width=0.23\textwidth]{figures/Restormer_FP/no_drop_pgd_20_GOPR0384_11_00-000002.png}
  \\

  \rotatebox{90}{\textbf{Restormer}} & \rotatebox{90}{+ FP} & \rotatebox{90}{+ SplitUp} &
  \includegraphics[width=0.23\textwidth]{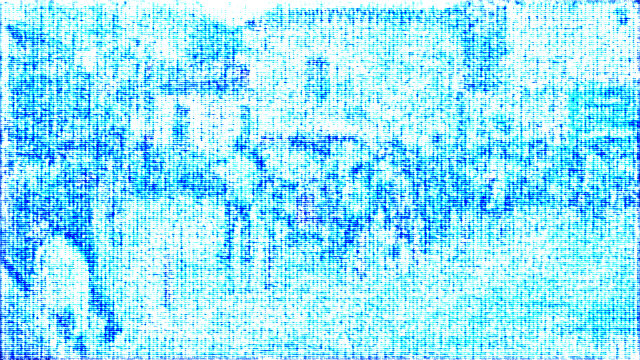}&
  \includegraphics[width=0.23\textwidth]{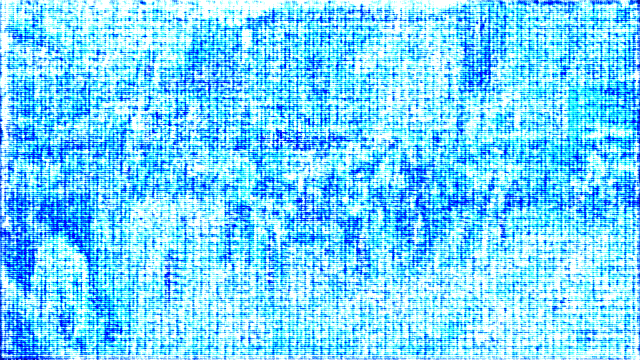}&
  \includegraphics[width=0.23\textwidth]{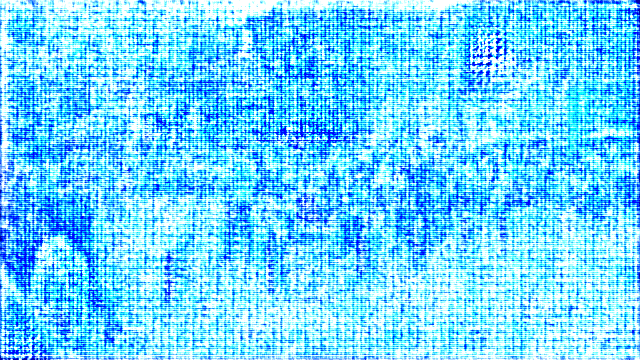}&
  \includegraphics[width=0.23\textwidth]{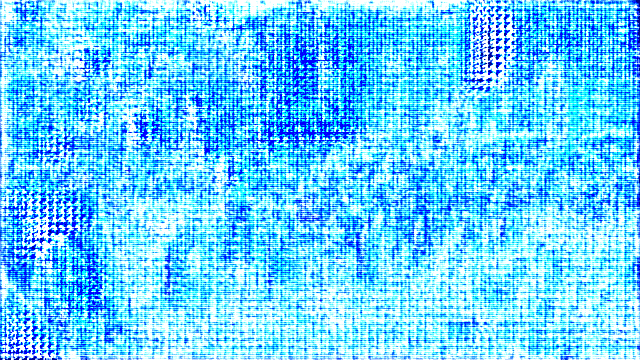}
  \\

  \rotatebox{90}{\textbf{Restormer}} & \rotatebox{90}{+ FP} & \rotatebox{90}{+ FreqAvgUp} &
  \includegraphics[width=0.23\textwidth]{figures/Restormer_FP_FreqAvgUp/no_drop_freqAvgUp_no_attack_GOPR0384_11_00-000002.png}&
  \includegraphics[width=0.23\textwidth]{figures/Restormer_FP_FreqAvgUp/no_drop_freqAvgUp_pgd_5_GOPR0384_11_00-000002.png}&
  \includegraphics[width=0.23\textwidth]{figures/Restormer_FP_FreqAvgUp/no_drop_freqAvgUp_pgd_10_GOPR0384_11_00-000002.png}&
  \includegraphics[width=0.23\textwidth]{figures/Restormer_FP_FreqAvgUp/no_drop_freqAvgUp_pgd_20_GOPR0384_11_00-000002.png}
  \\
\end{tabular}
}
\caption{\underline{\textbf{Image Deblurring}} results using \textbf{Restormer}. Different upsampling methods are being compared qualitatively on normal blurry input images and input images adversarially attacked using PGD. Symbolic notations are the same as those in \cref{tab:restormer_decoder_ablation}.}
\label{fig:restormer_pgd_attack_upsampling}
\end{figure*}

\begin{figure*}[htb]
    \centering % <-- added
\scalebox{0.92}{
   \begin{tabular}{@{}c@{\hspace{0.2cm}}c@{\hspace{0.1cm}}c@{\hspace{0.1cm}}c@{\hspace{0.1cm}}c@{\hspace{0.1cm}}c@{\hspace{0.1cm}}c@{}}
    \multicolumn{3}{c}{MODEL} & NO ATTACK & 5 iterations & 10 iterations & 20 iterations\\

  \rotatebox{90}{\textbf{Restormer}} & & &
  \includegraphics[width=0.23\textwidth]{figures/old_figures/Restormer_no_attack_GOPR0384_11_00-000002.png}&
  \includegraphics[width=0.23\textwidth]{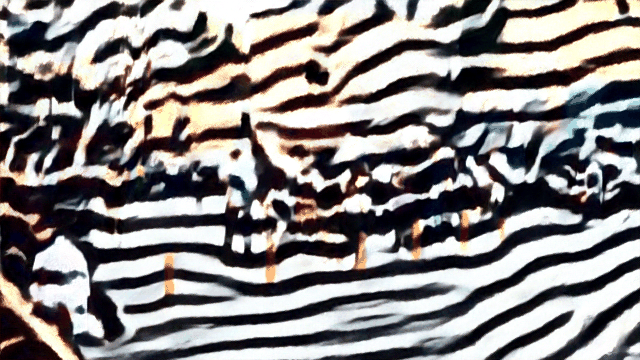}&
  \includegraphics[width=0.23\textwidth]{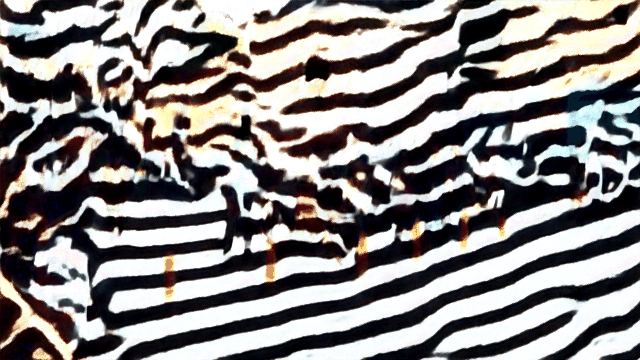}&
  \includegraphics[width=0.23\textwidth]{figures/old_figures/Restormer_cospgd_20_GOPR0384_11_00-000002.png}\\

  \rotatebox{90}{\textbf{Restormer}} & \rotatebox{90}{ + FLC} & &
  \includegraphics[width=0.23\textwidth]{figures/old_figures/Restormer_FLC_no_attack_GOPR0384_11_00-000002.png}&
  \includegraphics[width=0.23\textwidth]{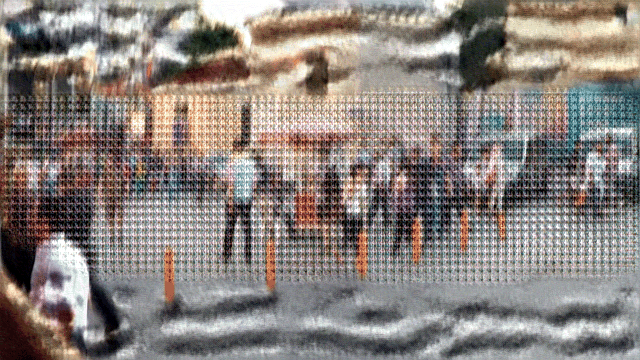}&
  \includegraphics[width=0.23\textwidth]{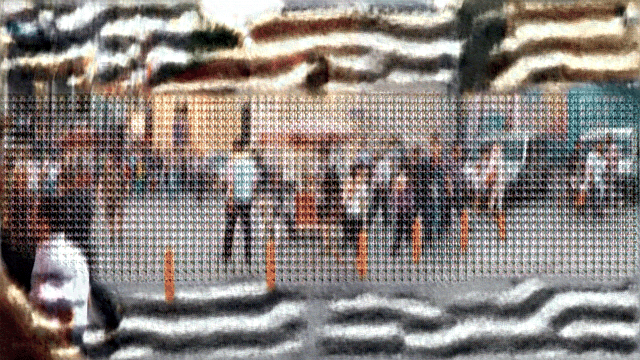}&
  \includegraphics[width=0.23\textwidth]{figures/old_figures/Restormer_FLC_cospgd_20_GOPR0384_11_00-000002.png}\\

  \rotatebox{90}{\textbf{Restormer}} & \rotatebox{90}{+ FP} &  &
  \includegraphics[width=0.23\textwidth]{figures/Restormer_FP/no_drop_no_attack_GOPR0384_11_00-000002.png}&
  \includegraphics[width=0.23\textwidth]{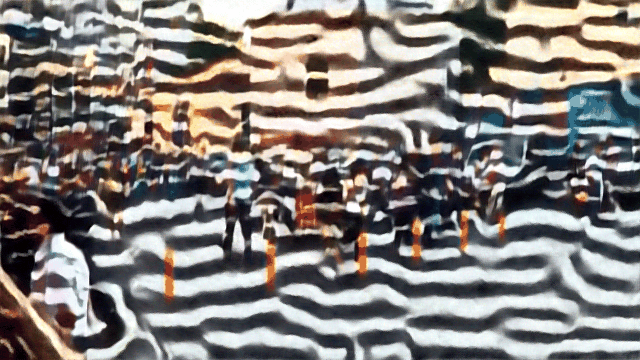}&
  \includegraphics[width=0.23\textwidth]{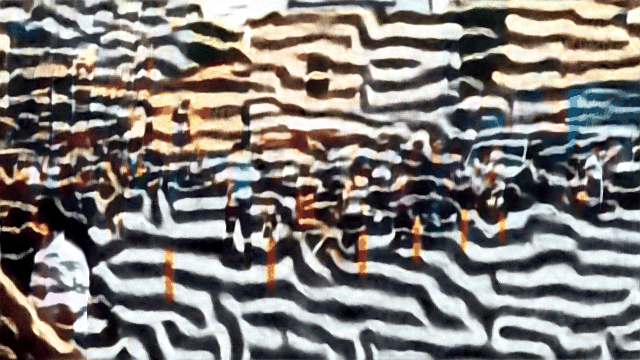}&
  \includegraphics[width=0.23\textwidth]{figures/Restormer_FP/no_drop_cospgd_20_GOPR0384_11_00-000002.png}
  \\

  \rotatebox{90}{\textbf{Restormer}} & \rotatebox{90}{+ FP} & \rotatebox{90}{+ DropHigh} &
  \includegraphics[width=0.23\textwidth]{figures/Restormer_FP_drop/with_drop_no_attack_GOPR0384_11_00-000002.png}&
  \includegraphics[width=0.23\textwidth]{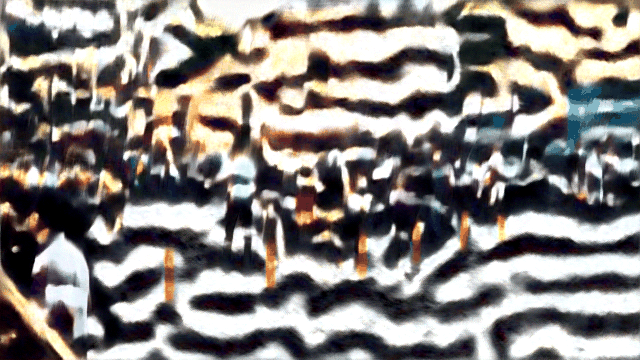}&
  \includegraphics[width=0.23\textwidth]{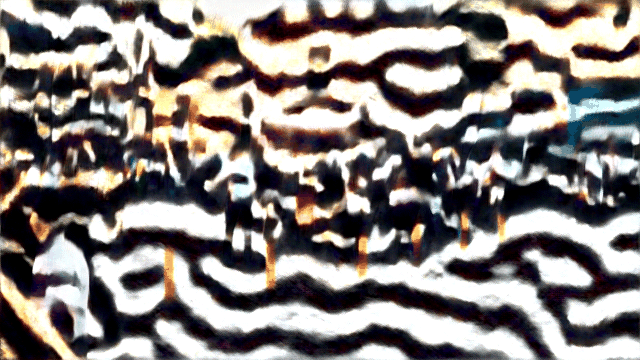}&
  \includegraphics[width=0.23\textwidth]{figures/Restormer_FP_drop/with_drop_cospgd_20_GOPR0384_11_00-000002.png}
  \\

  \rotatebox{90}{\textbf{Restormer}} & \rotatebox{90}{+ FP} & \rotatebox{90}{\tiny + FirstLayerDrop} &
  \includegraphics[width=0.23\textwidth]{figures/Restormer_FP_first_step_drop/first_layer_drop_no_attack_GOPR0384_11_00-000002.png}&
  \includegraphics[width=0.23\textwidth]{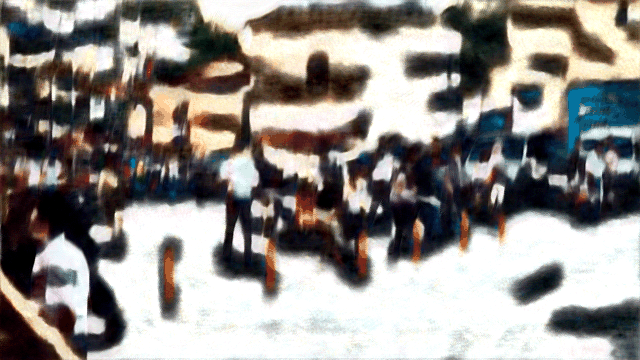}&
  \includegraphics[width=0.23\textwidth]{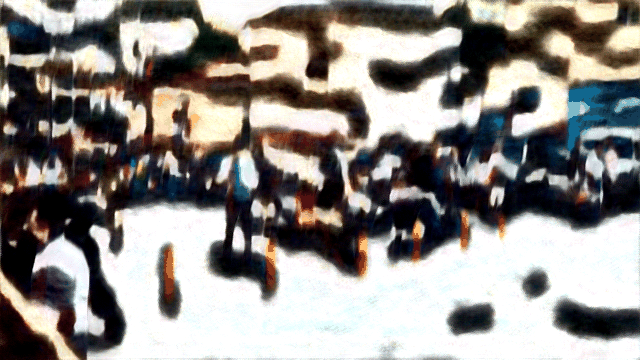}&
  \includegraphics[width=0.23\textwidth]{figures/Restormer_FP_first_step_drop/first_layer_drop_cospgd_20_GOPR0384_11_00-000002.png}
  \\

  \rotatebox{90}{\textbf{Restormer}} &  \rotatebox{90}{+ FreqAvgUp} & &
  \includegraphics[width=0.23\textwidth]{figures/deblurring_restormer_freqAvgUp/restormer_freqAvgUp_no_attack_GOPR0384_11_00-000002_restored_after_attack.png}&
  \includegraphics[width=0.23\textwidth]{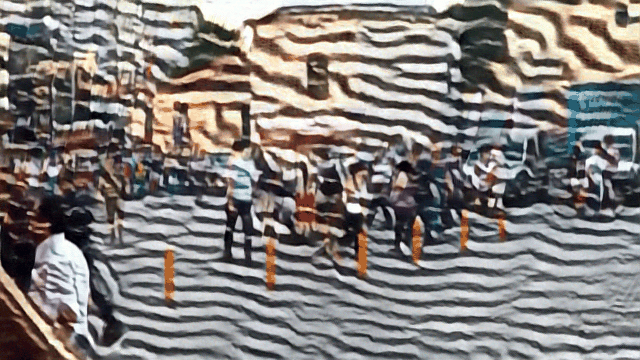}&
  \includegraphics[width=0.23\textwidth]{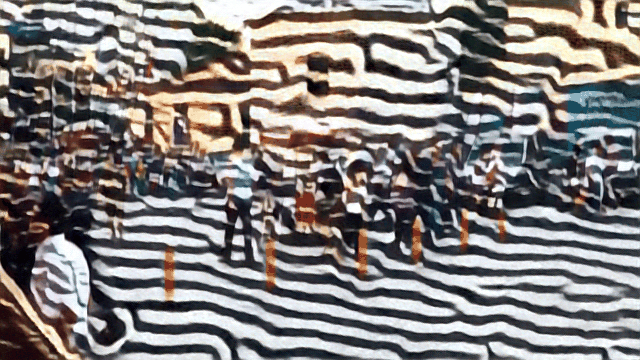}&
  \includegraphics[width=0.23\textwidth]{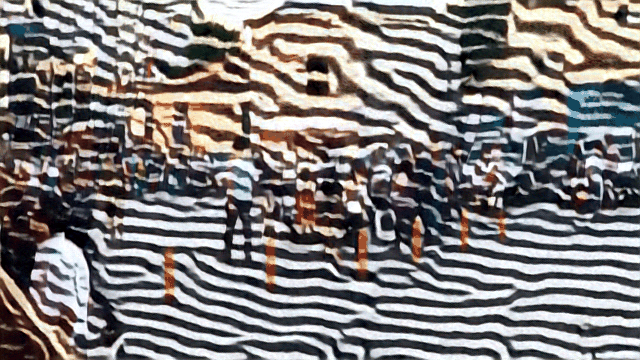}
  \\

  \rotatebox{90}{\textbf{Restormer}} & \rotatebox{90}{+ FP} & \rotatebox{90}{+ FreqAvgUp} &
  \includegraphics[width=0.23\textwidth]{figures/Restormer_FP_FreqAvgUp/no_drop_freqAvgUp_no_attack_GOPR0384_11_00-000002.png}&
  \includegraphics[width=0.23\textwidth]{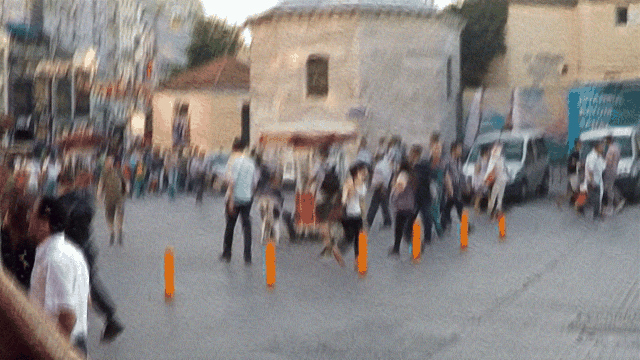}&
  \includegraphics[width=0.23\textwidth]{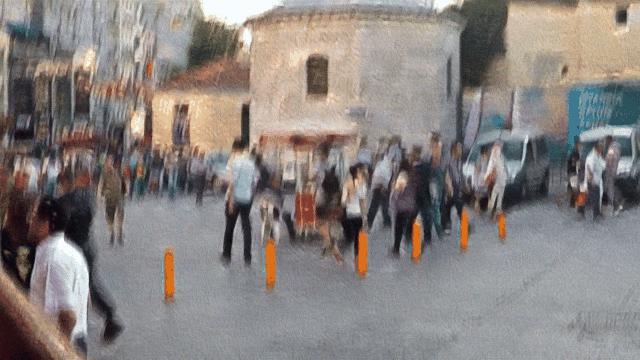}&
  \includegraphics[width=0.23\textwidth]{figures/Restormer_FP_FreqAvgUp/no_drop_freqAvgUp_cospgd_20_GOPR0384_11_00-000002.png}
  \\

\end{tabular}
}
\caption{\underline{\textbf{Image Deblurring}} results using \textbf{Restormer}. Different downsampling methods are being compared qualitatively on normal blurry input images and input images adversarially attacked using CosPGD.}
\label{fig:restormer_cospgd_attack_different_downnsamplings}
\end{figure*}
\begin{figure*}[htb]
    \centering % <-- added
\scalebox{0.92}{
   \begin{tabular}{@{}c@{\hspace{0.2cm}}c@{\hspace{0.1cm}}c@{\hspace{0.1cm}}c@{\hspace{0.1cm}}c@{\hspace{0.1cm}}c@{\hspace{0.1cm}}c@{}}
    \multicolumn{3}{c}{MODEL} & NO ATTACK & 5 iterations & 10 iterations & 20 iterations\\
  \rotatebox{90}{\textbf{Restormer}} & & &
  \includegraphics[width=0.23\textwidth]{figures/old_figures/Restormer_no_attack_GOPR0384_11_00-000002.png}&
  \includegraphics[width=0.23\textwidth]{figures/old_figures/Restormer_cospgd_5_GOPR0384_11_00-000002.png}&
  \includegraphics[width=0.23\textwidth]{figures/old_figures/Restormer_cospgd_10_GOPR0384_11_00-000002.png}&
  \includegraphics[width=0.23\textwidth]{figures/old_figures/Restormer_cospgd_20_GOPR0384_11_00-000002.png}\\

  \rotatebox{90}{\textbf{Restormer}} & \rotatebox{90}{+ FP} &  &
  \includegraphics[width=0.23\textwidth]{figures/Restormer_FP/no_drop_no_attack_GOPR0384_11_00-000002.png}&
  \includegraphics[width=0.23\textwidth]{figures/Restormer_FP/no_drop_cospgd_5_GOPR0384_11_00-000002.png}&
  \includegraphics[width=0.23\textwidth]{figures/Restormer_FP/no_drop_cospgd_10_GOPR0384_11_00-000002.png}&
  \includegraphics[width=0.23\textwidth]{figures/Restormer_FP/no_drop_cospgd_20_GOPR0384_11_00-000002.png}
  \\

  \rotatebox{90}{\textbf{Restormer}} & \rotatebox{90}{+ FP} & \rotatebox{90}{+ SplitUp} &
  \includegraphics[width=0.23\textwidth]{figures/Restormer_FP_SplitUp/no_drop_splitUp_no_attack_GOPR0384_11_00-000002.png}&
  \includegraphics[width=0.23\textwidth]{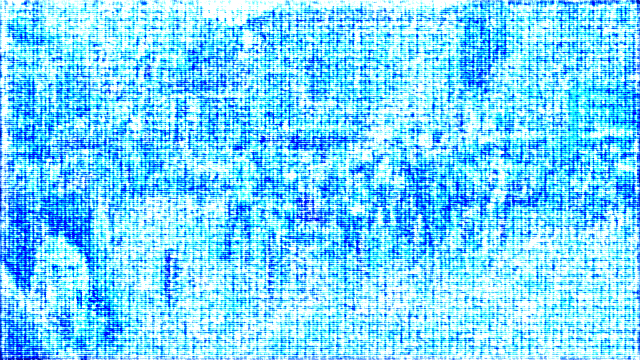}&
  \includegraphics[width=0.23\textwidth]{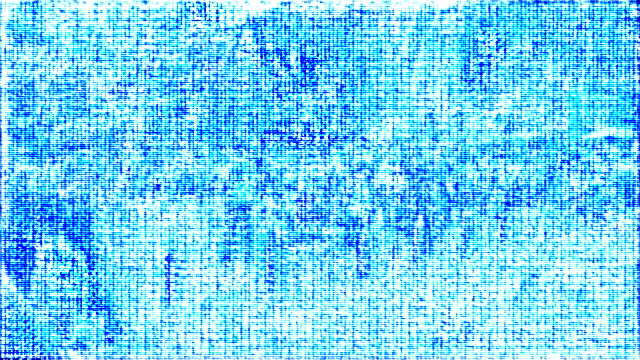}&
  \includegraphics[width=0.23\textwidth]{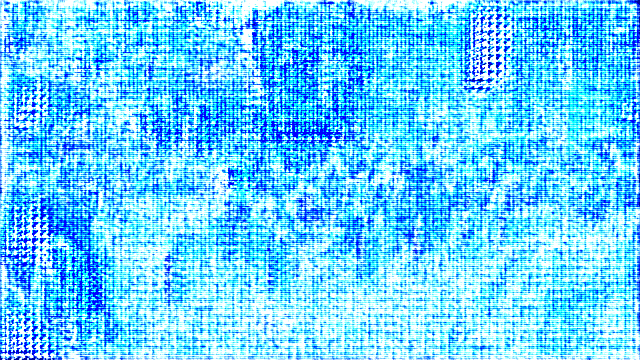}
  \\

  \rotatebox{90}{\textbf{Restormer}} & \rotatebox{90}{+ FP} & \rotatebox{90}{+ FreqAvgUp} &
  \includegraphics[width=0.23\textwidth]{figures/Restormer_FP_FreqAvgUp/no_drop_freqAvgUp_no_attack_GOPR0384_11_00-000002.png}&
  \includegraphics[width=0.23\textwidth]{figures/Restormer_FP_FreqAvgUp/no_drop_freqAvgUp_cospgd_5_GOPR0384_11_00-000002.png}&
  \includegraphics[width=0.23\textwidth]{figures/Restormer_FP_FreqAvgUp/no_drop_freqAvgUp_cospgd_10_GOPR0384_11_00-000002.png}&
  \includegraphics[width=0.23\textwidth]{figures/Restormer_FP_FreqAvgUp/no_drop_freqAvgUp_cospgd_20_GOPR0384_11_00-000002.png}
  \\
\end{tabular}
}
\caption{\underline{\textbf{Image Deblurring}} results using \textbf{Restormer}. Different upsampling methods are being compared qualitatively on normal blurry input images and input images adversarially attacked using CosPGD. Symbolic notations are the same as those in \cref{tab:restormer_decoder_ablation}.}
\label{fig:restormer_cospgd_attack_upsampling}
\end{figure*}

\begin{figure*}[htb]
    \centering % <-- added
\scalebox{0.75}{
   \begin{tabular}{@{}c@{\hspace{0.2cm}}c@{\hspace{0.1cm}}c@{\hspace{0.1cm}}c@{\hspace{0.1cm}}c@{\hspace{0.1cm}}c@{\hspace{0.1cm}}c@{\hspace{0.1cm}}c@{}}
    \multicolumn{4}{c}{MODEL} & NO ATTACK & 5 iterations & 10 iterations & 20 iterations\\

  \rotatebox{90}{\phantom{bbbb}\textbf{Restormer}} & & & &
  \includegraphics[height=3cm, width=0.24\textwidth]{figures/deraining_images/baseline_restormer_no_attack_22_restored_after_attack.png}&
  \includegraphics[height=3cm, width=0.24\textwidth]{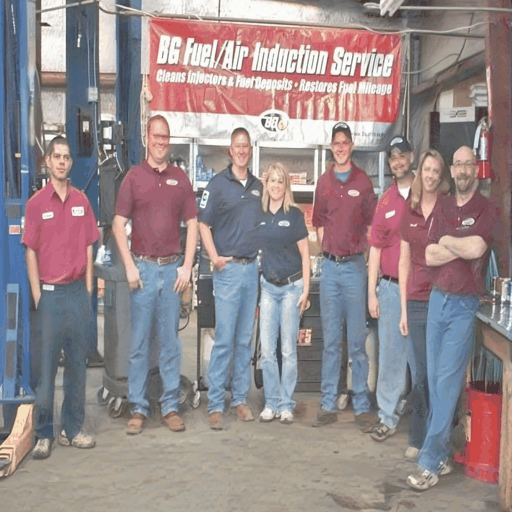}&
  \includegraphics[height=3cm, width=0.24\textwidth]{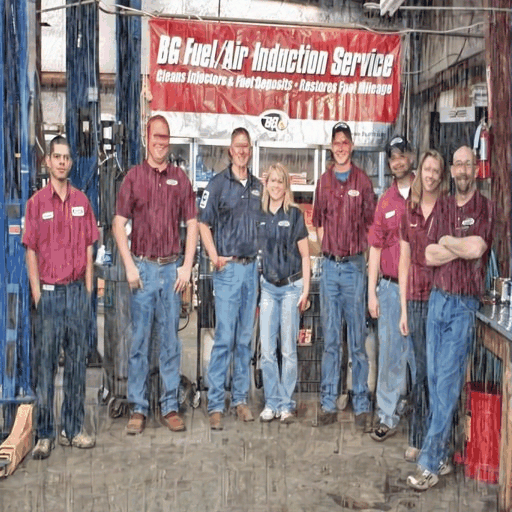}&
  \includegraphics[height=3cm, width=0.24\textwidth]{figures/deraining_images/baseline_restormer_cospgd_20_22_restored_after_attack.png}\\

  \rotatebox{90}{\phantom{bbbb}\textbf{Restormer}} & \rotatebox{90}{\phantom{bbbb}+ FP} &  & &
   \includegraphics[height=3cm, width=0.24\textwidth]{figures/deraining_images/restormer_fp_no_attack_22_restored_after_attack.png}&
  \includegraphics[height=3cm, width=0.24\textwidth]{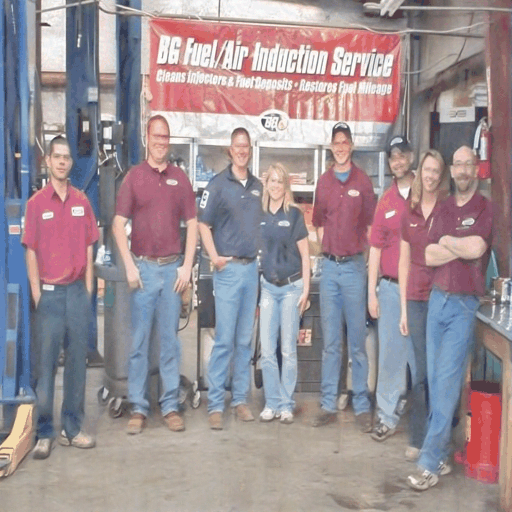}&
  \includegraphics[height=3cm, width=0.24\textwidth]{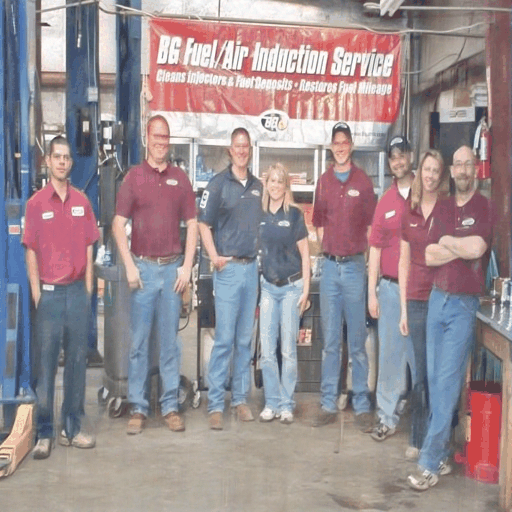}&
  \includegraphics[height=3cm, width=0.24\textwidth]{figures/deraining_images/restormer_fp_cospgd_20_22_restored_after_attack.png}\\

  \rotatebox{90}{\phantom{bbbb}\textbf{Restormer}} & \rotatebox{90}{\phantom{bbbb}+ FP} & \rotatebox{90}{\phantom{bbbb}+ DropHigh} & &
   \includegraphics[height=3cm, width=0.24\textwidth]{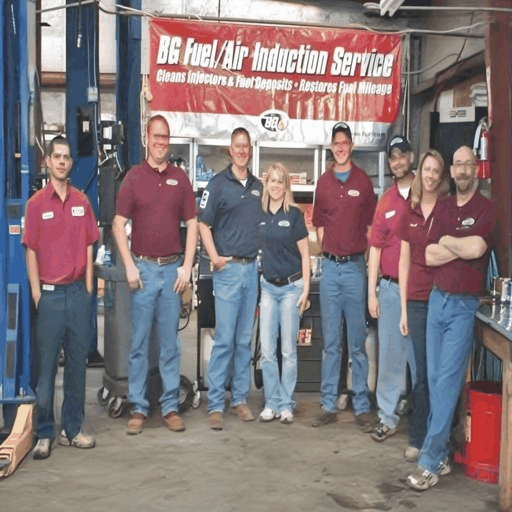}&
  \includegraphics[height=3cm, width=0.24\textwidth]{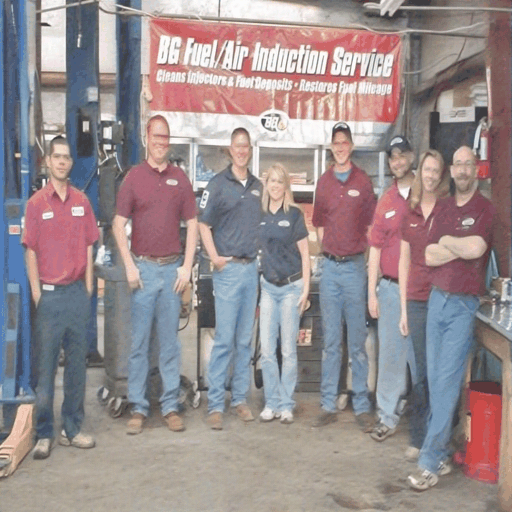}&
  \includegraphics[height=3cm, width=0.24\textwidth]{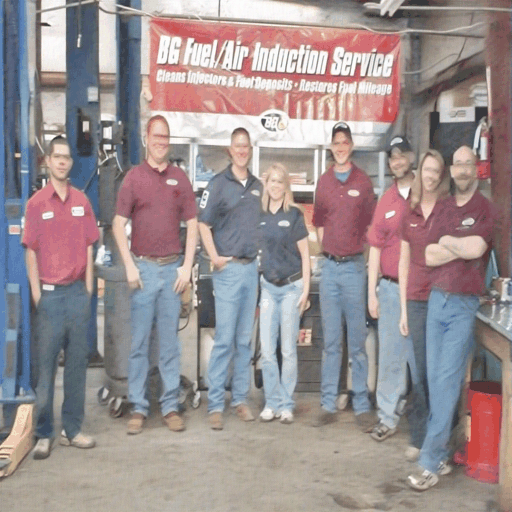}&
  \includegraphics[height=3cm, width=0.24\textwidth]{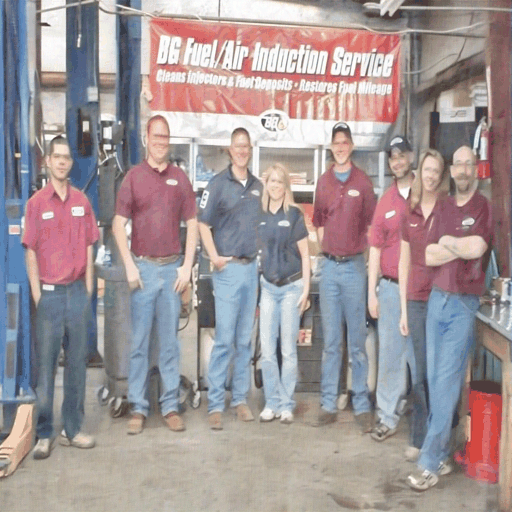}\\

  \rotatebox{90}{\phantom{bb}\textbf{Restormer}} & \rotatebox{90}{\phantom{bb}+ FP} & \rotatebox{90}{\phantom{bb}+ FirstLayerDrop} & &
   \includegraphics[height=3cm, width=0.24\textwidth]{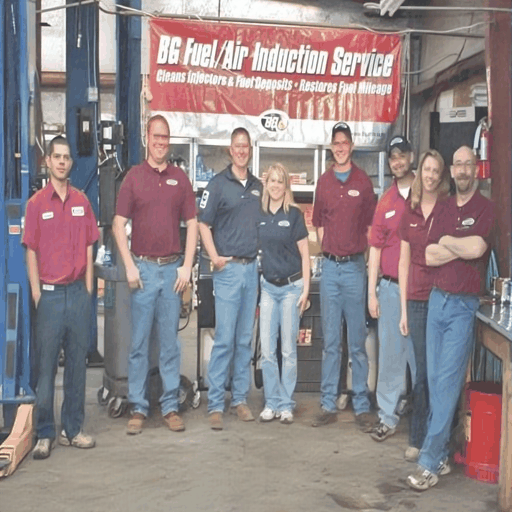}&
  \includegraphics[height=3cm, width=0.24\textwidth]{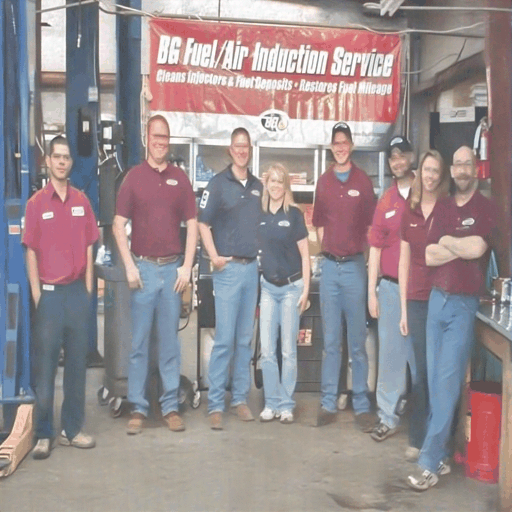}&
  \includegraphics[height=3cm, width=0.24\textwidth]{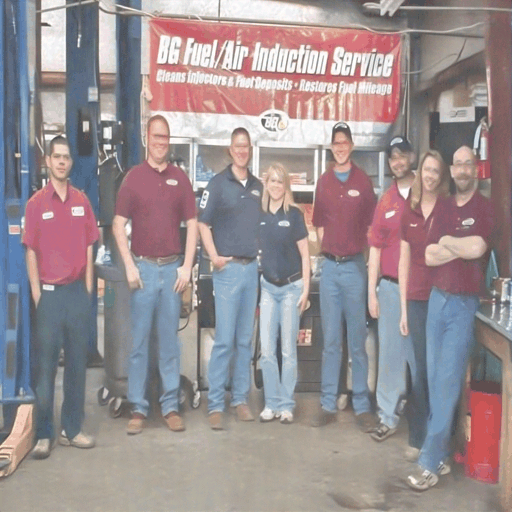}&
  \includegraphics[height=3cm, width=0.24\textwidth]{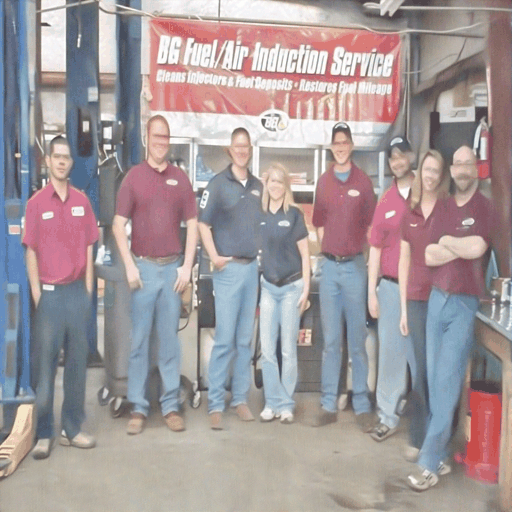}\\

  \rotatebox{90}{\phantom{bbbb}\textbf{Restormer}} & \rotatebox{90}{\phantom{bbbb}+ FP} & \rotatebox{90}{\phantom{bbbb}+ DropHigh} &  \rotatebox{90}{\phantom{bbbb}+ FreqAvgUp} & 
   \includegraphics[height=3cm, width=0.24\textwidth]{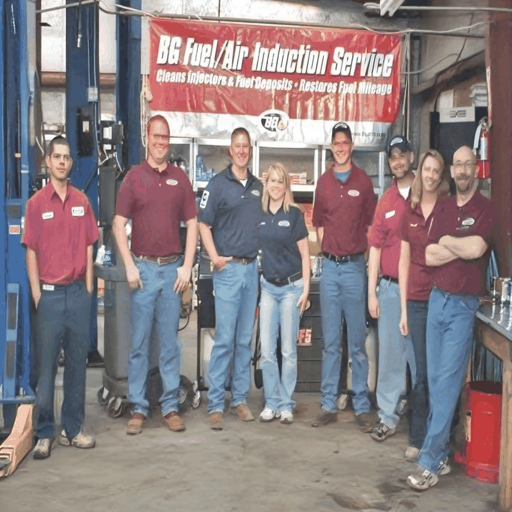}&
  \includegraphics[height=3cm, width=0.24\textwidth]{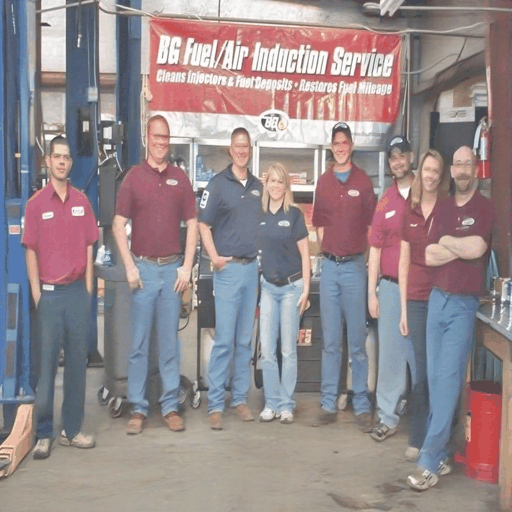}&
  \includegraphics[height=3cm, width=0.24\textwidth]{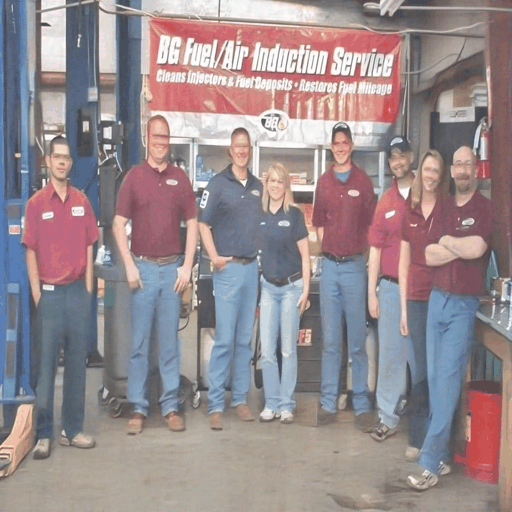}&
  \includegraphics[height=3cm, width=0.24\textwidth]{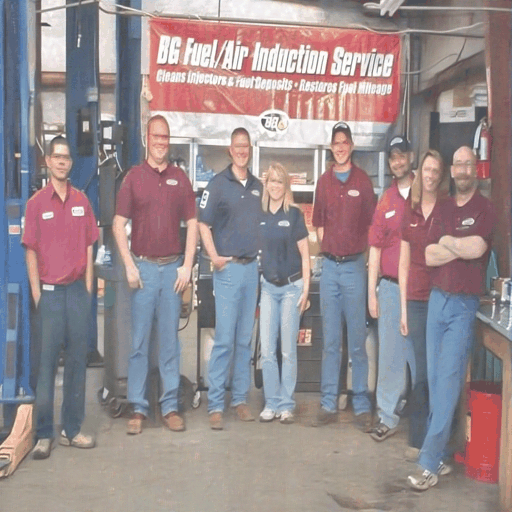}\\

  \rotatebox{90}{\phantom{bb}\textbf{Restormer}} & \rotatebox{90}{\phantom{bb}+ FP} & \rotatebox{90}{\phantom{bb}+ FirstLayerDrop} &  \rotatebox{90}{\phantom{bb}+ FreqAvgUp} & 
   \includegraphics[height=3cm, width=0.24\textwidth]{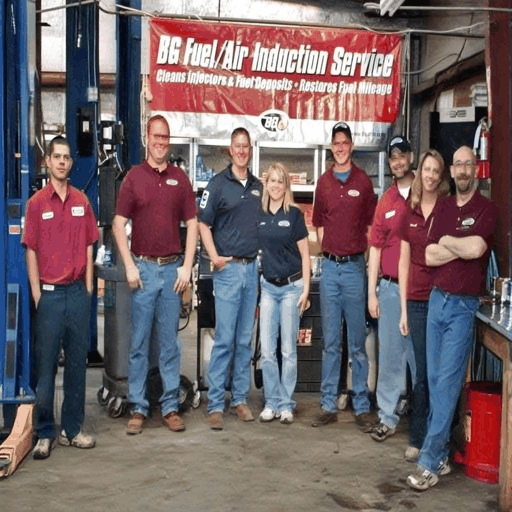}&
  \includegraphics[height=3cm, width=0.24\textwidth]{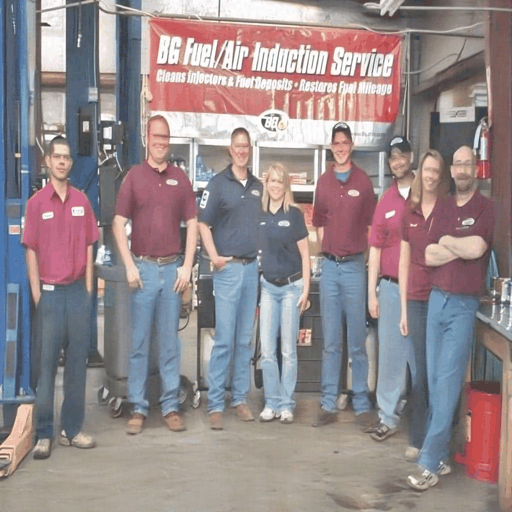}&
  \includegraphics[height=3cm, width=0.24\textwidth]{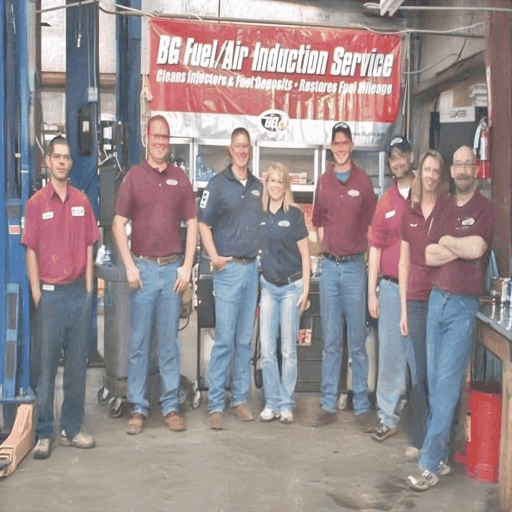}&
  \includegraphics[height=3cm, width=0.24\textwidth]{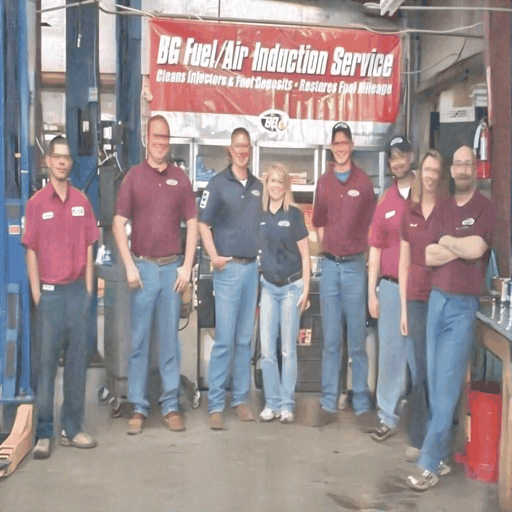}\\

    \rotatebox{90}{\phantom{bbbb}\textbf{Restormer}} & \rotatebox{90}{\phantom{bbbb}+ FP} & \rotatebox{90}{\phantom{bbbb}+ FreqAvgUp} & &
   \includegraphics[height=3cm, width=0.24\textwidth]{figures/deraining_images/BoA_restormer_no_attack_22_restored_after_attack.png}&
  \includegraphics[height=3cm, width=0.24\textwidth]{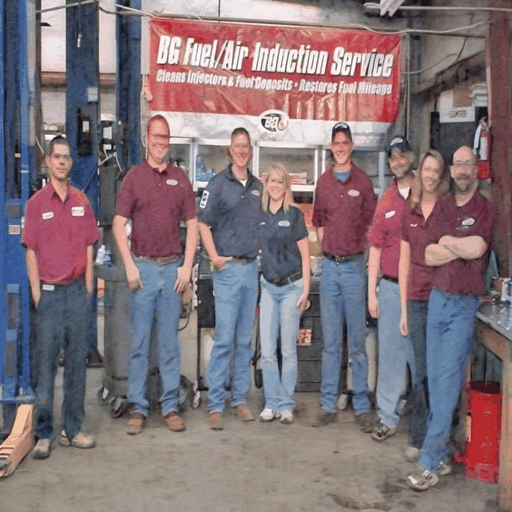}&
  \includegraphics[height=3cm, width=0.24\textwidth]{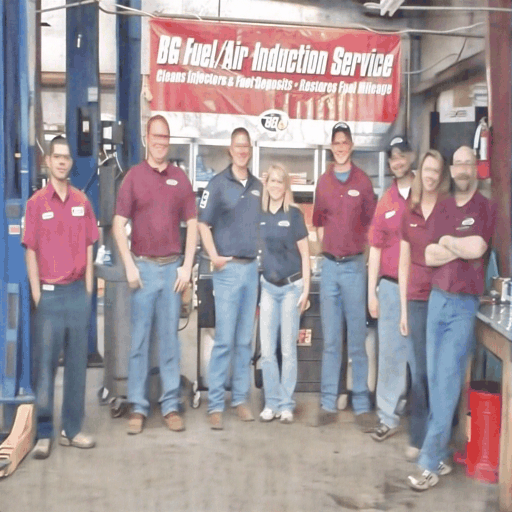}&
  \includegraphics[height=3cm, width=0.24\textwidth]{figures/deraining_images/BoA_restormer_cospgd_20_22_restored_after_attack.png}\\

\end{tabular}
}
\caption{\underline{\textbf{Image Deraining}} results using \textbf{Restormer}. Different sampling methods are being compared qualitatively on normal rainy input images from the Test1200 dataset and input images adversarially attacked using CosPGD attack.}
\label{fig:restormer_deraining_cospgd_full}
\end{figure*}
\begin{figure*}[htb]
    \centering % <-- added
\scalebox{0.75}{
   \begin{tabular}{@{}c@{\hspace{0.2cm}}c@{\hspace{0.1cm}}c@{\hspace{0.1cm}}c@{\hspace{0.1cm}}c@{\hspace{0.1cm}}c@{\hspace{0.1cm}}c@{\hspace{0.1cm}}c@{}}
    \multicolumn{4}{c}{MODEL} & NO ATTACK & 5 iterations & 10 iterations & 20 iterations\\

  \rotatebox{90}{\phantom{bbbb}\textbf{Restormer}} & & & &
  \includegraphics[height=3cm, width=0.24\textwidth]{figures/deraining_images/baseline_restormer_no_attack_22_restored_after_attack.png}&
  \includegraphics[height=3cm, width=0.24\textwidth]{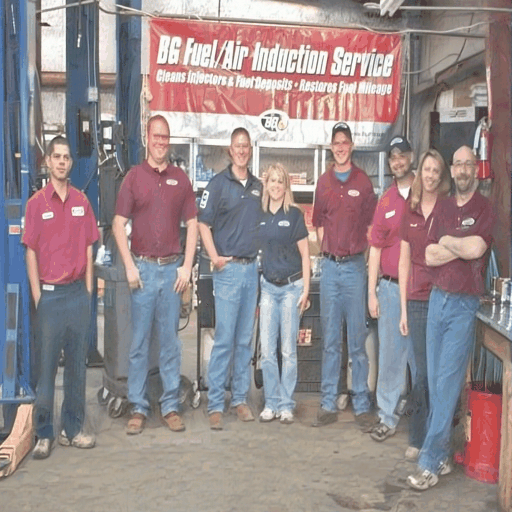}&
  \includegraphics[height=3cm, width=0.24\textwidth]{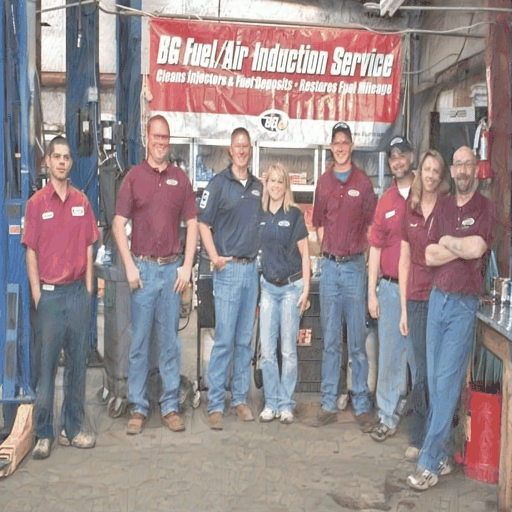}&
  \includegraphics[height=3cm, width=0.24\textwidth]{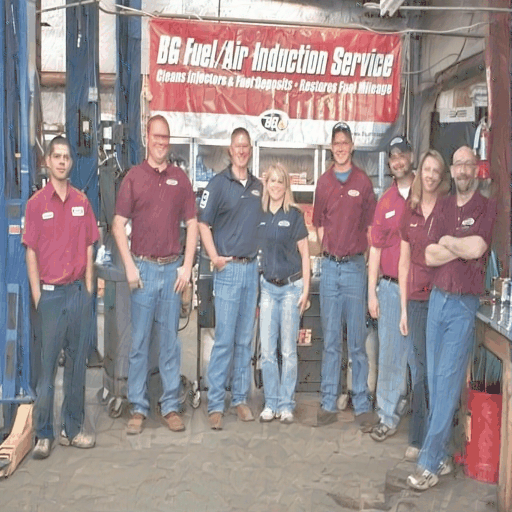}\\

  \rotatebox{90}{\phantom{bbbb}\textbf{Restormer}} & \rotatebox{90}{\phantom{bbbb}+ FP} &  & &
   \includegraphics[height=3cm, width=0.24\textwidth]{figures/deraining_images/restormer_fp_no_attack_22_restored_after_attack.png}&
  \includegraphics[height=3cm, width=0.24\textwidth]{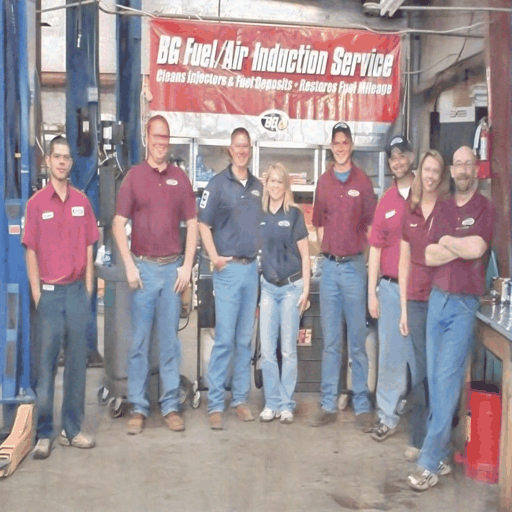}&
  \includegraphics[height=3cm, width=0.24\textwidth]{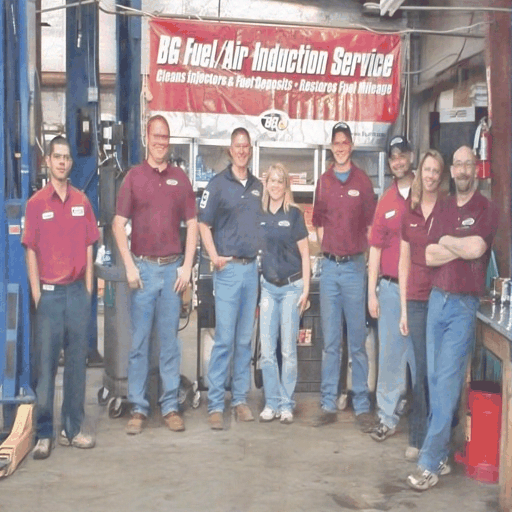}&
  \includegraphics[height=3cm, width=0.24\textwidth]{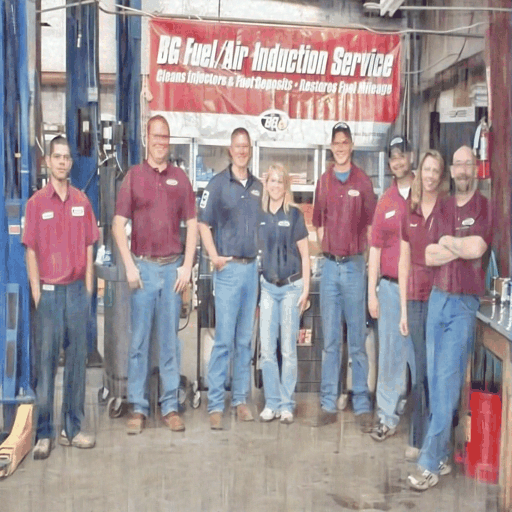}\\

  \rotatebox{90}{\phantom{bbbb}\textbf{Restormer}} & \rotatebox{90}{\phantom{bbbb}+ FP} & \rotatebox{90}{\phantom{bbbb}+ DropHigh} & &
   \includegraphics[height=3cm, width=0.24\textwidth]{figures/deraining_images/restormer_fp_drophigh_no_attack_22_restored_after_attack.png}&
  \includegraphics[height=3cm, width=0.24\textwidth]{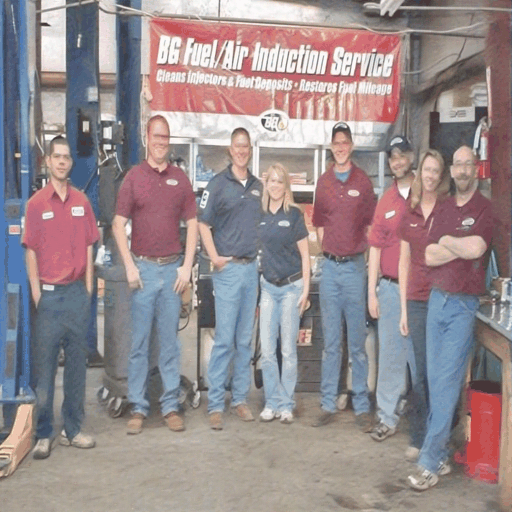}&
  \includegraphics[height=3cm, width=0.24\textwidth]{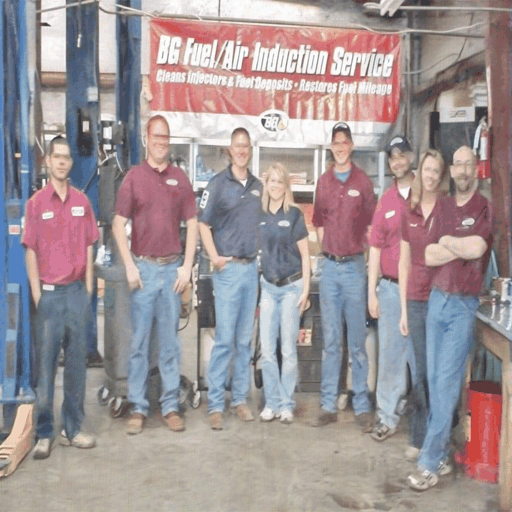}&
  \includegraphics[height=3cm, width=0.24\textwidth]{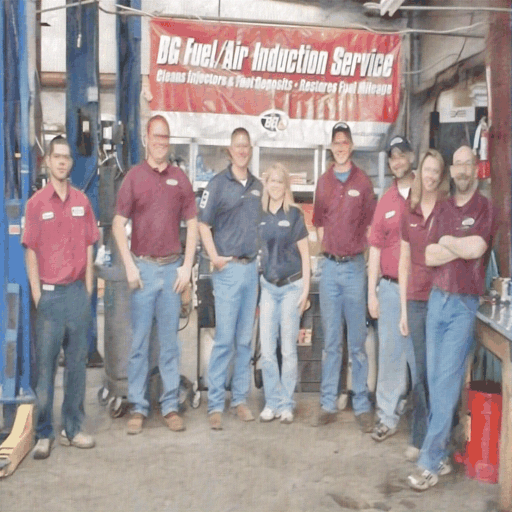}\\

  \rotatebox{90}{\phantom{bb}\textbf{Restormer}} & \rotatebox{90}{\phantom{bb}+ FP} & \rotatebox{90}{\phantom{bb}+ FirstLayerDrop} & &
   \includegraphics[height=3cm, width=0.24\textwidth]{figures/deraining_images/restormer_fp_firstlayerdrop_no_attack_22_restored_after_attack.png}&
  \includegraphics[height=3cm, width=0.24\textwidth]{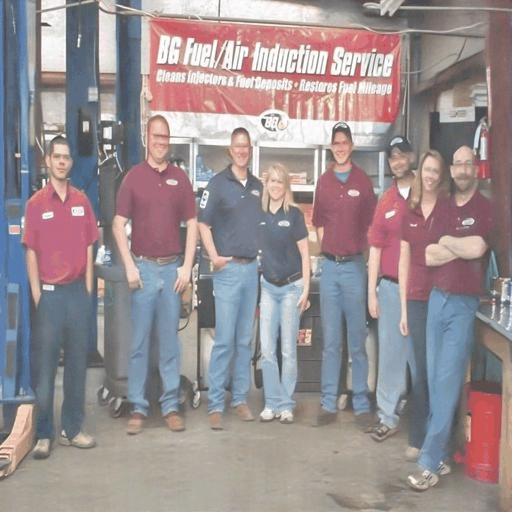}&
  \includegraphics[height=3cm, width=0.24\textwidth]{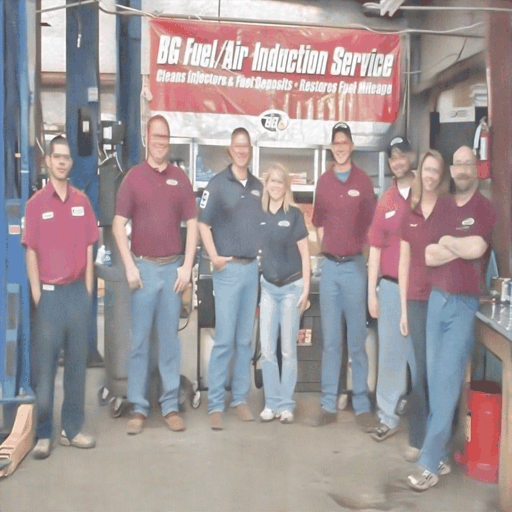}&
  \includegraphics[height=3cm, width=0.24\textwidth]{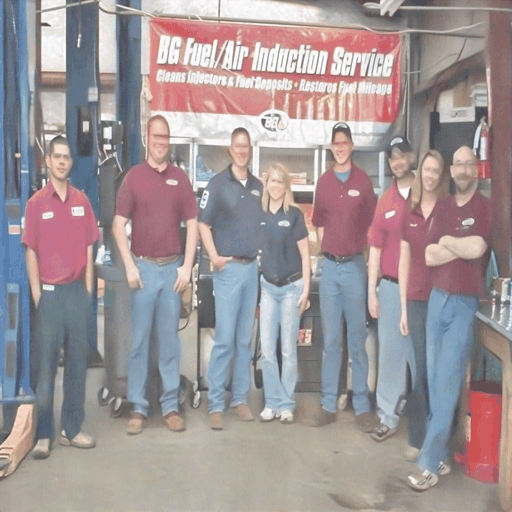}\\

  \rotatebox{90}{\phantom{bbbb}\textbf{Restormer}} & \rotatebox{90}{\phantom{bbbb}+ FP} & \rotatebox{90}{\phantom{bbbb}+ DropHigh} &  \rotatebox{90}{\phantom{bbbb}+ FreqAvgUp} & 
   \includegraphics[height=3cm, width=0.24\textwidth]{figures/deraining_images/restormer_fp_drophigh_freqavgup_no_attack_22_restored_after_attack.png}&
  \includegraphics[height=3cm, width=0.24\textwidth]{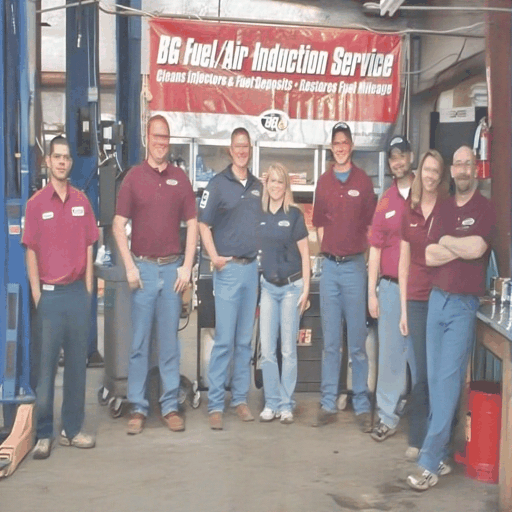}&
  \includegraphics[height=3cm, width=0.24\textwidth]{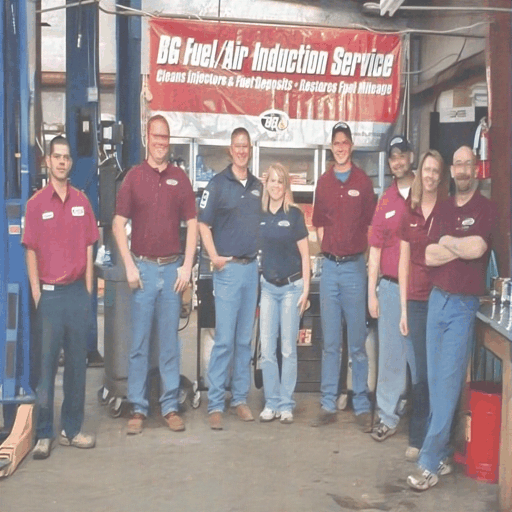}&
  \includegraphics[height=3cm, width=0.24\textwidth]{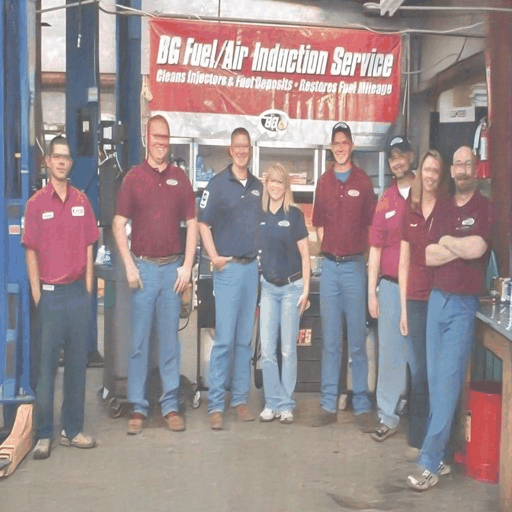}\\

  \rotatebox{90}{\phantom{bb}\textbf{Restormer}} & \rotatebox{90}{\phantom{bb}+ FP} & \rotatebox{90}{\phantom{bb}+ FirstLayerDrop} &  \rotatebox{90}{\phantom{bb}+ FreqAvgUp} & 
   \includegraphics[height=3cm, width=0.24\textwidth]{figures/deraining_images/restormer_fp_firstlayerdrop_freqavgup_no_attack_22_restored_after_attack.png}&
  \includegraphics[height=3cm, width=0.24\textwidth]{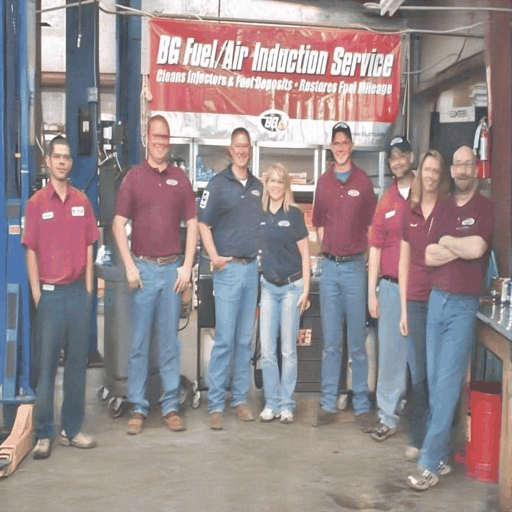}&
  \includegraphics[height=3cm, width=0.24\textwidth]{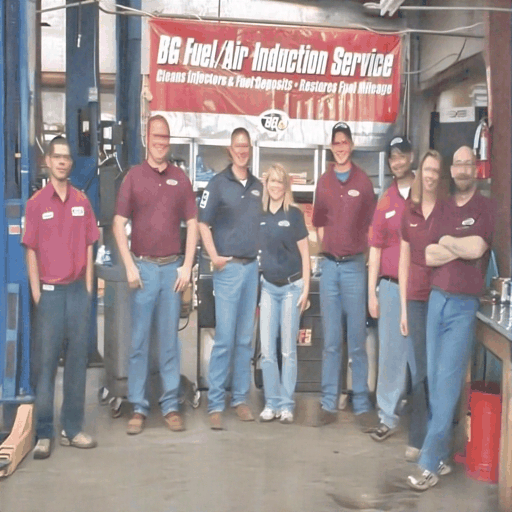}&
  \includegraphics[height=3cm, width=0.24\textwidth]{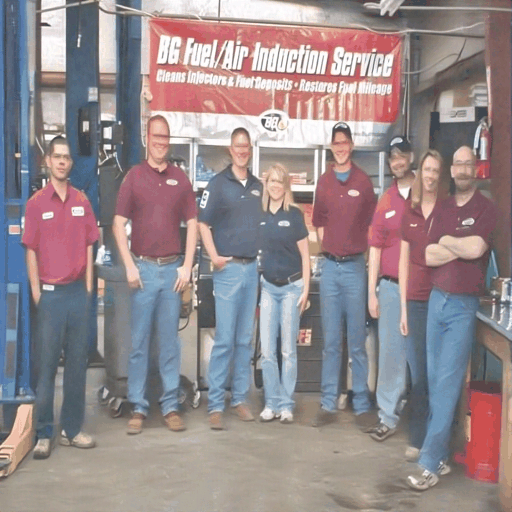}\\

   \rotatebox{90}{\phantom{bbbb}\textbf{Restormer}} & \rotatebox{90}{\phantom{bbbb}+ FP} & \rotatebox{90}{\phantom{bbbb}+ FreqAvgUp} & &
   \includegraphics[height=3cm, width=0.24\textwidth]{figures/deraining_images/BoA_restormer_no_attack_22_restored_after_attack.png}&
  \includegraphics[height=3cm, width=0.24\textwidth]{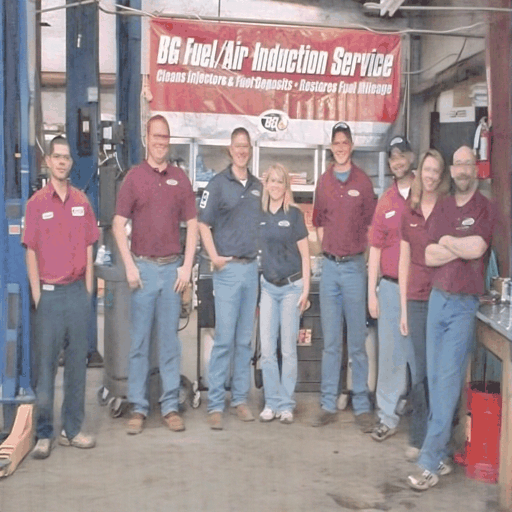}&
  \includegraphics[height=3cm, width=0.24\textwidth]{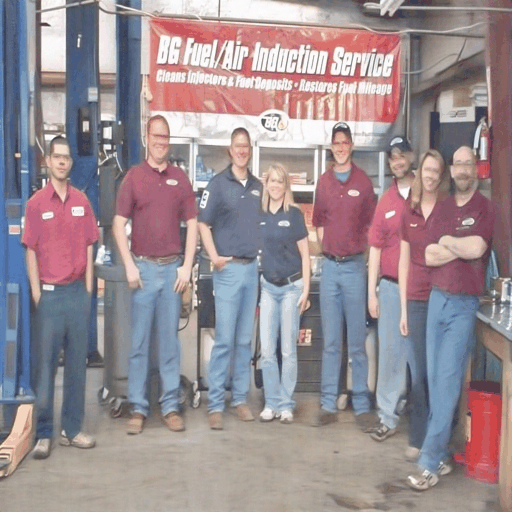}&
  \includegraphics[height=3cm, width=0.24\textwidth]{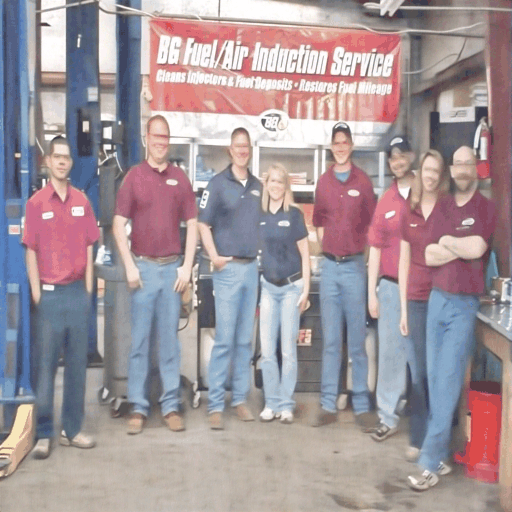}\\

\end{tabular}
}
\caption{\underline{\textbf{Image Deraining}} results using \textbf{Restormer}. Different sampling methods are being compared qualitatively on normal rainy input images from the Test1200 dataset and input images adversarially attacked using PGD attack.}
\label{fig:restormer_deraining_pgd_full}
\end{figure*}

\subsection{Validation of Focus on Low-Frequencies. }
\label{subsec:appendix:proof_low_frequencies_focus}
Since the proposed methods are learning a parameter for mixing the low-frequency and high-frequency information from the feature maps, we investigate whether the model focuses on low-frequency information.
We do so by observing the values of $\alpha$ and $\beta$ parameters learned for each downsampling and upsampling step.
As shown in \cref{eqn:downsampling:learnable} and \cref{eqn:upsampling:beta_mixing}, a value of $\alpha$ and $\beta$ less than 0.5 would mean the model is focusing more on low-frequency information.
In \cref{tab:alpha_values}, we report these values learned for \textbf{FrequencyPreservedPooling (FP)} and its variants.
Similarly, in \cref{tab:alpha_beta_values}, we report the values learned for \textbf{FrequencyPreservedPooling} along with the upsampling operation variants \textbf{FreqAvgUp} and \textbf{SplitUp}.
In both cases, we observe that in all but one instance, the learned value for mixing the information is less than 0.5 strongly indicating that the model focuses on low-frequencies.

\begin{table}[htbp]
    \centering
    \caption{\underline{\textbf{Image Deblurring}} studies. Values learnt for the mixing of feature maps, $\alpha$ during downsampling by different downsampling operations. Restormer has three downsampling steps and thus three values of $\alpha$. The initial value of $\alpha$ is 0.3, to induce a bias towards low frequencies. Here the upsampling operation used is PixelShuffle. Notations \textbf{+DropHigh} and \textbf{+FirstLayerDrop} mean the same as in \cref{tab:restormer_encoder_ablation}.}
    \scalebox{1.0}{ %0.75 for ECAI
    \begin{tabular}{lccc}
    \toprule
       Model  & $\alpha$ 0 & $\alpha$ 1 & $\alpha$ 2    \\
       \midrule

        Restormer + FP & 0.0686 & 0.1703 & 0.0886  \\
        %Drop FLC Restormer + ADV & 0.0233 & 0.1260 & 0.0825  \\
        %\midrule
        ~~~~ + FP + DropHigh & 0.5239 & 0.2861 & 0.1382  \\
        %No Drop FLC Restormer + ADV & 0.4553 & 0.3147 & 0.1333  \\
        %\midrule
        ~~~~ + FP + FirstLayerDrop & 0.1312 & 0.2815 & 0.1246  \\
        %First Layer Drop FLC Restormer + ADV & 0.0435 & 0.3103 & 0.1456  \\

        \bottomrule
    \end{tabular}
    }
    \label{tab:alpha_values}
\end{table}
\begin{table}[htbp]
    \centering
    \caption{\underline{\textbf{Image Deblurring}} studies. Values learnt for the mixing of feature maps, $\alpha$ during downsampling and $\beta$ during upsampling by different upsampling operations. Restormer has three downsampling steps and three upsampling steps, thus three values of each. The initial value of both $\alpha$ and $\beta$ is 0.3, to induce a bias towards low frequencies. Here the model used is ``BoA Restormer''.}
    \scalebox{0.81}{
    \begin{tabular}{lccc|ccc}
    \toprule
       Upsampling Method  & $\alpha$ 0 & $\alpha$ 1 & $\alpha$ 2  & $\beta$ 0 & $\beta$ 1 & $\beta$ 2  \\
       \midrule

        FreqAvgUp & 0.4866 & 0.2805 & 0.2125  & 0.2517 & 0.2032 & 0.2246  \\

        %Freq Avg Up + ADV & 0.4358 & 0.3342 & 0.2458  & 0.2461 & 0.1917 & 0.2313  \\

        SplitUp & 0.4990 & 0.2915 & 0.2281  & 0.2656 & 0.3306 & 0.3035  \\

        %Split Up + ADV & 0.4517 & 0.3274 & 0.2629  & 0.2725 & 0.3357 & 0.2568  \\
        
        \bottomrule
    \end{tabular}
    }
    \label{tab:alpha_beta_values}
\end{table}
Please note, that we provide an analysis of the importance of symmetry in \cref{subsec:analysis:symmetry}.

\subsection{Further Analysis on the Importance of Symmetry}
\label{subsec:analysis:symmetry}
\begin{table*}[htbp]
    \centering
    \caption{\underline{\textbf{Image Deblurring}} studies. Different Upsampling techniques used in the decoder compared against clean performance and against CosPGD and PGD attacks with various attack strengths. Attack strength increases with the number of attack iterations~(itrs). The baseline Restormer uses Pixel Shuffle for upsampling, and Pixel Unshuffle for downsampling. First, we replace the downsampling operation with \textbf{FrequencyPreservedPooling (FP)}, while still using Pixel Shuffle for upsampling. Next, we ablate over the choice of the upsampling operation by replacing \textbf{PixelShuffle} with \textbf{SplitUp} and \textbf{FreqAvgUp}.}
    %\scriptsize
    \scalebox{0.7}{
    \begin{tabular}{@{}lcc|cc|cc|cc|cc|cc|cc@{}}
    \toprule
    \multirow{3}{*}{Architecture} & \multicolumn{2}{c|}{Test} & \multicolumn{6}{c|}{CosPGD} & \multicolumn{6}{c}{PGD} \\
   & & & \multicolumn{2}{c|}{5 attack itrs } & \multicolumn{2}{c|}{10 attack itrs } & \multicolumn{2}{c|}{20 attack itrs } & \multicolumn{2}{c|}{5 attack itrs } & \multicolumn{2}{c|}{10 attack itrs } & \multicolumn{2}{c}{20 attack itrs } \\
   & PSNR & SSIM & PSNR & SSIM & PSNR & SSIM & PSNR & SSIM & PSNR & SSIM & PSNR & SSIM & PSNR & SSIM \\
    \toprule
         
        Restormer & 31.99 & \textbf{0.9635} & 11.36 & 0.3236 & 9.05 & 0.2242 & 7.59 & 0.1548 & 11.41 & 0.3256 & 9.04 & 0.2234 & 7.58 & 0.1543 \\
         
        %Restormer + FP + Pixel Shuffle upsampling & 29.95 & 0.9395 & 12.17 & 0.3024 & 10.35 & 0.2117 & 9.44 & 0.1651 & 12.17 & 0.3018 & 10.35 & 0.2112 & 9.45 & 0.1655 \\                

        Restormer + FP & 29.95 & 0.9395 & 12.17 & 0.3024 & 10.35 & 0.2117 & 9.44 & 0.1651 & 12.17 & 0.3018 & 10.35 & 0.2112 & 9.45 & 0.1655 \\                
 
        ~~~~~~~ + FP + SplitUp upsampling & 6.26 & 0.2370 & 6.31 & 0.1971 & 6.06 & 0.1854 & 5.79 & 0.1702 & 6.29 & 0.1964 & 6.04 & 0.1847 & 5.78 & 0.1697 \\

        ~~~~~~~ + FP + FreqAvgUp upsampling & 26.99 & 0.8806 & 23.9 & 0.7578 & 21.76 & 0.6875 & 21.00 & 0.6351 & 22.84 & 0.7290 & 18.91 & 0.6390 & 18.86 & 0.601 \\

    \bottomrule
    \end{tabular}    
    }
    \label{tab:restormer_decoder_ablation}
\end{table*}

Following, we ablate over the symmetry introduced in the architecture, by making the downsampling and upsampling operations mirror images to each other.
One might hypothesize, given the downsampling operation \textbf{FrequencyPreservedPooling} treats the low and high-frequency information differently, merely treating them differently again during upsampling might be sufficient, and performing an entirely symmetric architecture may not be necessary.
Thus, we device \textbf{SplitUp} as an upsampling operation, that merely splits the feature maps into low and high frequencies and learns to combine them using a learnable parameter $\beta$.
However, as shown \cref{fig:restormer_cospgd_attack_upsampling}, this leads to improper sampling, causing severe artifacts in the restored images, for both when under adversarial attack and in normal conditions.
This observation is reflected in the quantitative results reported in \cref{tab:restormer_decoder_ablation}, where the performance of \textbf{SplitUp} is inadequate.

\subsection{Additional Experimental Setup Details}
\label{sec:appendix:exp_setup_details}

\paragraph{Evaluation Metrics. }We report the Peak Signal-to-Noise Ratio (PSNR) and Structural similarity (SSIM)~\cite{ssim} scores of the deblurred images w.r.t. to the ground truth images, averaged over all images in the test set.
A higher PSNR indicates a better quality image or an image closer to the ground truth image.
A higher SSIM score corresponds to better higher similarity between the two compared images.

\paragraph{Training Regime. }To train our proposed architectures, for the Restormer variants, we follow the same training regime and hyperparameters as those used by \cite{zamir2022restormer} for training their proposed Restormer. For the NAFNet variants, we follow the same training regime as used by \cite{chen2022simple}.

\paragraph{Compute Resources. }For training and evaluation, we used a single NVIDIA A100 GPU or NVIDIA A40 GPU with 48GB VRAM for each run except runs using FSNet\cite{cui2024image}. Due to the excessive memory requirements of FSNet, for these experiments, we used a single NVIDIA A100 GPU with 80GB VRAM.

\paragraph{Dataset. } For the \textbf{Image Deblurring} task we use the GoPro image deblurring dataset~\cite{gopro}.
This dataset consists of 3214 real-world images with realistic blur and their corresponding ground truth (deblurred images) captured using a high-speed camera.
The dataset is split into 2103 training images and 1111 test images.

For the \textbf{Image Denoising} task, we use the Smartphone Image Denoising Dataset (SSID)~\cite{ssid}.
This dataset consists of 160 noisy images taken from 5 different smartphones and their corresponding high-quality ground truth images.

For the \textbf{Image Deraining} task, we follow the regime of \cite{zamir2022restormer} and use the Rain13K dataset for training and use the Rain100H~\cite{rain100h_l}, Rain100L~\cite{rain100h_l}, Test100~\cite{test100} and Test1200~\cite{test1200} datasets for testing.

\newpage
\FloatBarrier

\subsection{Providing code for the work}
\label{sec:appendix:code}

Following, we provide the Python code for the two prominent downsampling operations, \textbf{FrequencyPreservedPooling} (please refer to \cref{subsec:appendix:code:fp_code}) and \textbf{FP + DropHigh} (please refer to \cref{subsec:appendix:code:fp_drop_high_code}) and one prominent upsampling operation, \textbf{FreqAvgUp} (please refer to \cref{subsec:appendix:code:FreqAvgUp}) introduced in this work.

\subsubsection{Code for FrequencyPreservedPooling}
\label{subsec:appendix:code:fp_code}
\begin{lstlisting}
import torch
import torchvision.transforms as T
from torch import nn as nn
from torch.nn import functional as F

class FrequencyPreservedPooling(nn.Module):
    def __init__(self, channels = None, test_wo_drop_alpha=False, transpose=True, test_drop_alpha=False, stop = False, half_precision = False, padding = "reflect"):
        self.transpose = transpose # Decision to transpose in the spatial domain before and after Frequency domain conversion for stability        
        self.test_wo_drop_alpha = test_wo_drop_alpha # During inference, to explicitely not drop High-Frequencies , used just for personal ablation studies  
        self.test_drop_alpha = test_drop_alpha # To Drop High-Frequencies explicitely during inference, used just for personal ablation studies        
        self.half_precision = half_precision # To use half precision
        self.padding = padding # To decide which padding to use, zero or mirror

        super(FrequencyPreservedPooling, self).__init__()
        self.alpha = nn.Parameter(torch.tensor(0.3), requires_grad = True) # The single parameter learned to mix the high-frequency and low-frequency information
        self.downsample_high = nn.PixelUnshuffle(2) # The pixel unshuffle path for the high-frequencies
        

    def forward(self, x):        
        device = x.device

        orig_x_size = x.shape
        orig_x = x.clone()
        x = F.pad(x, (3*x.shape[-1]//4 +1, 3*x.shape[-1]//4, 3*x.shape[-2]//4 +1, 3*x.shape[-2]//4), mode=self.padding) # Padding before FFT for stability

        if self.transpose:
            x = x.transpose(2,3)

        in_freq = torch.fft.fftshift(torch.fft.fft2(x.to(torch.float32), norm='forward'))

        low_part = in_freq[:,:,int(x.shape[2]/4):int(x.shape[2]/4*3),int(x.shape[3]/4):int(x.shape[3]/4*3)] # Low pass frequencies cut-out
        low_part =  torch.abs(torch.fft.ifft2(torch.fft.ifftshift(low_part), norm='forward'))

        if self.half_precision:
            low_part = low_part.half()

        if self.transpose:
            low_part = low_part.transpose(2,3)     

        low_part = torch.cat((low_part, low_part, low_part, low_part), dim=1)        
        if self.test_drop_alpha:
            return T.CenterCrop((orig_x_size[-2]//2, orig_x_size[-1]//2))(low_part)

        zeroed_high = torch.zeros_like(in_freq)
        zeroed_high[:,:,int(x.shape[2]/4):int(x.shape[2]/4*3),int(x.shape[3]/4):int(x.shape[3]/4*3)] = in_freq[:,:,int(x.shape[2]/4):int(x.shape[2]/4*3),int(x.shape[3]/4):int(x.shape[3]/4*3)] # Low pass filter

        zeroed_high =  torch.abs(torch.fft.ifft2(torch.fft.ifftshift(zeroed_high), norm='forward'))
        if self.half_precision:
            zeroed_high = zeroed_high.half()
        if self.transpose:
            zeroed_high = zeroed_high.transpose(2,3)
        zeroed_high = T.CenterCrop((orig_x_size[-2], orig_x_size[-1]))(zeroed_high)

        high_part = orig_x - zeroed_high     
        high_part = self.downsample_high(high_part)
        self.alpha = self.alpha.to(device)
        high_part = high_part*self.alpha    
        return (T.CenterCrop((orig_x_size[-2]//2, orig_x_size[-1]//2))(low_part)*(1- self.alpha)) + high_part        
\end{lstlisting}

\subsubsection{Code for FP + DropHigh}
\label{subsec:appendix:code:fp_drop_high_code}
\begin{lstlisting}
import torch
import torchvision.transforms as T
from torch import nn as nn
from torch.nn import functional as F

class FrequencyPreservedPooling_DropHigh(nn.Module):
    # Implementation is the same as FrequencyPreservedPooling, except self.drop which drops high-frequencies by propagating only the low-frequencies for about 30% of the time.
    
    def __init__(self, channels = None, test_wo_drop_alpha=False, transpose=True, test_drop_alpha=False, stop = False, half_precision = False, padding = "reflect"):
        self.transpose = transpose # Decision to transpose in the spatial domain before and after Frequency domain conversion for stability 

        self.test_wo_drop_alpha = test_wo_drop_alpha # During inference, to explicitely not drop High-Frequencies , used just for personal ablation studies  
        self.test_drop_alpha = test_drop_alpha # To Drop High-Frequencies explicitely during inference, used just for personal ablation studies        
        self.half_precision = half_precision # To use half precision
        self.padding = padding # To decide which padding to use, zero or mirror
     
        super(FrequencyPreservedPooling_DropHigh, self).__init__()
        self.alpha = nn.Parameter(torch.tensor(0.3), requires_grad = True) # The single parameter learned to mix the high-frequency and low-frequency information
        self.downsample_high = nn.PixelUnshuffle(2) # The pixel unshuffle path for the high-frequencies
        self.drop = 0 # Decision to drop or not, with 30% probability, taken later

    def forward(self, x):        
        device = x.device

        orig_x_size = x.shape
        orig_x = x.clone()

        x = F.pad(x, (3*x.shape[-1]//4 +1, 3*x.shape[-1]//4, 3*x.shape[-2]//4 +1, 3*x.shape[-2]//4), mode=self.padding)


        if self.transpose:
            x = x.transpose(2,3)

        in_freq = torch.fft.fftshift(torch.fft.fft2(x.to(torch.float32), norm='forward'))#.half()

        low_part = in_freq[:,:,int(x.shape[2]/4):int(x.shape[2]/4*3),int(x.shape[3]/4):int(x.shape[3]/4*3)]
 
        low_part =  torch.abs(torch.fft.ifft2(torch.fft.ifftshift(low_part), norm='forward'))#.half()
        if self.half_precision:
            low_part = low_part.half()

        if self.transpose:
            low_part = low_part.transpose(2,3)     

        low_part = torch.cat((low_part, low_part, low_part, low_part), dim=1)
        self.drop = torch.tensor(np.random.choice(2, replace=True, p=[0.3, 0.7])).to(device) # Decision to drop or not, with 30% probability, taken NOW
        if self.test_drop_alpha:
            return T.CenterCrop((orig_x_size[-2]//2, orig_x_size[-1]//2))(low_part)
        elif self.drop == 0 and not self.test_wo_drop_alpha:
            return T.CenterCrop((orig_x_size[-2]//2, orig_x_size[-1]//2))(low_part)
        else:
            zeroed_high = torch.zeros_like(in_freq)
            zeroed_high[:,:,int(x.shape[2]/4):int(x.shape[2]/4*3),int(x.shape[3]/4):int(x.shape[3]/4*3)] = in_freq[:,:,int(x.shape[2]/4):int(x.shape[2]/4*3),int(x.shape[3]/4):int(x.shape[3]/4*3)]

                           

            zeroed_high =  torch.abs(torch.fft.ifft2(torch.fft.ifftshift(zeroed_high), norm='forward'))
            if self.half_precision:
                zeroed_high = zeroed_high.half()
            if self.transpose:
                zeroed_high = zeroed_high.transpose(2,3)
            zeroed_high = T.CenterCrop((orig_x_size[-2], orig_x_size[-1]))(zeroed_high)

            high_part = orig_x - zeroed_high
            

            high_part = self.downsample_high(high_part)


            self.alpha = self.alpha.to(device)
            high_part = high_part*self.alpha                        
            
            return (T.CenterCrop((orig_x_size[-2]//2, orig_x_size[-1]//2))(low_part)*(1- self.alpha)) + high_part
\end{lstlisting}

\subsubsection{Code for FreqAvgUp}
\label{subsec:appendix:code:FreqAvgUp}
\begin{lstlisting}
import torch
import torchvision.transforms as T
from torch import nn as nn
from torch.nn import functional as F

class FreqAvgUpsample(nn.Module):
    def __init__(self, n_feat, padding='zero', transpose=False):
        super(FreqAvgUpsample, self).__init__()
        self.padding = 'constant' if padding =='zero' else 'mirror' # Padding in the spatial domain before FFT for numerical stability
        self.body = nn.Conv2d(n_feat, n_feat*2, kernel_size=3, stride=1, padding=1, bias=False)   # Doubling the number of channels before upsampling to match the implementation of PixelShuffle     
        self.beta = nn.Parameter(torch.tensor(0.3), requires_grad = True) # The single parameter learned for mixing low and high-frequency information
        self.shuffle = nn.PixelShuffle(2) # Path for propagating the high-frequency information
        self.transpose = transpose     # alternate steps transpose in the spatial domain for numerical stability of FFT and IFFT   

    def forward(self, x):
        dtype = x.dtype
        x = self.body(x)        
        channels = x.shape[1]
        
        if self.transpose:
            x = x.transpose(2,3)
        freq = torch.fft.fft2(x.to(torch.float32), norm='forward')

        avg_list, avg_channel_list = [], []
        for i in range(0, freq.shape[1], 4):
            avg = torch.mean(freq[:,i:i+4,:,:], dim=1)
            avg = torch.unsqueeze(avg, dim=1)
            avg_channels = torch.cat([avg]*4, dim=1)
            avg_list.append(avg)  # This contains the mean frequencies of a set of 4 channels, this slightly resembles a low-pass filter
            avg_channel_list.append(avg_channels)    # This contains the mean frequencies of a set of 4 channels, stacked 4 times along the channel dimension to mimic reverse of PixelUnshuffle    
        
        avg_list = torch.cat(avg_list, dim=1) # Like a low-pass filter
        avg_channel_list = torch.cat(avg_channel_list, dim=1) 
                
        freq = freq - avg_channel_list # High-pass filter
        freq = torch.fft.ifft2(freq, norm='forward').to(dtype)
        highFreq = self.shuffle(freq) # Continuing the high-frequency path

        padding = F.pad(avg_list, (x.shape[-1], 0, x.shape[-2], 0), mode=self.padding) # Padding the low-frequencies to increase resolution in the frequency domain
        freqUp = torch.fft.ifft2(padding, norm='forward').to(dtype)        

        if self.transpose:
            freqUp = freqUp.transpose(2, 3)
            highFreq = highFreq.transpose(2, 3)
        
        return freqUp*(1-self.beta) + self.beta*highFreq # Mixing low and high-frequency information
\end{lstlisting}

\vfill

\end{document}